\documentclass[12pt,a4paper,twoside,openright]{extreport}

\usepackage{amsmath}                            
\usepackage{csquotes}                           
\usepackage{enumitem}                           
\usepackage[
    a4paper,
    top=2cm,bottom=2cm,
    outer=2cm,inner=3cm,
    includeheadfoot
]{geometry}                                     
\usepackage{graphicx}                           
\usepackage{listings}
\usepackage{xcolor}

\lstnewenvironment{minted}[2][]{%
  \lstset{#1}%
}{}
\usepackage[output-decimal-marker={.}]{siunitx} 
\usepackage{subcaption}                         
\usepackage{amssymb}
\usepackage{tabularx}
\usepackage{booktabs}

\usepackage[english]{babel}
\usepackage[backend=bibtex,style=ieee]{biblatex}
\bibliography{bibliografia}

\usepackage{fontspec}
\usepackage{setspace}
\title{Augmented Reality Interfaces for Human-Robot Collaboration: Development of a ROS 2-Based Sensor Streaming Framework and Validation via SLAM Algorithms}
\author{Alessandro Rubert}
\date{September 22, 2026}
\newcommand{\supervisor}{Prof. Stefano Ghidoni}
\newcommand{\assistantsupervisor}{Dott. Matteo Terreran, PhD}

\begin{document}
    \pagenumbering{roman}
    \pagestyle{empty} 

    \begin{titlepage}
    \newgeometry{hmargin=2.5cm,vmargin=2cm}
        \begin{figure}
            \centering
            \begin{subfigure}[b]{0.4\textwidth}
                \includegraphics[width=\textwidth]{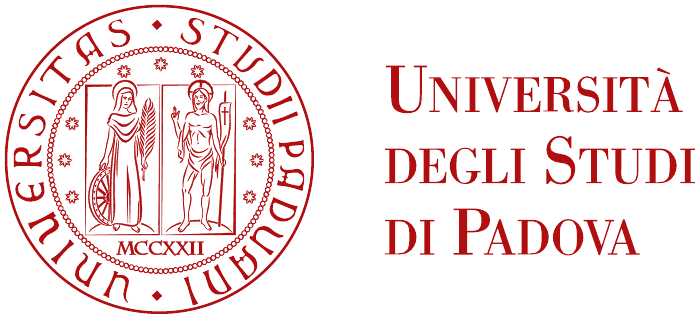}
            \end{subfigure}
            \hfill
            \begin{subfigure}[b]{0.3\textwidth}
                \includegraphics[width=\textwidth]{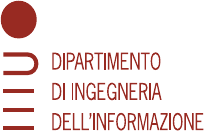}
            \end{subfigure}
        \end{figure}
    
        \vspace*{\stretch{0.5}}
    
        \begin{center}
            \makeatletter 

            \textsc{DEPARTMENT OF INFORMATION ENGINEERING}\\
            \vspace*{\stretch{0.1}}
            \textsc{BACHELOR'S DEGREE IN COMPUTER ENGINEERING}\\
            \vspace*{\stretch{0.1}}
            \textsc{BACHELOR'S THESIS}
    
            \vspace*{\stretch{0.5}}
            \LARGE
            \textbf{\@title}
    
            \vspace*{\stretch{1}}
            \normalsize
            \begin{tabular*}{\textwidth}{l @{\extracolsep{\fill}} r}
                \textbf{Supervisor} & \textbf{Candidate} \\
                \supervisor       & \@author           \\
                \\
                \textbf{Co-supervisor} \\
                \assistantsupervisor \\
            \end{tabular*}
    
            \vspace*{\stretch{2}}
            \textsc{ACADEMIC YEAR 2025-2026} \\
            \vspace*{\stretch{0.1}}
            Graduation Date: \@date
        
            \makeatother 
        \end{center}
    \restoregeometry
\end{titlepage}

    \cleardoublepage
    
    \vspace*{\stretch{1}}
\begin{flushright}
    \textit{``Simplicity is the ultimate sophistication.'' \\ Leonardo da Vinci}
\end{flushright}
\vspace{\stretch{4}}
    \cleardoublepage

    \pagestyle{plain} 

    \chapter*{Abstract}
    In recent years, Human-Robot Collaboration (HRC) has taken on a central role in Industry 4.0 and collaborative robotics, demanding communication channels that are increasingly bidirectional, intuitive, and efficient. In this context, Augmented Reality (AR) presents itself as a fundamental enabling technology, capable of both displaying information to the operator and gathering spatial data about the surrounding environment. This thesis presents the development of a sensor streaming framework that connects the Magic Leap 2 AR headset with the ROS 2 (Robot Operating System) ecosystem. Using the Unity development environment and the ROS-TCP-Connector package, an on-board application for the headset was developed, capable of acquiring real-time data from the integrated sensors (pose tracking, cameras, and environmental sensors) and publishing it to dedicated ROS 2 topics. In order to test the accuracy, latency, and robustness of the generated data stream, the framework was validated using SLAM (Simultaneous Localization and Mapping) algorithms known in the literature. The experimental results demonstrate that the proposed architecture ensures stable data transmission, laying the groundwork for safe real-time interaction and shared spatial awareness, and opening up new perspectives for the control and supervision of robotic systems in complex HRC scenarios.
    \cleardoublepage

    \tableofcontents
    \cleardoublepage
    
    \listoffigures
    \cleardoublepage 
    
    \pagenumbering{arabic}

    \chapter{Introduction}

\section{Robotics in enterprises}

\subsection{Hard automation. Industry 3.0}
Between the 1960s and the 1980s, the first industrial robots made their appearance in factories. These machines were robust, heavy, and designed to systematically repeat simple operations and programmed movements without perceiving the environment in any way. This type of robotics is defined as ``hard'' because the lines are dedicated to a single product and involve high reprogramming costs. These technological solutions were used to replace humans in the most arduous, hazardous, and repetitive tasks, such as handling incandescent die-cast parts or spot welding. Subsequently, between the 1980s and the 2000s, industrial robots reached maturity thanks to the widespread adoption of microprocessors and the first computer vision systems. For example, 6-axis articulated arms emerged during this period. These were later equipped with cameras that allowed movements to be corrected to adapt to slight misalignments of the parts. Thanks to these changes, the machines became faster and more accurate. All these innovations made it possible to have entire production lines fully automated, capable of ensuring millimetric precision and operating through the night as well. However, for safety reasons, these machines were strictly segregated inside metal cages or enclosed cells equipped with photocells and interlocks that halted the machine in the event of an opening, preventing fatal collisions with operators and ensuring functional safety.

\subsection{Flexible and collaborative automation. Industry 4.0}
From the early 2000s onwards, market demands caused the paradigm of hard automation to become obsolete. Indeed, today's market is far more dynamic. A clear example of this change is represented by the automotive sector. Historically, when the assembly line was designed for the production of a single model with very few variants (such as colour options alone), the adoption of a rigid production line was the ideal and efficient choice. Today, on the other hand, automotive companies launch car models on the market that have thousands of possible optional combinations. Other examples include the world of E-commerce and Fast Fashion. For instance, if a lipstick or a shoe goes viral on a social network, demand surges from 100 to 50,000 units within a few days. Two weeks later, the trend is already over, and the line configured for packaging lipsticks finds itself having to handle completely different shapes. In consumer electronics as well, production is extremely dynamic. Companies producing computers, smartphones, or household appliances launch a new model with different shapes and specifications every 9--12 months, resulting in modifications to the entire production line. In such dynamic contexts, hard automation---which involves design phases lasting several months only to be dismantled upon the release of a new model---proves to be unfeasible as well as inefficient and expensive. Therefore, a paradigm shift has been necessary, one that is still ongoing today. It is necessary to transition from hard automation to flexible automation that can be reprogrammed and adapted to requirements in a very short time. This is made possible by collaborative robots, that is, machines entirely designed to work in close contact with human beings without causing them harm. Cobots stand out for being easily reprogrammable by human operators, often through intuitive learning or simplified interfaces. Should the need arise to overhaul the entire production line, these machines can be quickly moved, reconfigured, and adapted to new processes without having to rebuild the infrastructure from scratch. The adoption of mobile robots moving within the company, dynamically managing internal logistics, has also facilitated this transition. These include Automated Guided Vehicles (AGVs), namely robots that follow fixed tracks on the floor to move, and Autonomous Mobile Robots (AMRs), which navigate autonomously using localization and mapping algorithms (SLAM: Simultaneous Localization and Mapping). This makes it possible to eliminate metal cages and, with the reclaimed space, design more compact factory layouts with higher production capacities. At the same time, a human operator can work in close contact with the robot. The human can perform operations that would be very difficult for the robot, while the robot can handle tasks that would be extremely tiring for a human. For example, in the manufacturing of an electronic device, the collaborative robot can drive micro-screws, which for a human would be tedious and draining; meanwhile, the human verifies that chips are placed correctly and that ribbon cables follow the correct routing (operations that would be very difficult for a robot). Certainly, for large-scale production of certain products, hard automation remains the best solution, but in more dynamic industries, being able to quickly move and reprogram robots is far more effective.

\begin{figure}[htbp]
  \centering
  \includegraphics[width=0.8\textwidth]{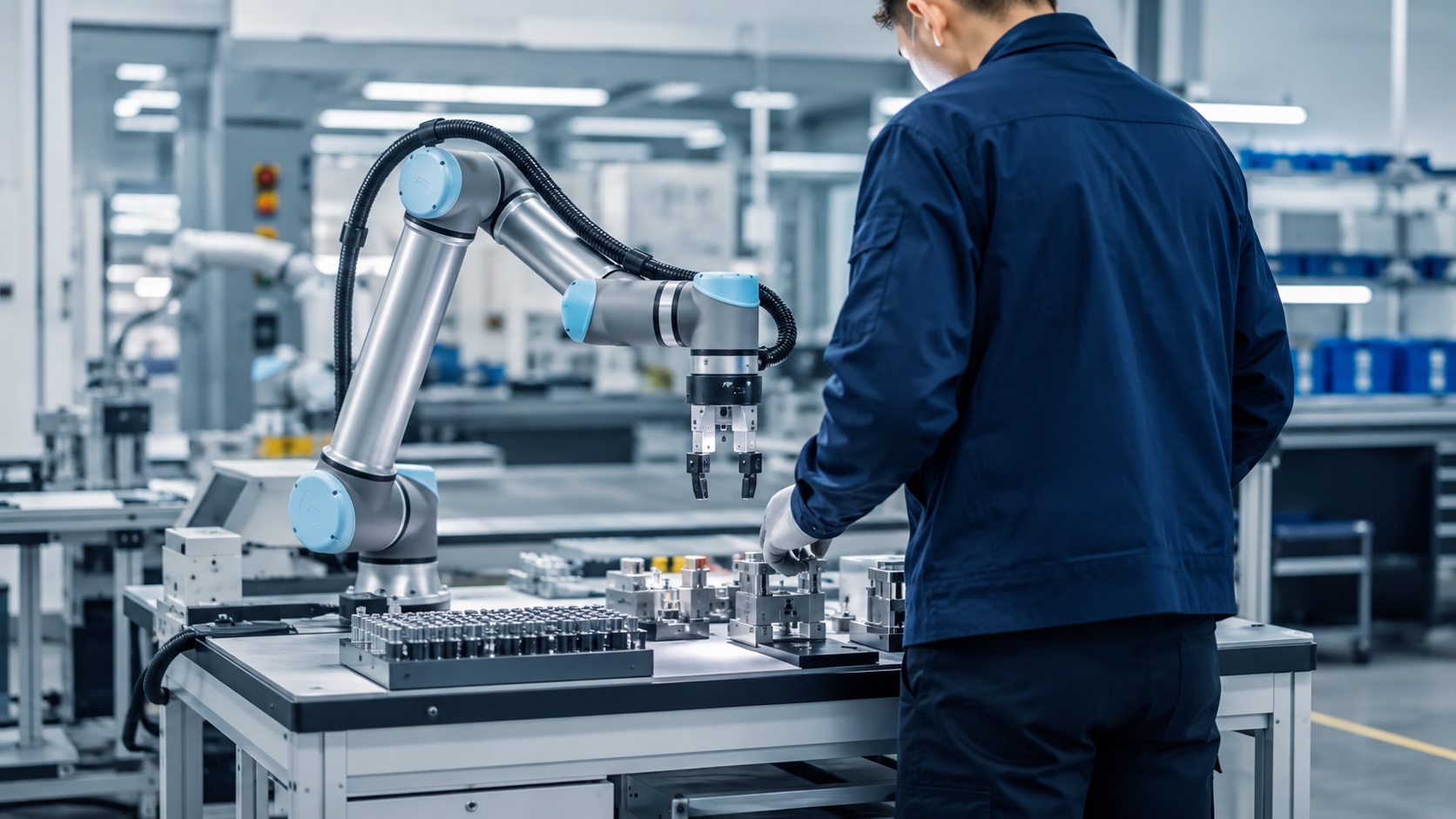}
  \caption[Cobot with human]{Example of a collaborative robot (cobot) integrated into a shared-workspace workstation with a human operator. Adapted from\cite{sito_cobot}.}
  \label{fig:Ur_robot_collaborativo}
\end{figure}

\subsection{Human-Robot Collaboration (HRC)}
At the core of Industry 4.0 lies Human-Robot Collaboration, which relies on combining the best qualities of both parties. Humans bring flexibility, intuition, and decision-making capabilities when facing unexpected events, whilst cobots ensure millimetric precision, strength, and consistency in repetitive and arduous tasks. From their combination emerges an efficient, safe, and high-value-added production process. To ensure an interaction that is both safe and efficient, the pressing need arises to implement a bidirectional communication interface. The critical bottleneck of this integration lies in mutual intentional transparency: on the one hand, the robot must predict and understand human actions and intentions (via intent recognition models and motion tracking) to proactively adapt its trajectory; on the other hand, the operator must be able to clearly anticipate the robot's future movements to avoid collisions, reduce cognitive load, and build a relationship of trust. In this scenario, Augmented Reality (AR) emerges as one of the most promising interfaces.

\section{Augmented reality and robotics: integration and communication limits}

\subsection{Augmented reality (AR)}
\begin{figure}[htbp]
    \centering
    \begin{subfigure}[b]{0.28\textwidth}
        \centering
        \includegraphics[width=\textwidth]{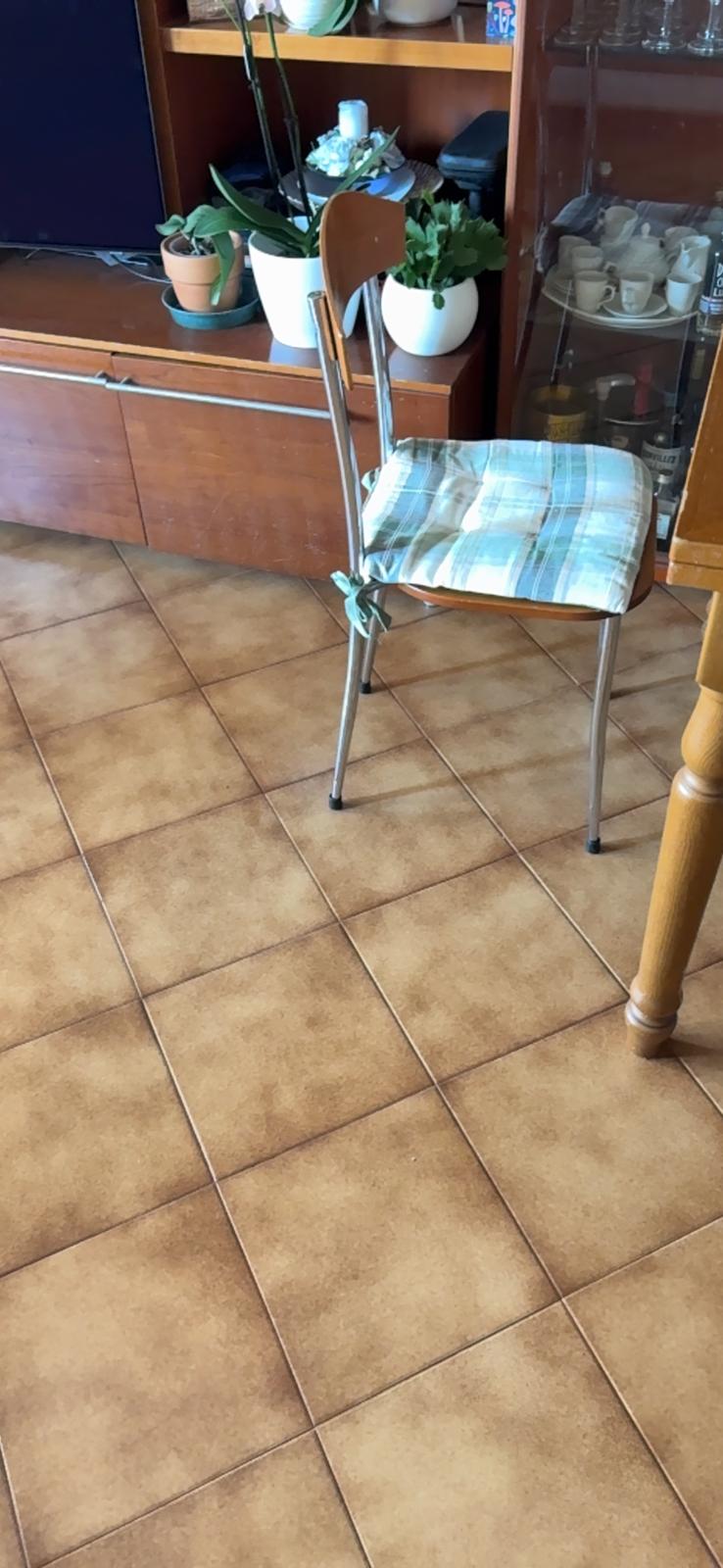}
        \caption{Real image without AR}
        \label{fig:ar1}
    \end{subfigure}
    \hfill
    \begin{subfigure}[b]{0.28\textwidth}
        \centering
        \includegraphics[width=\textwidth]{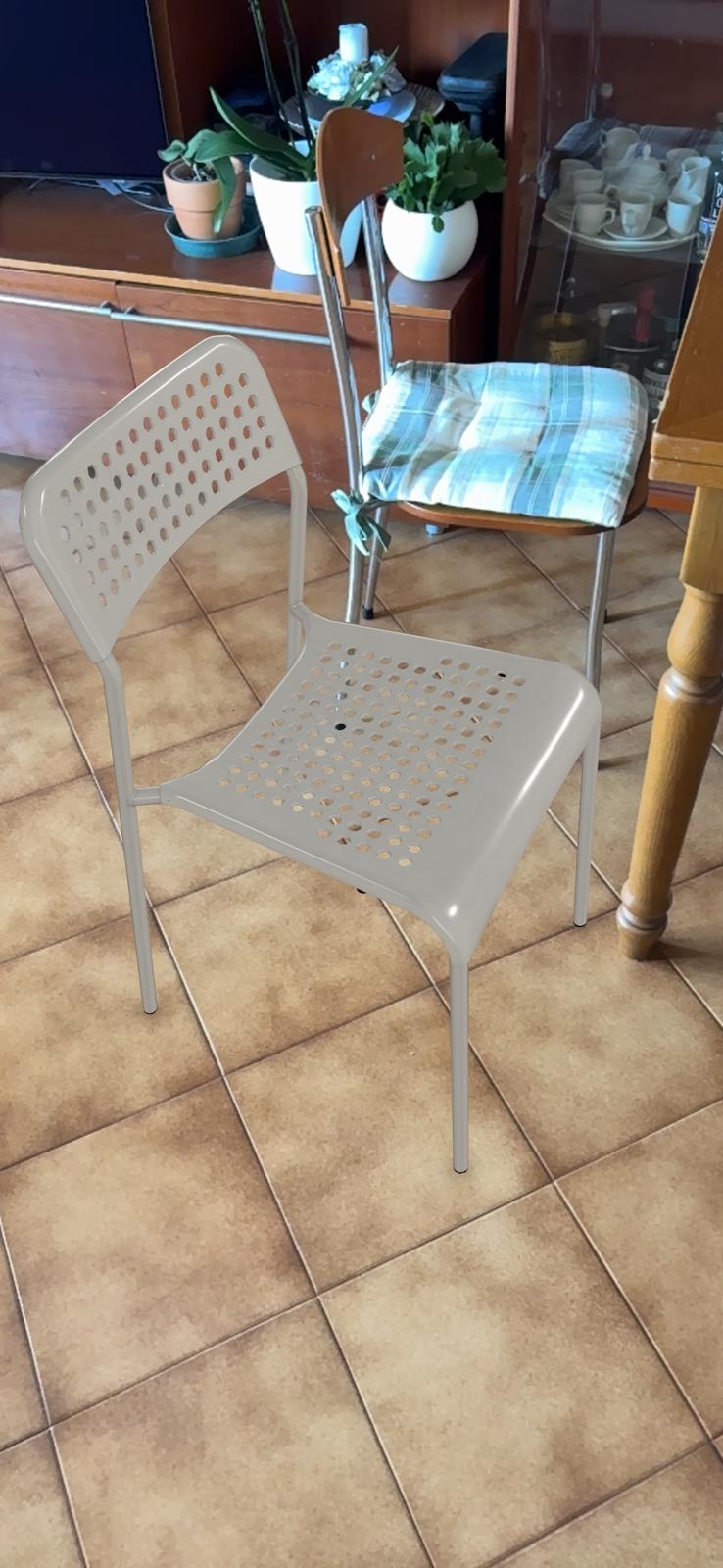}
        \caption{AR first angle}
        \label{fig:ar2}
    \end{subfigure}
    \hfill 
    \begin{subfigure}[b]{0.28\textwidth}
        \centering
        \includegraphics[width=\textwidth]{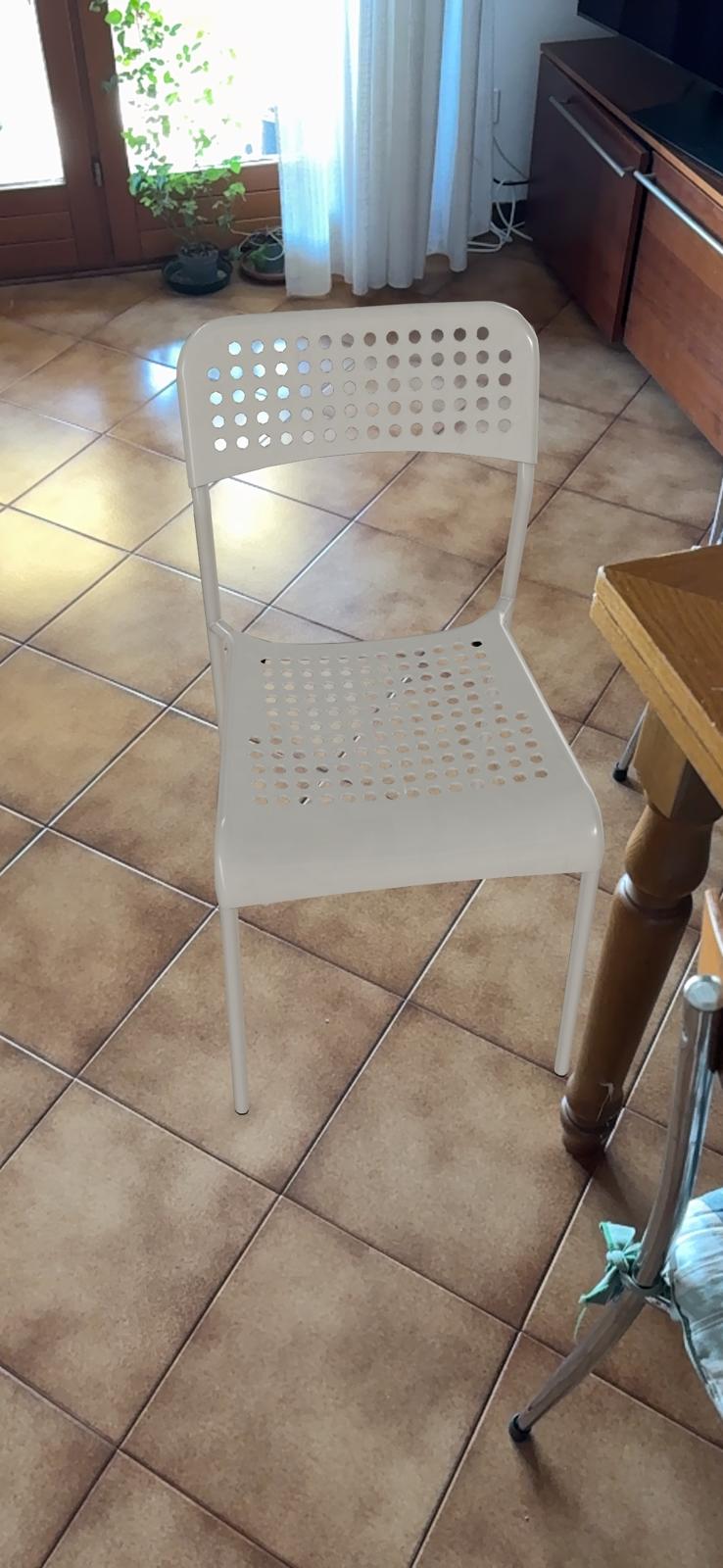}
        \caption{AR second angle}
        \label{fig:ar3}
    \end{subfigure}

    \caption[Examples of Augmented Reality]{Superimposition of 3D elements via AR through the IKEA website (adapted from \cite{sito_IKEA}, author's screenshot).}
    \label{fig:esempio_AR}
\end{figure}

Augmented Reality (AR) is a technology that enhances the perception of the outside world. Through dedicated devices (such as headsets or smartphones), digital elements like 3D images and holograms are reproduced, superimposing themselves onto the surrounding physical environment.
Unlike Virtual Reality (VR), which entirely replaces the real environment with a synthetic simulation that isolates the user, Augmented Reality integrates and enriches their visual or sensory field with different types of sensory elements (graphical, auditory, or haptic). A practical example of an everyday application is provided by the IKEA website\cite{sito_IKEA}, which allows users to view its products directly inside their own homes via the AR mode on their smartphones (Figure~\ref{fig:esempio_AR}).

\subsection{AR as a interface whith robotics}
The integration of Augmented Reality (AR) into collaborative robotics redefines human-machine interaction, transforming a headset into a true bidirectional bridge and advanced sensory system. Traditionally, spatial perception in collaborative robotics workstations relies on complex and expensive environmental systems consisting of numerous fixed cameras installed along the working perimeter, which allow the robot to know where the operator is located and what they are doing. Conversely, an AR headset worn by the operator drastically simplifies this architecture: the cameras and depth sensors integrated into the device continuously track head pose, hands, and the surrounding environment in real time. In this way, the headset itself ``sees'' the space on behalf of the robot, reconstructing the 3D map of the dynamic context directly from the user's perspective. Furthermore, the robot is not limited to merely receiving commands or trajectory parameters, but also transmits data regarding its progress status and motion intentions, which are projected directly into the operator's field of view, ensuring a continuous, shared information flow free of cumbersome external infrastructure.

\subsection{The lack of native communication}
Although high-end AR headsets on the market, such as the Magic Leap 2 used in this thesis project, offer advanced spatial tracking and environmental perception capabilities, these devices do not feature native integration with standard robotic ecosystems such as ROS 2 (Robot Operating System).
To bridge this architectural gap, the implementation of an intermediate communication middleware becomes necessary. This software ``bridge'' is responsible for extracting the data collected by the headset from the AR devices in real time, primarily video streams, and then converting and transmitting it to the robotic system. In this way, the robot can use this data to optimise trajectory planning, improve localization, or execute complex manipulation and human-robot collaboration tasks.

\section{Objectives}
\subsection{The developed framework}
The main objective of this thesis project, carried out at the IAS-Lab of the University of Padua, is the design and development of a software framework capable of bridging the communication gap between the high-end Augmented Reality headset Magic Leap 2 and the robotic ecosystem ROS 2 (Robot Operating System).
To overcome the lack of native integration, a modular architecture was developed based on network cooperation between the wearable device and a computer (which hypothetically handles managing other equipment such as robotic arms or cameras) running ROS 2. Communication between the two environments can take place either wirelessly or via a wired connection, ensuring in the latter case maximum bandwidth stability and extremely low latency values for the most critical applications.
The operation of the framework relies on dynamic and optimised management of sensory resources:
\begin{itemize}
    \item Dynamic configuration from ROS 2: A dedicated ROS 2 node resides on the computer, reading a configuration file in \texttt{.yaml} format. Within this file, operational communication parameters are defined, such as the selection of sensors to enable (e.g., cameras, inertial data, depth sensors) and the required resolution of the data streams.
    \item Starting the stream on Magic Leap 2: When the application on the headset starts, the latter receives the configuration sent by the computer and adapts its processing accordingly, initiating the streaming of only the requested data. This approach avoids saturating network bandwidth and prevents unnecessary computational load on the headset for unused sensors.
\end{itemize}
The work carried out has led to the implementation of a complete and ready-to-use application, conceived from the outset with a highly scalable architecture and designed to be easily extended with new functionalities or additional data processing modules.
Finally, with a view to fostering scientific reproducibility and encouraging open-source collaboration, the entire project has been released in a public repository on \texttt{GitHub}~\cite{MagicLeap2_ROS2_Repo}. In this way, the scientific community and developers will be able to replicate the proposed infrastructure, test it with other robotic systems, or enhance it by integrating new features.

\subsection{Validation}
In order to assess the operational correctness of the developed framework and the quality of the data streams transmitted in real time, an experimental validation phase was conducted. In particular, the aim was to verify the actual usability of the sensory data within a typical robotic localization and mapping algorithm. For this purpose, the video streams generated by the Magic Leap 2 world cameras and published to ROS 2 via the framework were fed as input to ORB-SLAM3~\cite{ORBSLAM3_TRO}, one of the benchmark SLAM (Simultaneous Localization and Mapping) systems in the literature, used on this occasion in monocular mode. To evaluate the algorithm's performance, the trajectory estimated by ORB-SLAM3 was compared against the ground truth provided by the poses and extrinsics internally estimated by the headset, originating from the native spatial tracking of the Magic Leap 2, which is renowned for its high accuracy and precision. The experimentation was conducted by running several tests along short-range trajectories in an indoor environment. The results of the comparison between the trajectory reconstructed in monocular mode and the device's native estimation confirmed proper data routing, reduced transmission latency, and full compatibility of the developed infrastructure with state estimation algorithms commonly employed in robotics.

\clearpage

\section{Thesis outline}
The remainder of this thesis is structured progressively across the following chapters. Chapter~2 surveys the state of the art, investigating the role of Augmented Reality in human-robot collaboration, egocentric vision paradigms for wearable systems and exoskeletons, and contemporary sensory streaming middleware, thereby contextualising the technological gap that the Magic Leap 2 headset aims to bridge. Chapter~3 introduces the foundational technologies leveraged throughout the project, detailing the hardware architecture and onboard sensor suite of the Magic Leap 2, the core tenets of ROS 2, the Unity development environment, the \texttt{ROS-TCP-Connector}, and the operational principles of the \texttt{ORB-SLAM3} visual localisation pipeline. Chapter~4 elaborates on the engineering and implementation of the framework, detailing requirement specifications, the dynamic parameter negotiation mechanism governed by configuration files, the characterisation of individual sensory streams, and operational guidelines for system deployment and code modularity. Chapter~5 presents the experimental validation of the middleware, detailing the trajectory evaluation methodology using the \texttt{evo} package and benchmarking monocular SLAM performance against native headset pose estimates across multiple real-world trajectories. Finally, Chapter~6 summarises the primary achievements, draws concluding remarks, and outlines promising future research directions in bidirectional interaction and advanced collaborative robotics.
    \cleardoublepage

    \chapter{State of the art}

\section{Augmented reality in human-robot collaboration (HRC)}
In recent years, the transition towards the Industry 4.0 paradigm has made direct cooperation between human and machine a key requirement to increase the flexibility and efficiency of production lines. In this scenario, operator safety and the fluidity of interaction represent the two primary challenges. Traditionally, human protection in Human-Robot Collaboration (HRC) contexts is managed through virtual fences, perimeter sensors, or monitored stop mechanisms compliant with ISO/TS 15066 standards. However, the passive stopping of the robot in the presence of the operator heavily penalises overall productivity.
Augmented Reality (AR) has established itself as the most promising technology to overcome this dichotomy, providing intuitive and bidirectional communication channels. The use of AR interfaces enables real-time visualization within the operator's field of view of the robot's internal state, planned trajectories, dynamic safety volumes, and step-by-step procedural instructions.
A relevant study in the industrial domain is that conducted by Hietanen et al.~\cite{Hietanen2020ARInteraction} (2020), in which an interactive AR-based interface was developed (comparing projection systems and wearable headsets such as HoloLens) for dynamic workspace division during the assembly of diesel engine components. The research demonstrated that the visual sharing of safety zones and manipulator intentions reduces robot idle time by over 50\% and overall execution times by 20--24\% compared to non-collaborative workstations.
In parallel, AR has evolved from a mere passive visualization tool into an active means of control and teaching (Learning from Demonstration, LfD). In this context, Yan et al.~\cite{Yan2024Complementary} (2024) proposed a complementary framework for HRC combining an AR headset (HoloLens 2) and a haptic interface to control a 7-degree-of-freedom manipulator (Franka Emika). The operator can intervene in the robot's null space via aerial gestures tracked by the headset, modifying the arm posture to avoid unexpected collisions without interrupting the primary task of the end-effector, and teaching new parametric trajectories modelled using Dynamic Movement Primitives (DMPs).

\section{Egocentric vision and perception for wearable systems and exoskeletons}
Beyond stationary industrial robots, Augmented Reality devices are finding increasing application in mobile and wearable robotic systems, particularly in lower-limb exoskeletons (Lower-Limb Exoskeletons, LLEs). Traditional assistive exoskeletons base their operation on predefined walking patterns derived from standard biomechanics, resulting in rigid behaviour poorly suited to unstructured environments.
To ensure genuine adaptability to the terrain, it is essential to equip these systems with autonomous environmental perception capabilities (Environment-Adaptive Gait Planning, EAGP). In this perspective, Trombin et al.~\cite{Trombin2024EAGP} (2024) introduced a collision-free trajectory generator (CFFTG) capable of adapting step height and length in real time based on the analysis of 3D point clouds acquired by RGB-D cameras.
However, sensors mounted solely at pelvis or foot level exhibit intrinsic limitations: they feature a restricted downward field of view and detect obstacles only at very short range, hindering forward path planning. To overcome this constraint, Mihailovic et al.~\cite{Mihailovic2025Egocentric} (2025) proposed an egocentric vision module integrating an AR headset camera (head-mounted) with the exoskeleton camera. Exploiting visual-inertial SLAM algorithms (ORB-SLAM3) executed in parallel on the streams, alongside 3D geometric matching techniques (DUSt3R), the system merges the maps generated from both perspectives. This configuration extends the perceptual horizon of the user, enables early recognition of stairs, ramps, and obstacles, and simultaneously projects visual feedback holograms (such as safe footholds) directly onto the headset's optical display.

\section{Sensor acquisition and streaming middleware for AR headsets}
For an AR headset to operate effectively as a sensory node within complex robotic control architectures, raw data from on-board sensors must be extracted at high frequencies and with minimal latency towards external workstations.
In the early years of commercial headset adoption, access to low-level hardware was strictly limited by proprietary operating systems. A pivotal breakthrough for the scientific community occurred with the introduction of ``Research Mode'' for the Microsoft HoloLens 2, documented by Ungureanu et al.~\cite{Ungureanu2020ResearchMode} (2020). Research Mode provided, for the first time, a set of low-level C++ APIs to directly access streams from the four infrared monochrome tracking cameras (Visible Light Cameras, VLC), the Time-of-Flight (ToF) depth sensor in both short-range and long-range modes, and raw inertial measurement unit (IMU) samples.
Leveraging these APIs, Dibene and Dunn~\cite{Dibene2022hl2ss} (2022) developed hl2ss (HoloLens 2 Sensor Streaming), a TCP socket-based server application for the Universal Windows Platform (UWP) capable of streaming in real time towards Python clients all visual feeds, spatial poses extrapolated from head tracking, and eye and hand tracking data. This work demonstrated the effectiveness of local network streaming for executing computationally demanding tasks (such as panoptic segmentation or dense TSDF 3D reconstruction) on dedicated workstations.

\section{The technological gap and the role of Magic Leap 2}
Despite the success and widespread adoption of Microsoft HoloLens 2 in the scientific literature of the past four years, the technological landscape has undergone a profound transformation: the HoloLens 2 device has reached the end of its commercial life cycle and has been discontinued, leaving the research community without a long-term supported benchmark platform based on Windows UWP.
The Magic Leap 2 headset represents the natural benchmark technological successor for enterprise- and industrial-grade augmented reality:
\begin{itemize}
    \item Advanced Perception Hardware: It integrates three wide-FOV, high-resolution monochrome tracking cameras ($1016 \times 1016$ at 30 Hz), a central RGB camera up to 4K, a metric ToF depth sensor, four IR cameras for Eye Tracking, and three distinct inertial measurement units.
    \item Open Software Architecture: Based on the Magic Leap OS operating system (derived from Android AOSP) and compliant with the OpenXR industry standard.
\end{itemize}
However, unlike the HoloLens landscape, within the Magic Leap 2 ecosystem there was no open-source tool dedicated to the centralised, modular, and synchronised streaming of raw sensors towards the standard robotic middleware ROS 2.
This thesis project aims precisely at bridging this technological gap: by designing a modular bridge based on Unity and native Android plug-ins, capable of dynamically negotiating transmission parameters from ROS 2 and validating the temporal and spatial integrity of video streams via real-time visual localization pipelines.

    \cleardoublepage

    \chapter{Technologies used}
In this thesis project, a framework was developed to enable the real-time streaming of data gathered by the Magic Leap 2 AR headset towards a computer equipped with ROS 2. The system was implemented as an application within the Unity environment to handle the graphical interface and sensory stream acquisition on board the headset. Communication between Unity and ROS 2 relies on the ROS-TCP-Connector package, whilst the quality and synchronization of the transmitted data were verified via the ORB-SLAM3 localization algorithm. This chapter describes the main technological tools employed, highlighting their relationships and the role each plays within the project.

\section{Magic Leap 2}
\begin{figure}[htbp]
  \centering
  \includegraphics[width=0.7\textwidth]{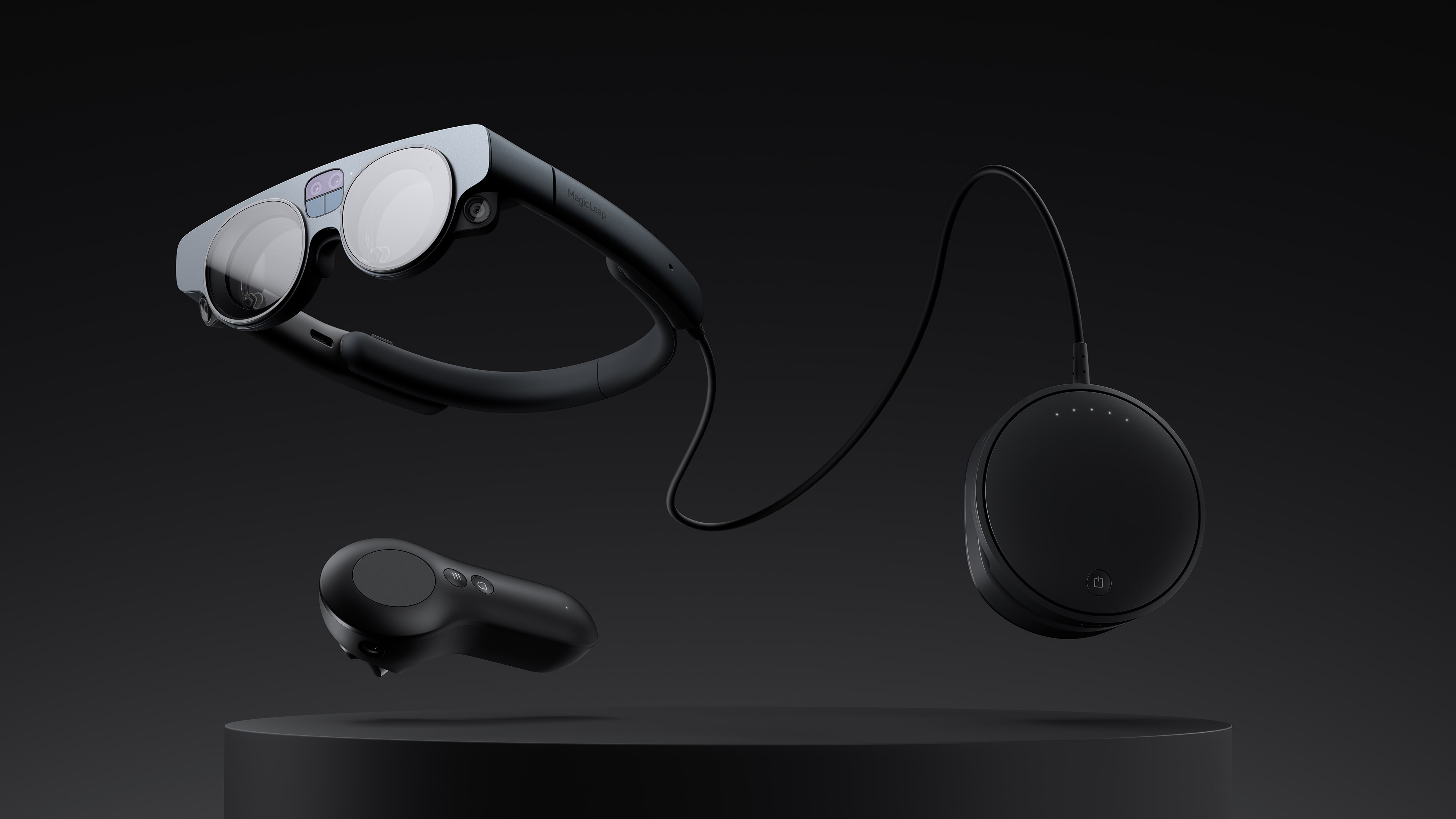}
  \caption[Magic Leap 2]{Magic Leap 2 AR headset. Adapted from \cite{sito_foto_MagicLeap2}.}
  \label{fig:MagicLeap2}
\end{figure}

In this thesis project, the Magic Leap 2 was used, an advanced augmented reality headset designed by the American company Magic Leap and primarily targeted at the enterprise and professional market (such as surgery, engineering, manufacturing, and defence).
Across the landscape of headsets, it is necessary to differentiate the Magic Leap 2 from more popular devices by outlining three types of approaches to augmented reality:
\begin{itemize}
    \item VR headsets with Video Passthrough (Apple Vision Pro, Meta Quest, HTC Vive XR Elite, Pimax Crystal)
    \item Traditional Smart Glasses (Ray-Ban Meta Display)
    \item True AR headsets with \textit{Optical See-Through} (Magic Leap 2)
\end{itemize}
The first category of headsets includes devices conceived for virtual reality. When wearing the device, the user looks at display screens, and the view of the external world is reconstructed artificially by capturing the environment through cameras. This inevitably introduces processing latency (delay) and a lower visual resolution compared to direct human sight. This sensory discrepancy is the primary cause of visual fatigue and \textit{motion sickness} (virtual reality kinetosis), limiting prolonged use.
The second category of headsets lacks advanced sensors for three-dimensional environment mapping (SLAM); these devices are incapable of performing spatial tracking. Consequently, graphical elements remain superimposed on the field of view and follow head movements, making it impossible to display virtual objects anchored and stationary in space.
The Magic Leap 2 (Figure~\ref{fig:MagicLeap2}) enables the user to view the surrounding environment directly through transparent lenses (similar to sunglasses) and projects spatially localized 3D objects into the environment: moving around the room, the virtual elements remain perfectly anchored and stationary in their position.

\subsection{Hardware configuration and input devices}
The hardware of the Magic Leap 2 adopts a distributed architecture: the glasses frame houses neither the main processing unit nor the power supply pack directly. The execution of complex Simultaneous Localization and Mapping (SLAM) algorithms, computational graphics processing, and application management are entirely delegated to an external ``compute module'' connected via cable, which encases the high-performance computing unit and the battery (Figure~\ref{fig:MagicLeap2}). To ensure precise, smooth, and ergonomic interaction within the augmented reality environment, the system also includes a dedicated controller featuring advanced tracking, designed to optimise the user experience and the manipulation of virtual elements (Figure~\ref{fig:MagicLeap2}).
The flagship feature of the company Magic Leap is its diffractive waveguide lenses (\textit{Waveguides}). These lenses contain several layers of ultra-thin glass onto which microstructures are etched via nanolithography. A micro-projector (LCOS optical engines) positioned on the frame projects digital images towards the edge of the lens. The projected light enters the waveguide and travels internally until it reaches diffractive gratings printed on the lenses, which bend the light towards the user's eye, superimposing the holograms onto the view of the real world \parencite{sito_MagicLeap2_waveguide_engineering}.
Furthermore, the lenses of this headset feature a layer dedicated to \textit{Dynamic Dimming}. This is a liquid crystal layer capable of controlling light transmission, which can dim the entire environment or create a black ``shadow mask'' directly behind the holograms. As a result, holograms do not appear semi-transparent even when positioned in brightly lit rooms or under direct sunlight.

\begin{figure}[htbp]
  \centering
  \includegraphics[width=0.7\textwidth]{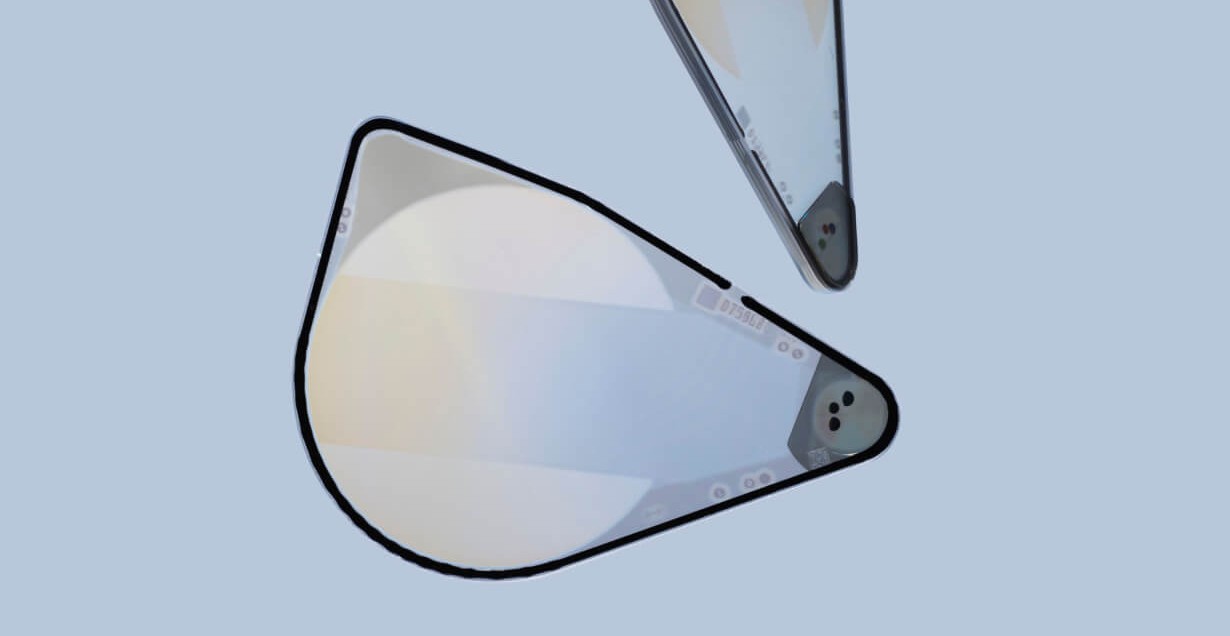}
  \caption[Magic Leap 2 lenses]{Magic Leap 2 diffractive waveguide lenses. Adapted from \cite{sito_MagicLeap2_waveguide}.}
  \label{fig:waveguide}
\end{figure}

\subsection{Sensors}
\begin{figure}[htbp]
  \centering
  \includegraphics[width=0.8\textwidth]{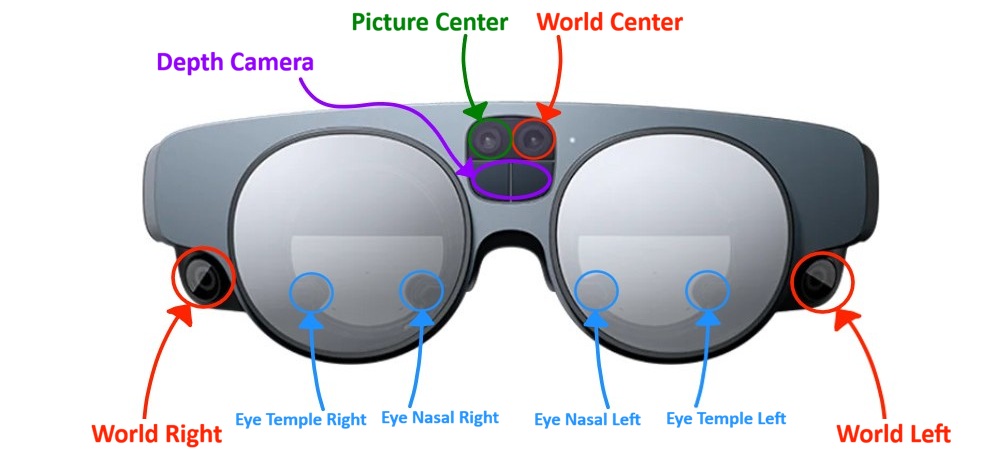}
  \caption[Frontal view of the Magic Leap 2]{Frontal view of the Magic Leap 2.}
  \label{fig:ML2_front}
\end{figure}

The Magic Leap 2 features several sensors that, when combined, allow the headset to localize itself in space with millimetric precision and, consequently, calculate in real time where to project objects onto the lenses so that they appear anchored to reality.
In particular, the sensory architecture of the device includes:
\begin{itemize}
    \item Inertial Measurement Units (IMUs): accelerometric and gyroscopic sensors to track head movement rotation and acceleration at very high frequencies. Specifically, there are two IMUs inside the headset and one in the compute module.
    \item Tracking cameras (World Cameras): three greyscale cameras with a resolution of 1016x1016 positioned centrally, on the right, and on the left of the headset (in red in Figure~\ref{fig:ML2_front}).
    \item RGB Camera (Picture): a high-resolution camera positioned on the front at the centre of the headset (in green in Figure~\ref{fig:ML2_front}).
    \item Depth sensor (Depth Camera): a ToF (Time of Flight) camera to accurately reconstruct the 3D geometry of surfaces and the distance of objects around the user. Positioned on the front and at the centre of the headset with a resolution of 544x480 (in purple in Figure~\ref{fig:ML2_front}).
    \item Eye Tracking cameras: cameras dedicated to the eyes to monitor the user's gaze, optimise rendering, and accurately calculate visual convergence. Featuring a resolution of 400x400, they are positioned two per eye: one near the nose and one near the temple (in light blue in Figure~\ref{fig:ML2_front}).
    \item Ambient and illumination sensors: to assess external lighting conditions and adjust lens brightness and dimming.
    \item Pressure sensors: to calculate the vertical displacements of the headset with greater precision.
\end{itemize}

\subsection{Operating system}
The Magic Leap 2 is equipped with Magic Leap OS, a proprietary operating system based on Android (\textit{AOSP - Android Open Source Project}). The choice of adopting an Android-derived architecture enables combining the stability of an extensively tested Linux kernel with a native C/C++ and Java development ecosystem, whilst simultaneously facilitating integration with graphics engines such as Unity. Unlike conventional mobile operating systems, Magic Leap OS has been heavily customised at the low level to handle the headset's complex distributed architecture. The operating system manages in real time the computational load balancing across processing units, optical calibration management (including dynamic dimming control), and the execution of localization algorithms, which are indispensable for augmented reality.

\clearpage

\subsection{Magic Leap SDK}
The SDK (Software Development Kit) for Magic Leap 2 is the set of tools, APIs, and libraries provided by Magic Leap to enable developers to build Augmented Reality (AR) applications aligned with industrial standards such as OpenXR (unlike its predecessor, the Magic Leap 1).
The developer kit provides three routes to develop applications on Magic Leap 2:
\begin{itemize}
    \item Unity (Recommended engine): This is the most extensively supported platform. The Magic Leap Unity SDK package integrates via the standard OpenXR and XR Plug-in Management system of Unity.
    \item Unreal Engine 5 (UE5): A Magic Leap Unreal SDK is available for Unreal Engine 5, leveraging OpenXR extensions.
    \item C/C++ Native SDK: For developers creating proprietary engines or requiring ultra-low-level performance and integrations.
\end{itemize}
The main features provided by the SDK are:
\begin{itemize}
    \item Spatial Mapping: Scans the 3D environment by identifying walls, floors, and furniture to handle physical collisions, detect planar surfaces, and position virtual objects permanently (Spatial Anchors).
    \item Body Tracking (Eye and Hand Tracking): Monitors gaze, pupils, and advanced hand movements (joints and gestures) to enable direct interaction without external devices.
    \item 6-DoF Controller: Offers precise 6-degree-of-freedom tracking by integrating optical and electromagnetic sensors to eliminate blind spots.
    \item Spatial Audio: Simulates 3D sound provenance and propagation for an immersive acoustic experience consistent with the surrounding space.
    \item Sensor Access: Provides developers with access to data gathered by the hardware (cameras, depth sensors, IMUs, microphones, and ambient light) for advanced analysis or custom algorithms.
\end{itemize}
Magic Leap provides the official documentation for using the SDK~\cite{MagicLeap2_Developer_Docs}.

\clearpage

\section{ROS 2 (Robot Operating System 2)}
ROS 2 (Robot Operating System 2) is the de facto standard open-source middleware framework for designing and developing professional and industrial robotic software. Despite its name, it is not a conventional operating system, but rather a set of libraries, tools, and conventions that allow distinct software components to communicate with each other, managing both individual sensors/actuators and entire robot fleets.

\subsection{Modular architecture}
ROS 2 is based on a modular software architecture. 
The standard underpinning communication in ROS 2 is DDS (\textit{Data Distribution Service}), an open, data-centric industrial standard developed by the Object Management Group (OMG) for real-time peer-to-peer communication. Unlike its predecessor ROS 1, which relied on a central coordinating node (ROS Master) and custom internal protocols, DDS ensures a fully distributed and decentralised architecture.
The fundamental computational unit is the node, an independent process dedicated to a single task (e.g., sensor reading, actuator control).
Nodes can discover each other automatically over the network and exchange data with a high degree of reliability and determinism, also enabling the definition of advanced Quality of Service (\textit{Quality of Service - QoS}) policies to configure priorities, fault tolerance, durability, and bandwidth management across communication flows.
To exchange data, nodes rely on specific communication paradigms: topics, which are asynchronous publish-subscribe (many-to-many) channels ideal for continuous streams of information such as sensory feeds. The framework also provides two other types of communication mechanisms: services and actions. Services operate on a request-response basis: a node requests an operation and waits for an immediate response before proceeding. This is suitable for short, instantaneous tasks, such as switching on an LED, resetting a sensor, or querying the current battery charge level. Actions are intended for long-running, complex tasks, such as navigating a robot from one point to another. Unlike services, during execution an action provides real-time feedback on the progress of the operation and allows the task to be cancelled at any time prior to returning the final result.
\clearpage
\subsection{Coordinate frames and transforms}
To track robot positions and motions over time, ROS 2 utilises the TF (Transform) infrastructure. A TF represents the position and orientation of an element with respect to another (Figure 3.4), mathematically corresponding to a rigid affine transformation in $SE(3)$. It defines the relationship between two coordinate frames (frames) by means of:
\begin{itemize}
    \item a three-dimensional vector $\mathbf{t} = [x, y, z]^T$ for translation;
    \item a unit quaternion $\mathbf{q} = [q_x, q_y, q_z, q_w]^T$ for 3D rotation.
\end{itemize}
The transformations are organised in a TF tree, a tree structure whose nodes are the coordinate frames (frames) and whose edges are the corresponding transforms. This ensures a unique path between any two frames, enabling ROS 2 to compute composite transformations and convert 3D coordinates instantaneously.
\begin{figure}[htbp]
  \centering
  \includegraphics[width=0.5\textwidth]{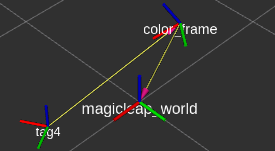}
  \caption[Example of Transform.]{Representation of the transforms (TF) of the Magic Leap 2 headset cameras with respect to the Magic Leap coordinate frame (world frame) and with respect to a visual marker designated tag4. Screenshot of data visualised via RViz.}
  \label{fig:Transform_esempio}
\end{figure}

\subsection{Tools}
ROS 2 provides several essential tools for development and debugging, including:

\begin{itemize}
    \item \texttt{RViz 2}: a 3D visualization tool that allows real-time monitoring of data published on topics (e.g., TFs, point clouds, images).
    \item \texttt{rosbag2}: a logging and playback (replay) tool for data streams exchanged between nodes, useful for analyzing tests even in the absence of the physical hardware.
\end{itemize}
 
\section{Unity}
Unity is a real-time cross-platform 3D graphics engine (game engine) and development environment, widely employed for creating interactive content and Augmented Reality (AR) and Virtual Reality (VR) applications. Within this thesis project, the Unity 2022.3.62f3 LTS (Long Term Support) version was used. Unity was not employed as a simulator, but as the primary runtime and execution environment on board the Magic Leap 2, responsible for the graphical rendering of holographic elements and the management of the application that acquires the headset's sensory data.

\subsection{Component-based architecture and execution loop}
Unity is structured around a component-based paradigm relying on a hierarchy of objects:
\begin{itemize}
    \item \texttt{GameObject} and Components: Any element present in the virtual environment is a GameObject. The behaviour and characteristics of the object are defined by attaching modular components to it (such as virtual cameras, spatial Transform components, or colliders). Custom application logic is implemented via C\# scripts that extend the base \texttt{MonoBehaviour} class.
    \item Main Thread and Frame Rate Optimisation: The Unity runtime processes game logic, device API interactions, and rendering on the Main Thread. In optical augmented reality applications, maintaining a high and steady frame rate (typically between 60 and 120 fps) is critical to prevent visual discrepancies between the real world and overlaid holograms; consequently, sensory buffer processing must be handled in a non-blocking manner.
\end{itemize}
Code execution follows a deterministic, event-driven lifecycle paced by the graphics engine rendering. Among the primary methods utilised for interfacing and data stream management are:
\begin{itemize}
    \item \texttt{Awake()} and \texttt{Start()}: Initialisation methods invoked upon object creation, employed to allocate resources in memory and initiate sensor capture services.
    \item \texttt{Update()}: Invoked at each rendering cycle (once per frame) at a variable frequency depending on the computational load; it is responsible for updating component states and handling real-time events.
    \item \texttt{FixedUpdate()}: Executed at regular, discrete time intervals (fixed $\Delta t$), independently of the rendering frame rate, employed for physics and operations requiring constant temporal periodicity.
\end{itemize}

\section{ROS-TCP-Connector and ROS-TCP-Endpoint}
To enable bidirectional message exchange between the Unity application running on board the Magic Leap 2 and the active ROS 2 nodes on the computing workstation, the open-source framework ROS-TCP-Connector~\cite{ROS_TCP_Connector}, developed within the Unity Robotics Hub project, was employed.
This tool acts as an inter-process communication bridge based on the TCP/IP transport protocol, decoupling the C\# graphics environment from the underlying robotic middleware.

\subsection{Client-server architecture}
Communication is structured according to a classic client-server network model consisting of two distinct entities:
\begin{itemize}
    \item \texttt{ROS-TCP-Connector} (Unity side - Client): A package integrated directly within the Unity environment in the form of a C\# library. It operates as a TCP client: once the remote machine's IP address and listening port (typically the standard port 10000) are configured, the connector establishes a persistent socket connection towards the workstation.
    \item \texttt{ROS-TCP-Endpoint} (ROS 2 side - Server): A ROS 2 node (developed in Python) running on the host machine. It acts as a TCP server listening on a socket: it accepts the incoming connection from the headset, receives the byte streams, deserialises them, and takes care of instantiating the corresponding publishers and subscribers within the ROS 2 \textit{computational graph}.
\end{itemize}
In this architecture, the network topology solely requires that the AR headset and the PC running ROS 2 be reachable within the same subnet (e.g., via a local Wi-Fi connection), making deployment flexible and free from complex hardware constraints.

\subsection{Serialization and message generation}
To enable data exchange between Unity and ROS 2, information must be structured according to the standard ROS 2 message format (defined in \texttt{.msg} files, such as \texttt{sensor\_msgs/Image} for visual frames or \texttt{geometry\_msgs/PoseStamped} for spatial coordinates).
The ROS-TCP-Connector package incorporates an automatic generation tool (ROS Message Generation) that converts ROS message definitions into corresponding C\# classes usable within the Unity environment. The general workflow of the communication pipeline comprises the following steps:
\begin{itemize}
    \item Encapsulation and serialization: Within Unity, application data is populated directly into the generated C\# class instances. These data structures are subsequently serialized into a binary byte stream.
    \item Transmission via socket: The TCP client embedded in Unity transmits the data packet over the network connection towards the host machine.
    \item Reception and publishing: The ROS-TCP-Endpoint server node running on ROS 2 receives the bytes from the socket, deserializes them, and handles actively publishing the message onto the corresponding topic of the computational graph. The same mechanism operates symmetrically whenever Unity needs to subscribe to topics published by ROS 2 nodes.
\end{itemize}

\section{ORB-SLAM3}

\subsection{SLAM algorithms (Simultaneous Localization and Mapping)}
The acronym SLAM (Simultaneous Localization and Mapping) identifies a class of algorithms and computational techniques that enable an autonomous device (such as a mobile robot or an augmented reality headset) to construct a map of an unknown environment whilst simultaneously estimating its own position and orientation (6-DoF pose) within that map in real time.
In Augmented Reality and Spatial Computing systems, SLAM represents the core enabling technology: without it, the headset could neither understand the geometry of the surrounding environment nor track user movements, making it impossible to stably anchor digital elements in physical space.
From an operational perspective, the algorithm processes in real time the continuous stream of raw data originating from on-board sensors (such as cameras, stereoscopic cameras, depth sensors, or IMU inertial units). In an initial visual processing stage (front-end), the system identifies and extracts salient reference points from the image (termed features or keypoints) and matches them across consecutive frames, estimating the relative motion instant by instant via visual odometry techniques. However, this local estimate is inevitably subject to an incremental accumulation of errors over time, known as ``drift''. To ensure global consistency in the reconstruction, the optimisation stage (back-end) intervenes, modelling the spatial relationships between camera poses and three-dimensional environment landmarks by minimising the overall reprojection error via non-linear optimisation techniques (\textit{Bundle Adjustment}). The process is completed by place recognition and loop closure: whenever the device revisits an area explored earlier, the system recognises the visual match, corrects the geometric drift accumulated along the entire trajectory, and coherently aligns the global map.

\subsection{ORB-SLAM3}
Among computer vision-based SLAM architectures, ORB-SLAM3~\cite{ORBSLAM3_TRO} represents one of the most advanced and comprehensive open-source state-of-the-art systems. It is a versatile library capable of performing visual, visual-inertial, and multi-map SLAM in real time on standard CPUs, ensuring high performance without the strict requirement for dedicated hardware acceleration. 
The strength of the architecture lies in the uniform use of ORB features (\textit{Oriented FAST and Rotated BRIEF}) across all system modules, from rapid frame-to-frame tracking to the place recognition stage. Features are distinct visual details (such as corners or edges) that the camera uses as reference points. The ORB method detects these points in the image extremely quickly and creates a binary descriptor for each, namely a compact sequence of bits encoding its surrounding appearance, which keeps recognition computationally lightweight. In this way, the computer can compare hundreds of points between frames almost instantaneously, recognising them even when the camera translates or rotates. This choice ensures excellent invariance to viewpoint changes and illumination variations whilst maintaining an exceptionally low computational overhead. 
Compared to prior iterations of the library, ORB-SLAM3 introduces tight integration with inertial sensors, enabling robust tracking even in challenging scenarios characterised by rapid camera motions or temporary visual occlusions.
A further key innovation is the Atlas system, a multi-map management engine that allows the system to operate seamlessly even during tracking loss. When visual tracking fails, ORB-SLAM3 does not abort execution but instead initialises a new local sub-map; as soon as the device revisits a previously mapped area, the algorithm seamlessly merges the disconnected sub-maps into a single coherent global model. Owing to this combination of robustness, metric accuracy, and versatility across diverse sensor configurations (monocular, stereo, and RGB-D), ORB-SLAM3 stands as one of the most widely adopted SLAM algorithms.
    \cleardoublepage

    \chapter{Framework design and implementation}

\section{Functional requirements and constraints}
The design of the communication architecture between the Magic Leap 2 headset and the ROS 2 ecosystem required an attentive preliminary analysis of both the application goals and the hardware and computational limitations typical of embedded devices and wireless transmission. Below, the identified functional requirements and the constraints that guided the software development are formalised.
\subsection{Functional requirements}
The functional requirements define the operational capabilities and services that the software bridge developed on board the headset must provide:
\begin{itemize}
    \item Multi-sensory acquisition: The system must interface with the headset hardware to extract heterogeneous data streams. Specifically, it must be capable of acquiring video streams (RGB and ambient tracking cameras), the 3D spatial pose estimate computed by the headset, and readings from auxiliary sensors (such as IMUs and the ambient light sensor).
    \item Data serialization and transmission over ROS 2: The extracted sensory information must be packetised and serialised in compliance with standard ROS 2 messages (in particular the \texttt{sensor\_msgs} suite and the corresponding types for images, poses, and raw readings). For each video stream, the bridge must ensure the synchronised publishing of the corresponding intrinsic calibration message (CameraInfo), enabling the proper interpretation of images by receiving nodes.
    \item Dynamic runtime configuration: To maximise operational flexibility and avoid repeated recompilation cycles of the Unity/Android package on the headset, the system must allow the configuration of streaming parameters (such as which sensors to enable, target resolutions, and framerates) directly from the ROS 2 side via a \texttt{config.yaml} configuration file parsed upon application start-up.
\end{itemize}

\subsection{Project and system constraints}
Architectural and technological constraints guided the implementation choices to guarantee maximum compatibility, efficiency, and scalability:
\begin{itemize}
    \item Standardisation and interoperability: To enable immediate integration of headset data with third-party algorithms within the robotics community (such as visual SLAM pipelines or navigation nodes), the topic architecture and the geometric transform (TF) tree were aligned with the de facto standards used by primary robotic vision sensors, foremost among which are the Intel RealSense family RGB-D cameras. This entailed the rigorous adoption of orientation conventions for optical frames (\texttt{\_optical\_frame}), ensuring that the camera coordinate frames comply with the standard optical convention and feature a consistent topic nomenclature.
    \item Network efficiency and resource management: Streaming multiple sensory feeds and high-resolution video frames over the network introduces a potential bottleneck regarding available bandwidth. The system was therefore constrained to optimise the transmitted payload and minimise the TCP serialization overhead, ensuring adequate update rates without saturating the local network or overloading the headset's CPU.
    \item Software modularity and extensibility: The Unity application was designed following a decoupled component-based pattern (independent scripts for each sensor modality). This design constraint ensures code maintainability and allows straightforward integration of new sensory feeds in the future (such as eye tracking or hand tracking) without requiring rewrites of the communication logic or interfering with verified modules.
\end{itemize}

\clearpage

\section{Framework architecture and parameter negotiation}
To ensure effective decoupling between the control logic on ROS 2 and execution on board the headset, the system was designed according to a modular model that initiates with an initial stage of dynamic negotiation and configuration of transmission parameters. The infrastructure comprises two nodes on the ROS 2 side (the communication endpoint and the configuration node) and a coordinated set of scripts on board the Unity application.

\subsection{System components and negotiation workflow}
The bridge operation relies on a state-based synchronization mechanism between ROS 2 and Unity:
\begin{enumerate}
    \item ROS 2 infrastructure start-up: The network endpoint (\texttt{ROS-TCP-Endpoint}), responsible for managing the low-level connection with Unity, and the configuration node are initialised. The latter loads into memory the operational parameters defined within a \texttt{config.yaml} file and begins listening on the topic dedicated to incoming requests from the headset.
    \item Configuration request (handshake): Upon launching the application on the Magic Leap 2, the coordinating script (\texttt{Configurator}) sends a synchronization message over a request topic (\texttt{/request}).
    \item Parameter serialization and dispatch: Upon receiving the request, the configuration node serialises the parameters parsed from the \texttt{YAML} file into a structured string (\texttt{JSON} format) and publishes it onto a designated response topic targeted at the headset (\texttt{/magic\_leap2/config}).
    \item Parsing and selective stream initiation: The \texttt{Configurator} on board Unity receives the \texttt{JSON} payload and parses it. The modular structure of the application provides a script (module) for managing each sensor or feature. These scripts all start up concurrently a priori, but before initiating reading routines and message transmission to ROS 2, they await the assignment of their respective parameters (enablement flags, resolutions, framerates) to internal variables by the \texttt{Configurator}.
    \item Sensor activation: Once parameter assignment is complete, the \texttt{Configurator} sets the initiation flags (\texttt{enabled}) to true solely for the activated modules. This transition unblocks the execution of streaming coroutines within the respective scripts, commencing the continuous publishing of sensory data.
\end{enumerate}
\begin{figure}[htbp]
  \centering
  \includegraphics[width=0.7\textwidth]{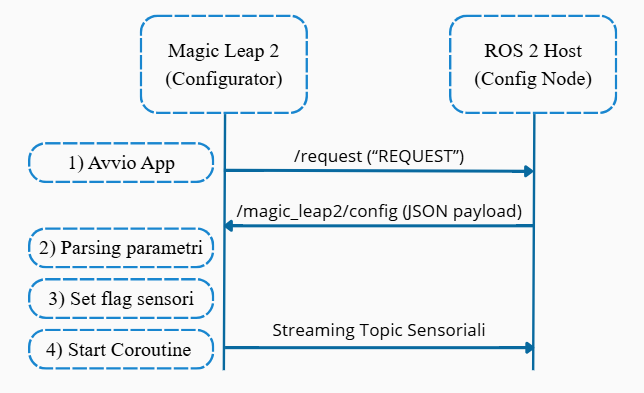}
  \caption[Parameter negotiation workflow]{Parameter negotiation workflow for transmission between the Magic Leap 2 and ROS 2. (Source: author's elaboration)}
  \label{fig:Flusso_Neg}
\end{figure}

\subsection{Structure of the configuration file (\texttt{config.yaml})}
Centralised configuration allows the user to define beforehand the computational load and network bandwidth occupancy without modifying or recompiling the code on Unity or Android. Through the \texttt{config.yaml} file, it is possible to specify:
\begin{itemize}
    \item Headset topic prefix: By default, the topics published by the headset follow the prefix \texttt{/magic\_leap2/...}, but this naming scheme can be customised as desired.
    \item Sensor enablement: Boolean flags to selectively activate or deactivate each component (tracking cameras, RGB camera, ToF sensor, IMUs, and ambient light sensor).
    \item Stream type (where supported by the headset): The headset allows certain sensors to offer multiple distinct streams with completely different functionalities; for instance, the \texttt{picture\_center} camera provides one stream for raw images and another displaying mixed reality, namely the camera frames overlaid with the elements projected by the headset.
    \item Image resolution and format (where supported by the headset): Pixel dimensions and video encoding for each enabled camera, tailored to the requirements of the receiving algorithm (e.g., to balance visual accuracy against network load).
    \item Publishing rates (where supported by the headset): Framerate parameters to set the invocation periodicity of the routines dispatching frames and environmental data.
\end{itemize}

An example of the \texttt{config.yaml} configuration file is provided below:
\begin{minted}{yaml}
/config_ML_node:
  ros__parameters:

    general:
      topic_name : "/magic_leap2"

    picture_center: 
      enabled: false
      # stream 0 is the simple RGB camera stream. 
      # stream 1 is the AR stream with the world as background and the user as
      # foreground.
      stream: 0
      # for stream 0 the possible resolution are 640x480, 1280x720, 1920x1080, 
      # 3840x2160, 2048x1536, 1280x960, 1440x1080, 2880x2160, 4096x3072.
      # for stream 1 the possible resolution are 648x720, 972x1080, 1944x2160, 
      # 960x720, 1440x1080, 2880x2160.
      resolution_width: 1920
      resolution_height: 1080
      update_rate: 30                   # possible values in range [30, 60]
      format: "jpeg"

    depth_center: 
      enabled: false
      # stream 0 is the long range depth stream.
      # stream 1 is the short range depth stream.
      stream: 1
      resolution_width: 544
      resolution_height: 480
      # for stream 0 the possible update rates are 1, 5.
      # for stream 1 the possible update rates are 5, 30, 60.
      update_rate: 30
      format: "Depth32" # possible values = "Depth32", "DepthRaw".

    world_center: 
      enabled: false
      stream: 0                          # possible values = 0, 1
      resolution_width: 1016
      resolution_height: 1016
      update_rate: 30
      format: "Grayscale"

    world_right: 
      enabled: false
      stream: 0                          # possible values = 0, 1
      resolution_width: 1016
      resolution_height: 1016
      update_rate: 30
      format: "Grayscale"

    world_left: 
      enabled: false
      stream: 0                           # possible values = 0, 1
      resolution_width: 1016
      resolution_height: 1016
      update_rate: 30
      format: "Grayscale"

    eye_temple_right:
      enabled: false
      stream: 0
      resolution_width: 400
      resolution_height: 400
      update_rate: 30
      format: "Grayscale"

    eye_temple_left:
      enabled: false
      stream: 0
      resolution_width: 400
      resolution_height: 400
      update_rate: 30
      format: "Grayscale"

    eye_nasal_right:
      enabled: false
      stream: 0
      resolution_width: 400
      resolution_height: 400
      update_rate: 30
      format: "Grayscale"

    eye_nasal_left:
      enabled: false
      stream: 0
      resolution_width: 400
      resolution_height: 400
      update_rate: 30
      format: "Grayscale"

    # the other sensors include gyroscope, accelerometer and 
    # ambient light sensor.
    other_sensors:
      enabled: false
\end{minted}

\subsection{Component-based architecture and execution model in Unity}
Within the Unity environment, the application is structured around a single central GameObject associated with dedicated, independent C\# scripts for each type of sensory stream (cameras, inertial sensors, and ambient light sensors). This organisation ensures strong modularity and isolation between acquisition processes.
Each module implements a common execution model based on the following phases:
\begin{itemize}
    \item Initialisation (Start): Configures internal data structures and ROS 2 publishers, maintaining a boolean enablement flag initially set to false.
    \item Unblocking and stream control (Update): At each rendering frame, the Update method monitors the status of the enablement flag. As soon as the configuration script sets the flag to true, a dedicated asynchronous coroutine is launched to acquire frames and publish \texttt{sensor\_msgs/Image} and \texttt{sensor\_msgs/CameraInfo} messages alongside their corresponding timestamp.
    \item Continuous pose updating: In parallel with frame transmission, the Update loop continues to extract the sensor's updated 6-DoF pose (expressed with respect to the reference optical frame) and publish it to ROS 2. This decoupling allows the geometric transform tree (TF) to remain updated at high frequencies, without being constrained by the frame rate of the camera feeds.
    \item Low-level sensor integration: For streams not natively handled via standard Unity APIs (such as IMUs and ambient light sensors), the dedicated script queries at runtime a native plug-in developed in Kotlin for the Android OS, exposing the readings to the publishing pipeline following the same synchronization paradigm.
\end{itemize}

\section{Implementation and characterisation of sensory modules}
To enable comprehensive perception of the environment and user dynamics, the developed application acquires the full suite of sensory streams provided by the Magic Leap 2 headset hardware and publishes them in real time onto dedicated ROS 2 topics. Specifically, the bridge manages:
\begin{itemize}
    \item Ambient tracking cameras (World Cameras): three low-level monochrome (greyscale) streams, complete with intrinsic calibration and spatial poses;
    \item Front RGB camera (Picture Center): high-definition images compressed in JPEG format, available both as a standard optical video stream and as a mixed-reality view with overlaid holographic elements;
    \item Depth sensor (Depth Camera): Time-of-Flight (ToF) data configurable for metric distance measurements (Depth32) or for capturing reflected infrared intensity (DepthRaw);
    \item Eye Tracking cameras: four infrared streams dedicated to the analysis of gaze and ocular movements;
    \item Auxiliary on-board sensors (IMUs and ambient light): linear accelerations, angular velocities, and external illuminance levels, extracted with hardware timestamps via a native Android plug-in.
\end{itemize}
The following subsections provide a detailed description of the implementation, topic structure, message formats, spatial transformations (TF Tree), and operational characteristics of each sensory module.
\clearpage
\subsection{Ambient tracking cameras (World Cameras)}
\begin{figure}[htbp]
    \centering
    \includegraphics[width=0.33\textwidth]{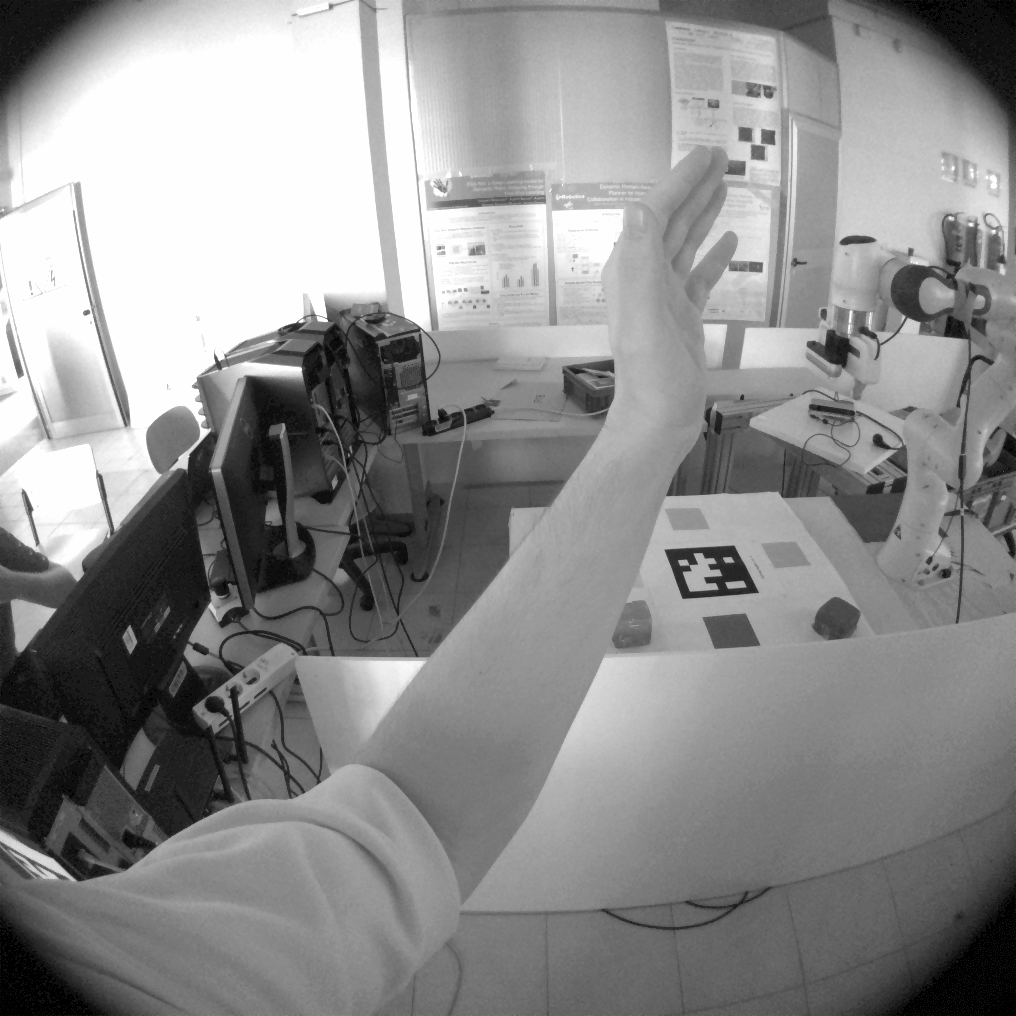}\hfill
    \includegraphics[width=0.33\textwidth]{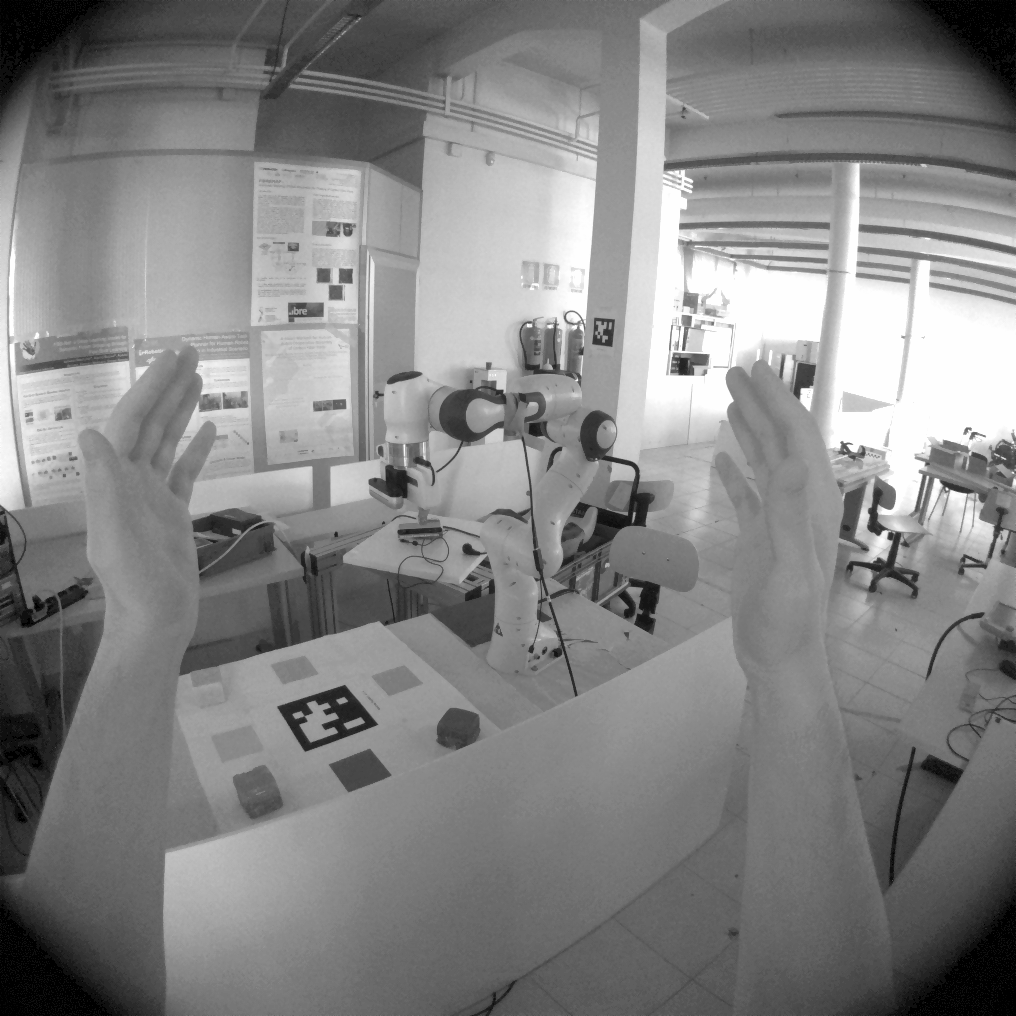}\hfill
    \includegraphics[width=0.33\textwidth]{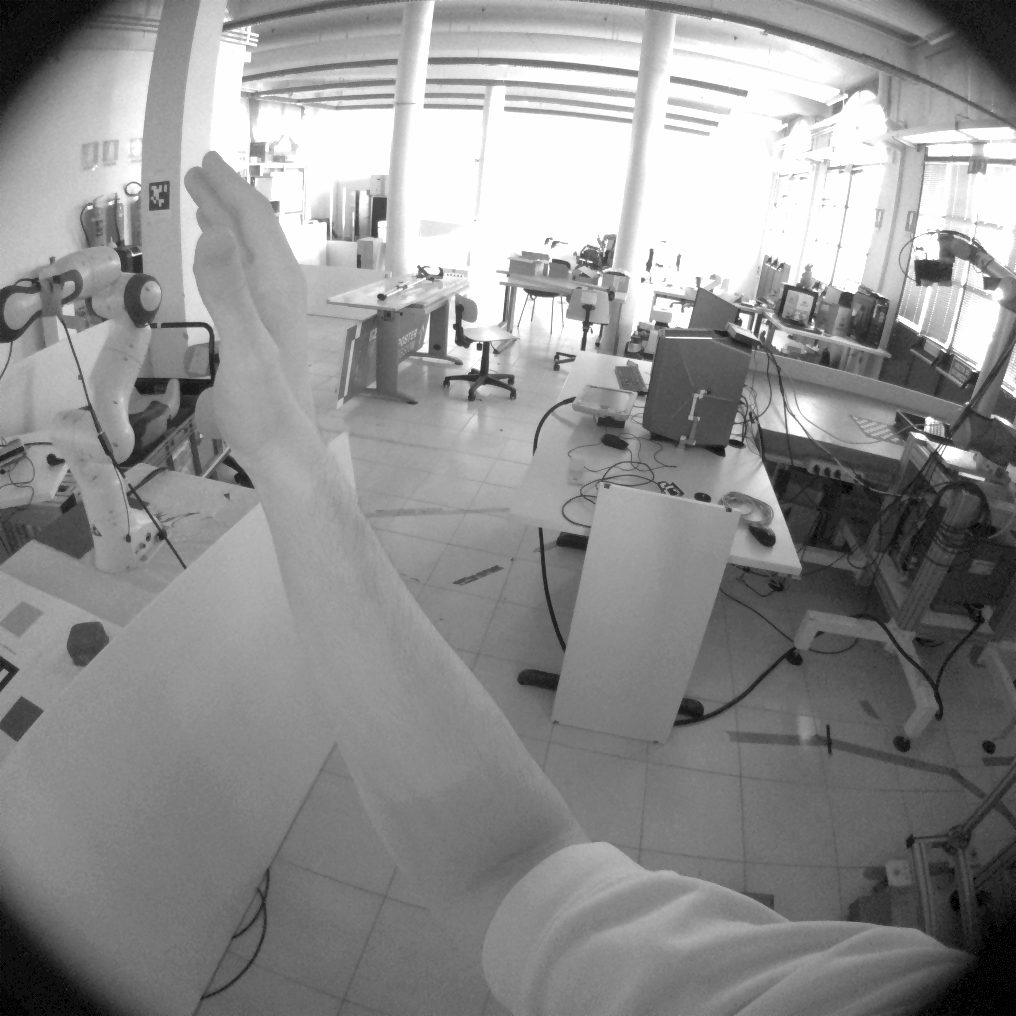}
    \caption{Example acquisition of greyscale frames originating from the three ambient tracking cameras (World Cameras): from left to right, \textit{World Left}, \textit{World Center}, and \textit{World Right}.}
    \label{fig:Telecamere_World_Left_Center_Right}
\end{figure}
\subsubsection{Description and operation}
The headset integrates three monochrome wide Field of View (FOV) cameras, optimised for low latency to enable spatial analysis and environmental understanding. As described in Section 3.1.2, the sensors are arranged along the front and sides of the headset and are conventionally designated as:
\begin{itemize}
    \item World Center: central camera;
    \item World Left: left lateral camera;
    \item World Right: right lateral camera.
\end{itemize}
Each sensor captures greyscale images at a fixed resolution of $1016 \times 1016$ pixels with a sampling frequency of 30 Hz. Considering a colour depth of 8 bits per pixel, the raw data stream generated by a single camera yields a throughput of approximately 250 Mbit/s.
\subsubsection{ROS 2 topics and messages}
Each camera can be enabled or disabled independently. Topics and the corresponding geometric transformations are registered and published exclusively if the respective sensory module is enabled. To avoid duplication, the topic structure for a generic camera <name> (where \texttt{<name>} $\in$ [ {\texttt{world\_center}, \texttt{world\_left}, \texttt{world\_right}} ]) is defined as follows:

\begin{table}[htbp]
  \centering
  \small 
  \begin{tabularx}{\textwidth}{@{} >{\raggedright\arraybackslash}p{4.8cm} >{\raggedright\arraybackslash}p{3.8cm} X @{}}
    \toprule
    \textbf{Topic} & \textbf{Message Type} & \textbf{Description} \\
    \midrule
    \texttt{/magic\_leap2/<name>/\newline image\_raw} & 
    \texttt{sensor\_msgs/}\newline\texttt{msg/Image} & 
    Raw stream of acquired frames. \\
    \addlinespace
    \texttt{/magic\_leap2/<name>/\newline camera\_info} & 
    \texttt{sensor\_msgs/}\newline\texttt{msg/CameraInfo} & 
    Intrinsic parameters and camera calibration model. \\
    \addlinespace
    \texttt{/magic\_leap2/<name>/\newline metadata} & 
    \texttt{std\_msgs/}\newline\texttt{msg/String} & 
    Sensor metadata (e.g., analogue/digital gain, exposure time). \\
    \bottomrule
  \end{tabularx}
\end{table}

\subsubsection{Spatial transformations (TF Tree)}
The bridge publishes the camera poses onto the standard \texttt{/tf} topic at a frequency of 60 Hz, expressed relative to the fixed coordinate frame \texttt{magicleap\_world} (origin defined upon application start-up).
For each enabled sensor, two specific coordinate frames are published to ensure full compatibility with both robotic standards and computer vision libraries:
\begin{itemize}
    \item \texttt{<name>\_frame}: oriented according to the standard ROS convention ($X$-axis forward, $Y$-axis to the left, $Z$-axis pointing upwards);
    \item \texttt{<name>\_optical\_frame}: oriented according to the standard optical convention ($Z$-axis along the optical axis pointing forward, $X$-axis to the right, $Y$-axis pointing downwards).
\end{itemize}

\subsubsection{Limitations and potential issues}
The experimental characterisation of the three ambient tracking cameras highlighted optimal performance: the video stream maintains a strictly stable sampling rate of 30 Hz, with negligible jitter and excellent responsiveness to lighting variations. The only critical issue is related to the impact on the transmission channel: as these are uncompressed, high-resolution streams ($1016 \times 1016$ pixels at 8 bits), the concurrent activation of all three modules generates an aggregate network load of approximately 750 Mbit/s. Such a data volume can saturate the bandwidth available on conventional Wi-Fi networks, making it advisable to selectively enable only the camera of interest (e.g., World Center alone) via the \texttt{config.yaml} file, or to employ a dedicated wired connection.
\clearpage

\subsection{High-resolution RGB front camera (Picture Center)}

\begin{figure}[htbp]
    \centering
    \includegraphics[width=0.48\textwidth]{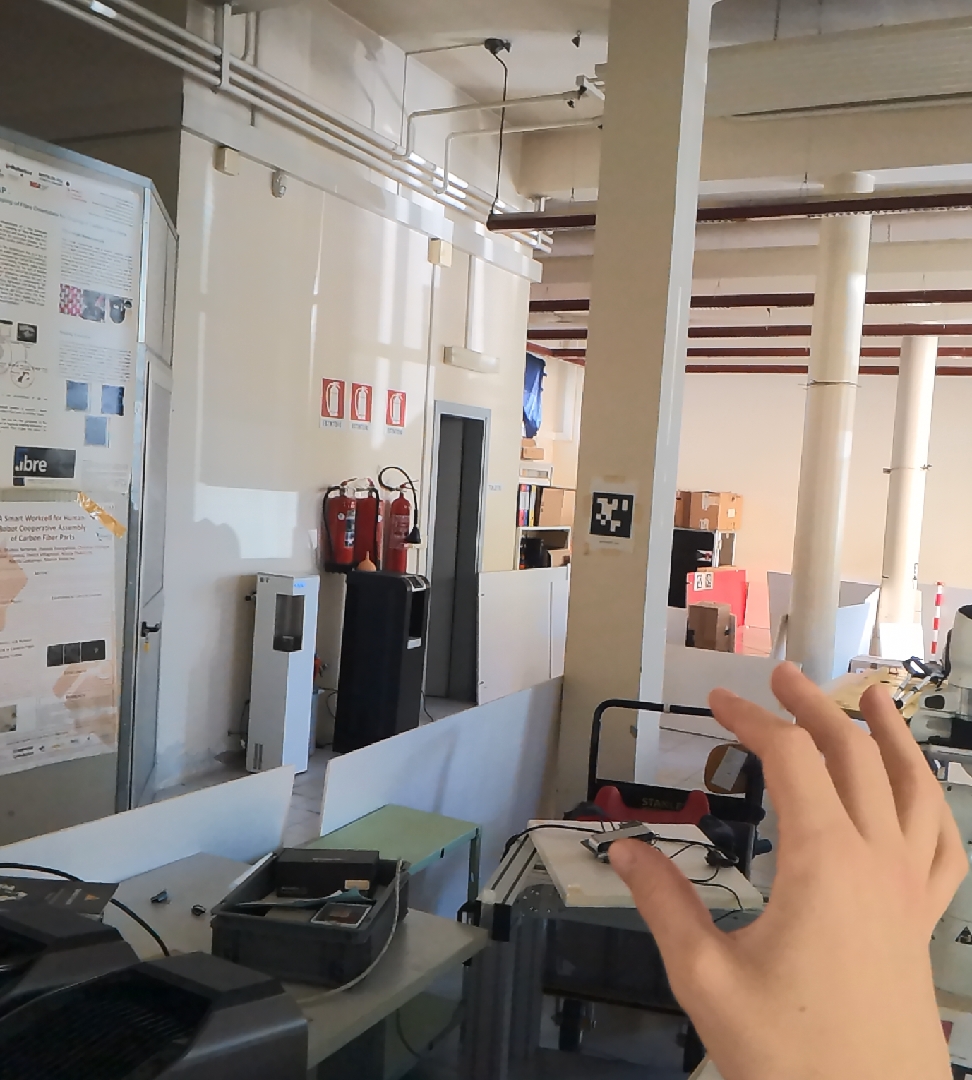}\hfill
    \includegraphics[width=0.48\textwidth]{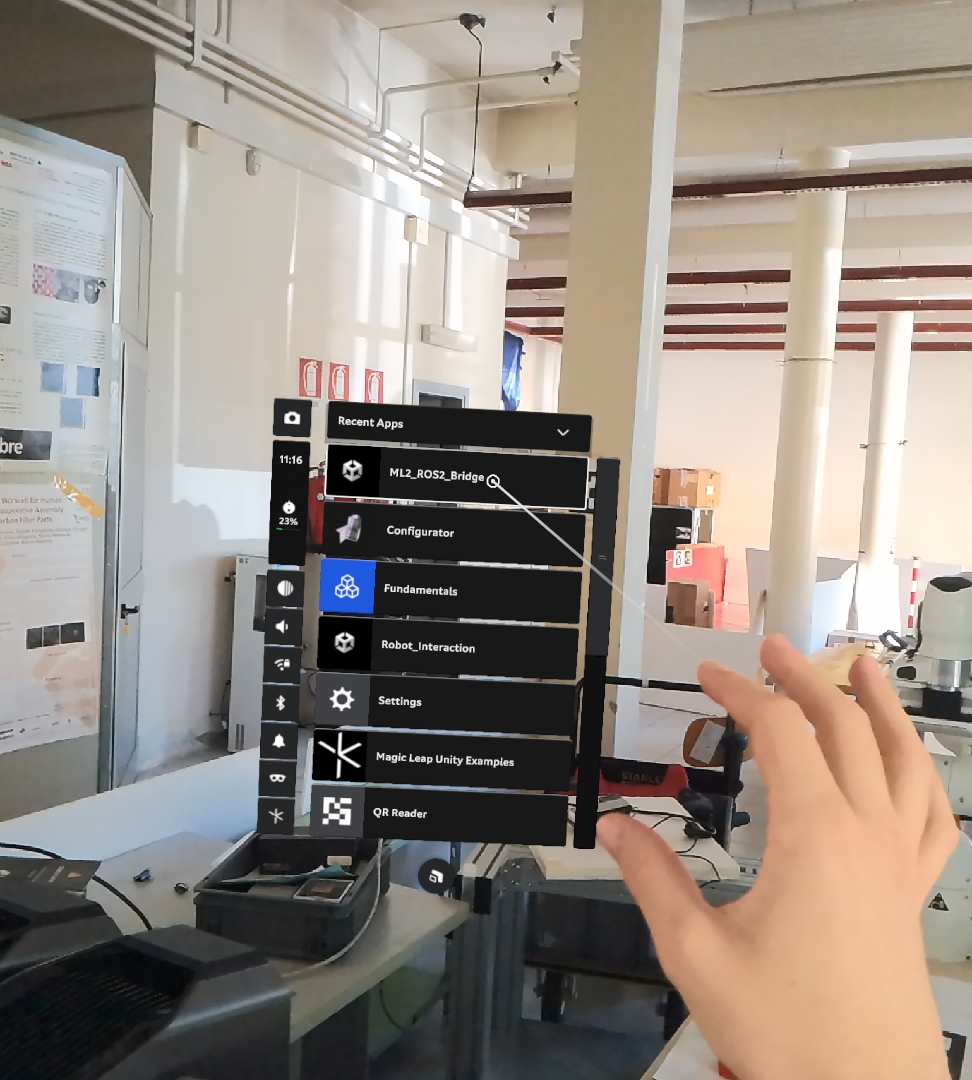}
    \caption{Streams acquired from the high-resolution RGB front camera (\textit{Picture Center}): on the left, the conventional video stream (\textit{Stream 0}); on the right, the mixed-reality stream (\textit{Stream 1}) featuring the overlay of user interface holographic elements.}
    \label{fig:Telecamera_RGB_e_Mixed_Reality}
\end{figure}
\subsubsection{Description and operation}
Positioned frontally at the centre of the headset (conventionally designated Picture Center), this high-resolution camera is designed for acquiring fine visual details, still photographs, and high-definition recordings. The acquisition pipeline natively converts and compresses frames into \texttt{JPEG} format directly on board the device to optimise the network payload. The headset APIs provide two distinct streaming modalities:
\begin{itemize}
    \item Stream 0 (Standard RGB Stream): Conventional optical video stream of the RGB camera.
    \item Stream 1 (Mixed Reality / AR Stream): Composite synthetic stream wherein graphical elements and holograms displayed to the user within the headset are overlaid onto the real-world video feed.
\end{itemize}
Both modalities permit the selection of nominal sampling frequencies of 30 Hz or 60 Hz and support a broad spectrum of resolutions:
\begin{itemize}
    \item Stream 0 resolutions: $640 \times 480$, $1280 \times 720$, $1920 \times 1080$, $3840 \times 2160$, $2048 \times 1536$, $1280 \times 960$, $1440 \times 1080$, $2880 \times 2160$, $4096 \times 3072$ pixels.
    \item Stream 1 resolutions: $648 \times 720$, $972 \times 1080$, $1944 \times 2160$, $960 \times 720$, $1440 \times 1080$, $2880 \times 2160$ pixels.
\end{itemize}
Owing to the JPEG compression embedded within the payload, bandwidth occupancy remains significantly constrained compared to an uncompressed raw stream, scaling as a function of the selected resolution and framerate.

\subsubsection{ROS 2 topics and messages}
Depending on the modality selected in the configuration file (\texttt{config.yaml}), the bridge registers and publishes data onto the corresponding ROS 2 topics:

Topics for standard RGB mode (Stream 0):

\begin{table}[htbp]
  \centering
  \small 
  \begin{tabularx}{\textwidth}{@{} >{\raggedright\arraybackslash}p{4.8cm} >{\raggedright\arraybackslash}p{3.8cm} X @{}}
    \toprule
    \textbf{Topic} & \textbf{Message Type} & \textbf{Description} \\
    \midrule
    \texttt{/magic\_leap2/color/\newline image\_raw/compressed} & 
    \texttt{sensor\_msgs/}\newline \texttt{msg/CompressedImage} & 
    Stream of acquired compressed frames. \\
    \addlinespace
    \texttt{/magic\_leap2/color/\newline camera\_info} & 
    \texttt{sensor\_msgs/}\newline\texttt{msg/CameraInfo} & 
    Intrinsic parameters and camera calibration model. \\
    \addlinespace
    \texttt{/magic\_leap2/color/\newline metadata} & 
    \texttt{std\_msgs/}\newline \texttt{msg/String} & 
    Sensor metadata (e.g., analogue/digital gain, exposure time). \\
    \bottomrule
  \end{tabularx}
\end{table}

Topics for Mixed Reality mode (Stream 1):

\begin{table}[htbp]
  \centering
  \small 
  \begin{tabularx}{\textwidth}{@{} >{\raggedright\arraybackslash}p{4.8cm} >{\raggedright\arraybackslash}p{3.8cm} X @{}}
    \toprule
    \textbf{Topic} & \textbf{Message Type} & \textbf{Description} \\
    \midrule
    \texttt{/magic\_leap2/mixed\_rea\newline lity/image\_raw/compressed} & 
    \texttt{sensor\_msgs/}\newline\texttt{msg/CompressedImage} & 
    Stream of acquired compressed frames. \\
    \addlinespace
    \texttt{/magic\_leap2/mixed\newline\_reality/camera\_info} & 
    \texttt{sensor\_msgs/}\newline\texttt{msg/CameraInfo} & 
    Intrinsic parameters and camera calibration model. \\
    \addlinespace
    \texttt{/magic\_leap2/\newline mixed\_reality/metadata} & 
    \texttt{std\_msgs/}\newline\texttt{msg/String} & 
    Sensor metadata (e.g., analogue/digital gain, exposure time). \\
    \bottomrule
  \end{tabularx}
\end{table}

\subsubsection{Spatial transformations (TF Tree)}
The bridge publishes the camera poses onto the standard \texttt{/tf} topic at a frequency of 60 Hz, expressed relative to the fixed coordinate frame \texttt{magicleap\_world} (origin defined upon application start-up). Two specific coordinate frames are published:
\begin{itemize}
    \item \texttt{color\_frame}: oriented according to the standard ROS convention ($X$-axis forward, $Y$-axis to the left, $Z$-axis pointing upwards);
    \item \texttt{color\_optical\_frame}: oriented according to the standard optical convention ($Z$-axis along the optical axis pointing forward, $X$-axis to the right, $Y$-axis pointing downwards).
\end{itemize}
\subsubsection{Limitations and potential issues}
The experimental characterisation of the module highlighted several operational criticalities primarily related to the thermal management of the device:
\begin{itemize}
    \item Overheating and thermal throttling: High-resolution acquisition combined with the continuous computational overhead of JPEG compression induces a rapid increase in the headset's internal temperature.
    \item Framerate degradation: When configuring a nominal frequency of 60 Hz, the sensor seldom sustains this operating regime at start-up, initially settling at lower rates. As the temperature rises, thermal throttling intervenes, progressively degrading the effective framerate over approximately ten minutes.
\end{itemize}
For these reasons, the module is ill-suited to applications demanding continuous high-frequency streams (such as real-time visual odometry); instead, it lends itself optimally to periodic visual inspection tasks, capturing ultra-high-resolution static frames, or recording short mixed-reality sequences for operator supervision.

\clearpage

\subsection{Depth sensor (Depth camera)}

\begin{figure}[htbp]
    \centering
    \begin{subfigure}[b]{0.48\textwidth}
        \centering
        \includegraphics[width=\textwidth]{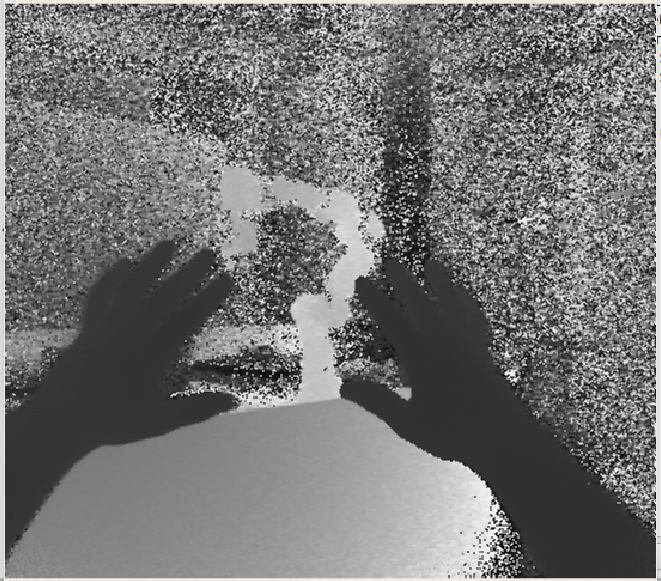}
        \caption{}
        \label{fig:Telecamera_Depth_Short}
    \end{subfigure}
    \hfill
    \begin{subfigure}[b]{0.48\textwidth}
        \centering
        \includegraphics[width=\textwidth]{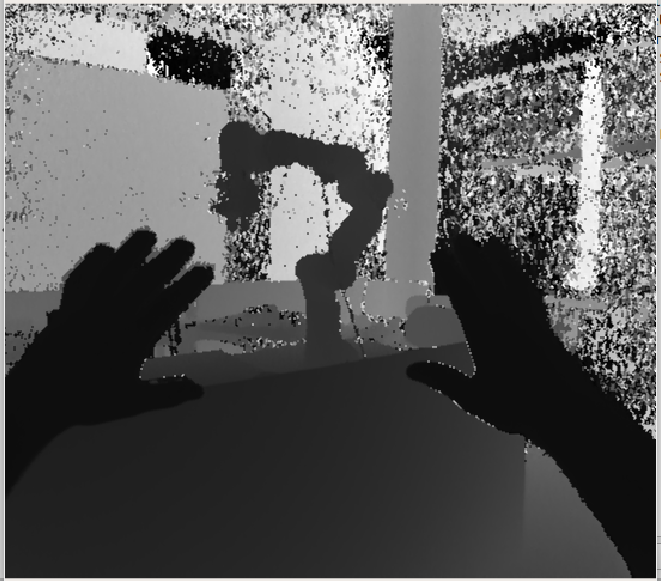}
        \caption{}
        \label{fig:Telecamera_Depth_Long}
    \end{subfigure}

    \vspace{0.5cm} 

    \begin{subfigure}[b]{0.48\textwidth}
        \centering
        \includegraphics[width=\textwidth]{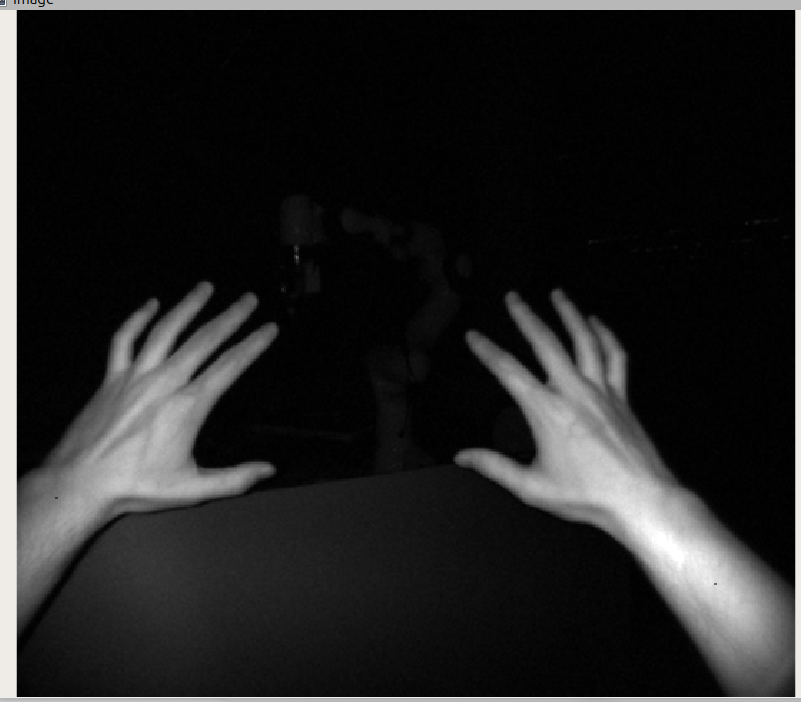}
        \caption{}
        \label{fig:Telecamera_Infra_Short}
    \end{subfigure}
    \hfill
    \begin{subfigure}[b]{0.48\textwidth}
        \centering
        \includegraphics[width=\textwidth]{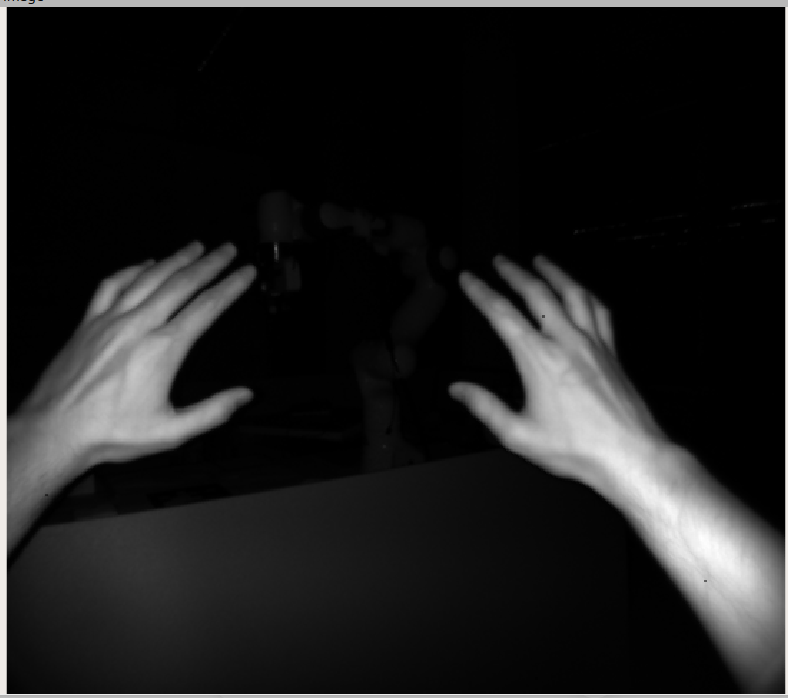}
        \caption{}
        \label{fig:Telecamera_Infra_Long}
    \end{subfigure}

    \caption{Visual comparison among the four operational modes of the Time-of-Flight (ToF) depth sensor: (a) Short Range with metric encoding \textit{Depth32}; (b) Long Range with metric encoding \textit{Depth32}; (c) Short Range with raw infrared intensity \textit{DepthRaw}; (d) Long Range with raw infrared intensity \textit{DepthRaw}.}
    \label{fig:Quattro_Immagini_Depth_Sensor}
\end{figure}

\subsubsection{Description and operation}
Positioned frontally (conventionally designated Depth Center), the depth sensor relies on Time-of-Flight (ToF) technology to reconstruct the three-dimensional geometry of the surrounding environment. The sensor operates at a fixed resolution of $544 \times 480$ pixels and provides data in two distinct output formats:
\begin{itemize}
    \item \texttt{Depth32} (Depth Map): A matrix of 32-bit floating-point values (\texttt{float32}) where each element represents the estimated metric distance of the point relative to the sensor's focal plane.
    \item \texttt{DepthRaw} (Infrared Intensity): A matrix of 32-bit values (\texttt{float32}) containing the raw quantity of reflected infrared light captured by the receiver, particularly suited to motion capture applications or the detection of retro-reflective markers.
\end{itemize}
The headset APIs provide two operational streaming modes:
\begin{itemize}
    \item Stream 0 (Long Range): A configuration optimised for medium-to-long-range environmental mapping (distances ranging between $0.5\text{ m}$ and $5.0\text{ m}$) with centimetre-level accuracy. It supports configurable sampling rates of 1 Hz and 5 Hz.
    \item Stream 1 (Short Range): A high-precision configuration with millimetre-level accuracy for close-range distances (between $0.2\text{ m}$ and $1.2\text{ m}$), ideal for fine interaction, hand tracking, and gesture estimation. It supports selectable sampling rates of 5 Hz and 30 Hz.
\end{itemize}
Considering a resolution of $544 \times 480$ pixels and a payload of 4 bytes per pixel (\texttt{float32} format), the theoretical uncompressed throughput generated by the sensor is approximately:
\begin{itemize}
    \item $\approx 8.35\text{ Mbit/s}$ at $1\text{ Hz}$;
    \item $\approx 41.78\text{ Mbit/s}$ at $5\text{ Hz}$;
    \item $\approx 250.68\text{ Mbit/s}$ at $30\text{ Hz}$.
\end{itemize}

\subsubsection{ROS 2 topics and messages}
Depending on the data format selected in the \texttt{config.yaml} file, the bridge registers and publishes information onto the corresponding topics:

Topics for the metric depth format (\texttt{Depth32}):
\begin{table}[htbp]
  \centering
  \small
  \begin{tabularx}{\textwidth}{@{} >{\raggedright\arraybackslash}p{4.8cm} >{\raggedright\arraybackslash}p{3.8cm} X @{}}
    \toprule
    \textbf{Topic} & \textbf{Message Type} & \textbf{Description} \\
    \midrule
    \texttt{/magic\_leap2/depth/\newline image\_raw} & 
    \texttt{sensor\_msgs/}\newline \texttt{msg/Image} & 
    Raw matrix of metric depth values (32FC1 encoding). \\
    \addlinespace
    \texttt{/magic\_leap2/depth/\newline camera\_info} & 
    \texttt{sensor\_msgs/}\newline \texttt{msg/CameraInfo} & 
    Intrinsic parameters and calibration model of the ToF sensor. \\
    \addlinespace
    \texttt{/magic\_leap2/depth/\newline metadata} & 
    \texttt{std\_msgs/}\newline \texttt{msg/String} & 
    Operational metadata (integration times, IR illumination modes). \\
    \bottomrule
  \end{tabularx}
\end{table}
\clearpage
Topics for the infrared intensity format (\texttt{DepthRaw}):
\begin{table}[htbp]
  \centering
  \small
  \begin{tabularx}{\textwidth}{@{} >{\raggedright\arraybackslash}p{4.8cm} >{\raggedright\arraybackslash}p{3.8cm} X @{}}
    \toprule
    \textbf{Topic} & \textbf{Message Type} & \textbf{Description} \\
    \midrule
    \texttt{/magic\_leap2/infra/\newline image\_raw} & 
    \texttt{sensor\_msgs/}\newline\texttt{msg/Image} & 
    Raw matrix of IR reflectance intensity (32FC1 encoding). \\
    \addlinespace
    \texttt{/magic\_leap2/infra/\newline camera\_info} & 
    \texttt{sensor\_msgs/}\newline\texttt{msg/CameraInfo} & 
    Intrinsic parameters and sensor calibration model. \\
    \addlinespace
    \texttt{/magic\_leap2/infra/\newline metadata} & 
    \texttt{std\_msgs/}\newline\texttt{msg/String} & 
    IR receiver metadata. \\
    \bottomrule
  \end{tabularx}
\end{table}

\subsubsection{Spatial transformations (TF Tree)}
The bridge publishes the ToF sensor poses onto the \texttt{/tf} topic at a frequency of 60 Hz relative to the \texttt{magicleap\_world} reference frame:
\begin{itemize}
    \item \texttt{depth\_frame}: oriented according to the standard ROS convention ($X$-axis forward, $Y$-axis to the left, $Z$-axis pointing upwards);
    \item \texttt{depth\_optical\_frame}: oriented according to the standard optical convention ($Z$-axis along the optical axis pointing forward, $X$-axis to the right, $Y$-axis pointing downwards).
\end{itemize}

\subsubsection{Limitations and potential issues}
Experimental validation demonstrated robust nominal behaviour, characterised by low latencies and a steady throughput free from fluctuations.
Nevertheless, an API-level discrepancy emerged regarding the Short Range mode (Stream 1): although the headset query primitives indicate that a 60 Hz framerate is selectable (and the application accepts this parameter without throwing exceptions or interrupting the stream), the actual publishing rate remains constrained by a maximum hardware/firmware threshold of 30 Hz.
\clearpage

\subsection{Eye tracking cameras and gaze estimation}
\begin{figure}[htbp]
    \centering
    \begin{subfigure}[b]{0.45\textwidth}
        \centering
        \includegraphics[width=\textwidth]{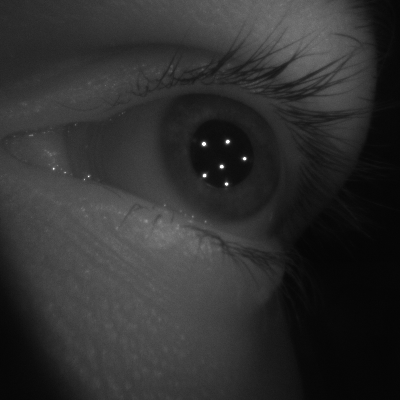}
        \caption{}
        \label{fig:Telecamera_Eye_Nasal_Right}
    \end{subfigure}
    \hfill
    \begin{subfigure}[b]{0.45\textwidth}
        \centering
        \includegraphics[width=\textwidth]{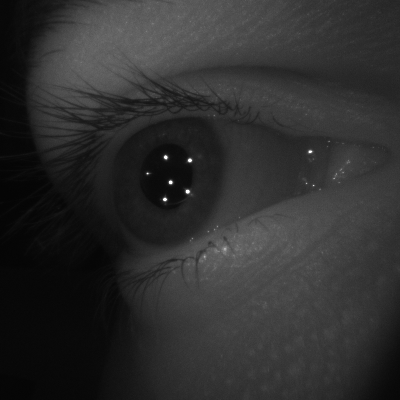}
        \caption{}
        \label{fig:Telecamera_Eye_Nasal_Left}
    \end{subfigure}

    \vspace{0.5cm} 

    \begin{subfigure}[b]{0.45\textwidth}
        \centering
        \includegraphics[width=\textwidth]{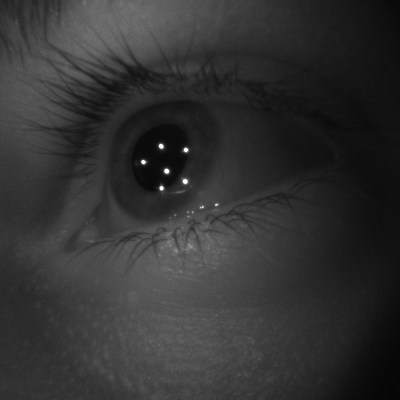}
        \caption{}
        \label{fig:Telecamera_Eye_Temple_Right}
    \end{subfigure}
    \hfill
    \begin{subfigure}[b]{0.45\textwidth}
        \centering
        \includegraphics[width=\textwidth]{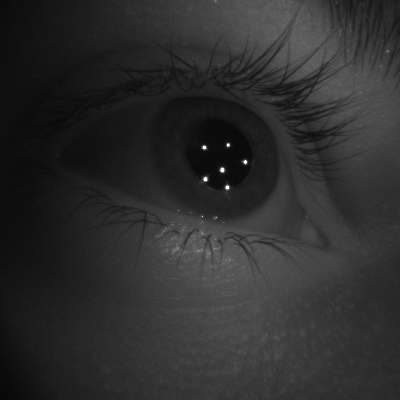}
        \caption{}
        \label{fig:Telecamera_Eye_Temple_Left}
    \end{subfigure}

    \caption{Infrared streams acquired by the four dedicated eye tracking cameras: (a) \textit{Eye Nasal Right}, (b) \textit{Eye Nasal Left}, (c) \textit{Eye Temple Right}, and (d) \textit{Eye Temple Left}.}
    \label{fig:Quattro_Immagini_Eye_Tracking}
\end{figure}
\clearpage
\subsubsection{Description and operation}
The headset integrates four monochrome infrared cameras specifically directed towards the user's eyes, dedicated to ocular tracking (eye tracking) and gaze direction estimation (gaze recognition). The sensors are arranged in pairs (one on the nasal side and one on the temporal side for each eye) and are conventionally designated as:
\begin{itemize}
    \item \texttt{eye\_nasal\_right}: nasal camera for the right eye;
    \item \texttt{eye\_temple\_right}: temporal camera for the right eye;
    \item \texttt{eye\_nasal\_left}: nasal camera for the left eye;
    \item \texttt{eye\_temple\_left}: temporal camera for the left eye.
\end{itemize}
Each sensor captures greyscale images at a fixed resolution of $400 \times 400$ pixels with a sampling frequency of 30 Hz. Considering a colour depth of 8 bits per pixel, the raw data stream generated by an individual camera yields a theoretical throughput of approximately 38.4 Mbit/s.
Cumulatively, in the event of concurrent activation of all four sensors, the generated network traffic amounts to approximately 153.6 Mbit/s.
\subsubsection{ROS 2 topics and messages}
Each camera can be independently enabled or disabled via the \texttt{config.yaml} configuration file. Topics are registered and published exclusively if the corresponding module is active. For a generic camera <name> (where $\texttt{<name>} \in \{\texttt{eye\_nasal\_right}, \texttt{eye\_temple\_right}, \texttt{eye\_nasal\_left}, \texttt{eye\_temple\_left}\}$), the topic structure is defined as follows:
\begin{table}[htbp]
  \centering
  \small 
  \begin{tabularx}{\textwidth}{@{} >{\raggedright\arraybackslash}p{4.8cm} >{\raggedright\arraybackslash}p{3.8cm} X @{}}
    \toprule
    \textbf{Topic} & \textbf{Message Type} & \textbf{Description} \\
    \midrule
    \texttt{/magic\_leap2/<name>/\newline image\_raw} & 
    \texttt{sensor\_msgs/}\newline\texttt{msg/Image} & 
    Raw stream of acquired monochrome frames (mono8 encoding). \\
    \addlinespace
    \texttt{/magic\_leap2/<name>/\newline metadata} & 
    \texttt{std\_msgs/}\newline\texttt{msg/String} & 
    Sensor metadata (e.g., exposure time and analogue/digital gains). \\
    \bottomrule
  \end{tabularx}
\end{table}
\clearpage
\subsubsection{Limitations and potential issues}
During the development and characterisation phase of the eye tracking modules, several architectural limitations imposed by the software ecosystem and the vendor's APIs emerged:
\begin{itemize}
    \item Absence of intrinsic parameters (\texttt{CameraInfo}): The Magic Leap APIs do not expose the intrinsic calibration matrix or the optical distortion coefficients for the ocular cameras, preventing the generation of the standard \texttt{sensor\_msgs/msg/CameraInfo} topic.
    \item Absence of geometric poses in the TF Tree: The operating system does not provide the spatial transformations relating the rigid placement of the four cameras to the central headset frame; consequently, dedicated reference frames are not published onto the \texttt{/tf} tree.
\end{itemize}
Notwithstanding these informational constraints regarding spatial metadata, the four video streams exhibit low latency and an exceptionally stable acquisition rate of 30 Hz, rendering them fully viable for external image processing pipelines or bespoke pupil tracking algorithms.

\subsection{Inertial measurement units (IMUs) and ambient light sensor}
\subsubsection{Description and operation}
The device hardware architecture incorporates three distinct Inertial Measurement Units (IMUs) alongside an ambient light sensor:
\begin{itemize}
    \item Headset IMUs (Left and Right): Two IMUs embedded directly into the spectacles frame, positioned respectively on the left and right sides of the chassis to accurately track head dynamics.
    \item Compute Pack IMU: A third IMU located inside the external processing unit (Compute Pack).
    \item Ambient Light Sensor: A photodiode integrated into the headset for monitoring external illuminance.
\end{itemize}
Accessing these sensors via the standard Magic Leap SDK APIs revealed critical limitations: notably, the inability to query all three IMUs concurrently and the omission of original hardware timestamps associated with the precise acquisition instant of each sample.
To overcome this constraint, a native Android plug-in was developed in Kotlin. The plug-in interfaces at a low level with the \texttt{SensorManager} service of Magic Leap OS (built on the Android Open Source Project), registering dedicated listeners for each accelerometer, gyroscope, and light sensor. At runtime, the C\# script within Unity queries the native plug-in and retrieves raw readings paired with their respective hardware timestamps at nanosecond precision.
Owing to the compact payload of scalar and vector numerical messages, the aggregate impact of these streams on network bandwidth remains negligible (on the order of a few hundred kbit/s).
\subsubsection{ROS 2 topics and messages}
Inertial and photometric data are serialised and published onto their respective standard ROS 2 topics within the \texttt{sensor\_msgs} suite:

\begin{table}[htbp]
  \centering
  \small 
  \begin{tabularx}{\textwidth}{@{} >{\raggedright\arraybackslash}p{4.8cm} >{\raggedright\arraybackslash}p{3.8cm} X @{}}
    \toprule
    \textbf{Topic} & \textbf{Message Type} & \textbf{Description} \\
    \midrule
    \texttt{/magic\_leap2/accel\_left/\newline imu} & 
    \texttt{sensor\_msgs/msg/Imu} & 
    Linear acceleration of the left IMU (headset). \\
    \addlinespace
    \texttt{/magic\_leap2/gyro\_left/\newline imu} & 
    \texttt{sensor\_msgs/msg/Imu} & 
    Angular velocity of the left IMU (headset). \\
    \addlinespace
    \texttt{/magic\_leap2/accel\_right/\newline imu} & 
    \texttt{sensor\_msgs/msg/Imu} & 
    Linear acceleration of the right IMU (headset). \\
    \addlinespace
    \texttt{/magic\_leap2/gyro\_right/\newline imu} & 
    \texttt{sensor\_msgs/msg/Imu} & 
    Angular velocity of the right IMU (headset). \\
    \addlinespace
    \texttt{/magic\_leap2/accel\_\newline compute\_pack/imu} & 
    \texttt{sensor\_msgs/msg/Imu} & 
    Linear acceleration of the Compute Pack IMU. \\
    \addlinespace
    \texttt{/magic\_leap2/gyro\_\newline compute\_pack/imu} & 
    \texttt{sensor\_msgs/msg/Imu} & 
    Angular velocity of the Compute Pack IMU. \\
    \addlinespace
    \texttt{/magic\_leap2/light\_sensor} & 
    \texttt{sensor\_msgs/msg/\newline Illuminance} & 
    Measured ambient illuminance level. \\
    \addlinespace
    \bottomrule
  \end{tabularx}
\end{table}
\subsubsection{Limitations and potential issues}
The sampling frequency exposed by the Android layer via the \texttt{SensorManager} for the inertial measurement units is constrained to approximately 40 Hz, presumably attributable to internal decimation or aggregation filters enforced by the operating system's Hardware Abstraction Layer (HAL).
Although this rate is lower than the nominal maximum regimes typical of inertial sensors (often exceeding 100--200 Hz), the 40 Hz frequency remains strictly constant and devoid of significant jitter throughout streaming. The retrieval of authentic hardware timestamps via the native plug-in ensures accurate temporal synchronisation with the visual streams, which constitutes a critical prerequisite for deployment in sensor fusion pipelines and Visual-Inertial Odometry (VIO) algorithms.

\section{User manual}
The complete framework developed to interface the Magic Leap 2 headset with the ROS 2 ecosystem is available in the official project GitHub repository~\cite{MagicLeap2_ROS2_Repo}.
This section outlines the step-by-step operational procedure for system installation, configuration, and execution, designed so that the end user is not required to interface with the Unity development environment or recompile the source code to deploy the bridge.
\subsection{ROS 2 environment setup}
On the host workstation (running ROS 2), the workspace must be configured by integrating the two software packages provided in the repository:
\begin{enumerate}
    \item Package importation: Within the \texttt{src} directory of the ROS 2 workspace, clone or copy:
    \begin{itemize}
        \item the \texttt{config\_ml2\_stream} package, responsible for managing and dispatching configuration parameters to the headset;
        \item the \texttt{ros\_tcp\_endpoint} package (version v0.7.0 for ROS 2), tasked with managing the TCP socket and routing incoming network messages onto ROS 2 topics.
    \end{itemize}
    \item Customisation of the \texttt{config.yaml} file: Prior to building or execution, navigate to the \texttt{config\_ml2\_stream} package and edit the \texttt{config.yaml} file. Within this file, the user can individually enable or disable the various sensors (World Cameras, RGB camera, Depth Camera, Eye Tracking, IMUs, and ambient light sensor) and specify operational parameters such as framerates, streaming modes, and target resolutions.
    \item Workspace compilation: Build the packages using the \texttt{colcon} build tool and source the resulting environment:
    \begin{minted}{bash}
cd ~/ros2_ws
colcon build --packages-select ros_tcp_endpoint config_ml2_stream
source install/setup.bash
    \end{minted}
\end{enumerate}

\clearpage

\subsection{Application installation and setup on Magic Leap 2}
To execute sensory streaming from the headset, installing the precompiled package (\texttt{.apk}) is sufficient, without requiring the Unity project to be reopened or rebuilt:
\begin{enumerate}
    \item APK installation: Connect the Magic Leap 2 headset to the PC via a USB-C cable and install the \texttt{.apk} file (located within the \texttt{Application} directory of the repository) using the official Magic Leap Hub 3 application.
    \item Network requirements: The precompiled application is configured out of the box to attempt a TCP connection towards the static IP address \texttt{192.168.0.203} on port 10000. It is therefore essential that the host workstation running ROS 2 is assigned this static IP address and that both devices (PC and headset) are connected to the same local subnet (via Wi-Fi or a wired network connection over USB-C).
    \item Granting system permissions (Android Permissions): As the application directly interfaces with video streams, spatial data, and low-level sensors, manually granting all permissions mandated by the Android operating system (Magic Leap OS) is compulsory prior to the very first launch:
    \begin{itemize}
        \item Don the headset and enter the Settings menu;
        \item Navigate to the Apps section and select the bridge application;
        \item Access the Permissions tab and explicitly grant all requested authorizations (specifically: \textit{Camera}, \textit{Spatial Mapping/Perception}, \textit{Sensors}, and \textit{Local Network}). Failure to grant any of these permissions will cause an immediate crash or freeze of the low-level acquisition routines.
    \end{itemize}
\end{enumerate}

\subsection{Start-up sequence and data streaming}
To ensure a successful network handshake and enable the headset to ingest the configuration parameters before streaming commences, the following start-up sequence must be strictly adhered to:
\begin{enumerate}
    \item Launching the ROS 2 communication endpoint: Open a primary terminal on the host workstation and execute the launch file for the TCP endpoint:
    \begin{minted}{bash}
ros2 launch ros_tcp_endpoint endpoint.py
    \end{minted}
    \item Launching the configuration server: Open a secondary terminal and execute the node tasked with parsing the \texttt{config.yaml} file and listening for incoming requests from the headset:
    \begin{minted}{bash}
ros2 launch config_ml2_stream magic_leap_config.launch.py
    \end{minted}
    \item Executing the application on board the headset: Once it has been verified that both ROS 2 nodes are active and listening, launch the application inside the Magic Leap 2. Upon start-up, the application:
    \begin{itemize}
        \item establishes the connection with the endpoint;
        \item automatically transmits the handshake request onto the \texttt{/request} topic;
        \item receives the JSON configuration payload from the ROS 2 node;
        \item unblocks the acquisition coroutines exclusively for the enabled sensors, initiating continuous publishing onto the respective ROS 2 topics.
    \end{itemize}
\end{enumerate}

\section{Modularity and upgrades}
The framework was engineered following a modular, decoupled architectural pattern, allowing future developers to customise the on-board headset application, adapt the default network configuration (e.g., the ROS 2 server IP address), or integrate novel sensory streams and advanced AR capabilities (such as hand tracking or on-device computer vision algorithms).

\subsection{Unity development environment setup}
To modify the source code and build a new \texttt{.apk} package, the Unity development environment must be configured according to the following steps:
\begin{enumerate}
    \item Unity project creation: Create a new project configured with Magic Leap 2 OpenXR support, adhering to the specifications and prerequisites outlined in the vendor's official documentation~\cite{sito_MagicLeap2_documentation_getting_started}.
    \item Script importation: Within the main scene, create an empty \texttt{GameObject} (acting as the logical container for the modules) and attach all the C\# scripts provided in the \texttt{Scripts} directory of the \texttt{GitHub} repository~\cite{MagicLeap2_ROS2_Repo}, ensuring they are all enabled within the Inspector.
    \item Native Android plug-in integration:
    \begin{itemize}
        \item Retrieve the \texttt{.aar} binary located within the \texttt{Android\_Plugin} directory of the repository~\cite{MagicLeap2_ROS2_Repo} and place it in the Unity project at the path \texttt{Assets/Plugins/Android/}.
        \item To enable Unity to properly link the Kotlin-based plug-in, navigate to \textit{Edit > Project Settings > Player}, select the Android platform tab (robot icon), and, under the \textit{Publishing Settings} section, tick the Custom Main Gradle Template checkbox.
        \item Open the generated \texttt{Assets/Plugins/Android/mainTemplate.gradle} file with a text editor, locate the dependencies block at the bottom of the file, and append the Kotlin standard library dependency:
        \begin{minted}{Groovy}
dependencies {
    // ... other dependencies auto-generated by Unity ...
    // Dependency for the native sensor plug-in:
    implementation "org.jetbrains.kotlin:kotlin-stdlib:1.9.20"
}
        \end{minted}
    \end{itemize}
    \item Installation and configuration of the \texttt{ROS-TCP-Connector} package:
        \begin{itemize}
            \item Open the Unity \textit{Package Manager} (\textit{Window > Package Manager}), click the + button in the upper left corner, and select \textit{Add package from git URL}.
            \item Provide the ROS connector repository URL:
            \begin{minted}{bash}
https://github.com/Unity-Technologies/ROS-TCP-Connector.git?
path=/com.unity.robotics.ros-tcp-connector
            \end{minted}
            \item Upon completing the installation, access the new Robotics menu item in Unity's top navigation bar: within the configuration window, select ROS2 as the protocol version and set the IP address of the host ROS 2 workstation should it differ from the default value (\texttt{192.168.0.203}).
        \end{itemize}
    For further details, refer to the official repository~\cite{ROS_TCP_Connector}.
\end{enumerate}

\subsection{Guidelines for code extension and upgrades}
The software architecture enables developers to introduce modifications or integrate new modules following a standardised three-step design pattern:
\begin{enumerate}
    \item Extending the configuration structure (\texttt{Configurator.cs}):
    \begin{itemize}
        \item To add novel parameters configurable via YAML (such as a custom framerate for an additional sensor or extra activation flags), developers must extend the internal \texttt{AppConfig} class defined within the \texttt{Configurator.cs} script.
        \item Upon application start-up, the JSON parsing routine automatically populates the \texttt{AppConfig} data structure with the updated values parsed from the \texttt{config.yaml} file.
    \end{itemize}
    \item Parameter dispatching to functional modules:
    \begin{itemize}
        \item Within the \texttt{Configurator.cs} script, insert the logic required to assign the newly ingested parameters to the public properties of the respective target scripts prior to setting their activation flags to \texttt{true}.
    \end{itemize}
    \item Implementing new sensory modules or AR features:
    \begin{itemize}
        \item Any new module (such as an Hand Tracking manager, on-device computer vision algorithms, or a 3D holographic renderer for HRC) can be implemented by creating a standalone, decoupled C\# script.
        \item The script simply needs to:
        \begin{itemize}
            \item expose a boolean activation flag (e.g., \texttt{is\_active}) managed directly by \texttt{Configurator.cs};
            \item handle data sampling and message serialisation towards the ROS 2 connector publishers within the Unity \texttt{Update} loop or via an asynchronous coroutine;
            \item interface with the native Android plug-in if access to low-level operating system hardware telemetry or metadata is required.
        \end{itemize}
    \end{itemize}
\end{enumerate}
    \cleardoublepage

    \chapter{Experimental Validation and Results}

\section{Experimental setup and acquisition procedure}
To validate the temporal accuracy, video stream integrity, and overall stability of the developed bridge, the framework was evaluated by integrating the architecture with a Visual Simultaneous Localisation and Mapping (Visual SLAM) algorithm. Specifically, trajectory estimation was conducted using ORB-SLAM3 in monocular mode (Mono SLAM), fed by streams from the ambient tracking cameras (World Cameras).
The experimental campaign was carried out within an unstructured indoor environment. To isolate the individual behaviour of each optical sensor and prevent interference or network bottlenecks stemming from bandwidth saturation, the tests were conducted by enabling a single stream at a time:
\begin{enumerate}
    \item Sensor configuration: Via the \texttt{config.yaml} file, the first camera (World Center) was enabled individually.
    \item Launch and streaming: The ROS 2 endpoint, configuration node, and on-board Magic Leap 2 application were initialised, starting the 30 Hz stream of the corresponding \texttt{/magic\_leap2/world\_center/image\_raw} topic alongside the \texttt{CameraInfo} messages.
    \item SLAM execution and data recording: The monocular ORB-SLAM3 node was launched, and concurrent recording to a \texttt{ros2 bag} file was initiated, capturing:
    \begin{itemize}
        \item the real-time poses estimated by the SLAM algorithm;
        \item the concurrent poses estimated by the Magic Leap 2 internal localisation system (published onto the \texttt{/tf} topic), utilised as the baseline reference trajectory (pseudo-ground truth).
    \end{itemize}
    \item Iteration across remaining modules: The test session was concluded and systematically repeated for the remaining two cameras (World Left and World Right), executing a distinct indoor exploratory path for each sensor.
\end{enumerate}
The poses supplied by the Magic Leap 2 internal system were chosen as the ground truth reference owing to their high reliability. By fusing data across multiple cameras, the depth sensor, and IMUs to rigidly anchor holograms in Euclidean space, the headset provides an operational accuracy substantially superior to monocular ORB-SLAM3.

\section{Analysis methodology and trajectory segmentation via \texttt{evo}}
The quantitative assessment of the estimation error was performed using the open-source package \texttt{evo}~\cite{grupp2017evo} (Python package for the evaluation of odometry and SLAM), the established benchmark standard within the robotics community for trajectory comparison and evaluation.
\subsection{Rationale for trajectory data segmentation}
Deploying purely monocular Visual SLAM in indoor environments entails well-documented theoretical limitations: the absence of direct metric depth perception and the presence of texture-deprived regions can induce transient visual feature tracking loss, leading to local drift or complete filter resets.
Because the primary objective of this experimental validation is to evaluate framework fidelity—specifically verifying that the transmitted video frames do not exhibit packet loss, anomalous jitter, or timestamp skew sufficient to corrupt downstream motion estimation—the analysis did not consider complete runs that were interrupted by algorithmic tracking failures. Instead, continuous, unbroken temporal segments wherein monocular tracking remained fully sustained were extracted from the recorded \texttt{rosbag} files and systematically evaluated.

\subsection{Error metrics and trajectory alignment}
The quantitative analysis was conducted by comparing the trajectory estimated by ORB-SLAM3 ($\mathbf{P}_{\text{est}}$) against the reference trajectory provided by the headset's on-board tracking system ($\mathbf{P}_{\text{ref}}$).

\subsubsection{Spatial and temporal alignment ($\text{Sim}(3)$ Umeyama)}
In monocular SLAM, the lack of depth sensors or direct inertial constraints renders the metric scale unobservable. Consequently, the estimated trajectory is defined up to an arbitrary scale factor $s > 0$ and an initial rigid body transformation relative to the reference coordinate frame.
To enable a consistent evaluation, the poses were temporally associated based on timestamps and aligned in Euclidean space via the Umeyama algorithm~\cite{umeyama1991least}, which minimizes the root-mean-square error to determine the optimal similarity transformation $\mathbf{S} \in \text{Sim}(3)$ comprising:
\begin{itemize}
    \item a rotation matrix $\mathbf{R} \in \text{SO}(3)$;
    \item a translation vector $\mathbf{t} \in \mathbb{R}^3$;
    \item a scale factor $s \in \mathbb{R}^+$.
\end{itemize}
The aligned estimated position at the $i$-th time step is therefore given by:
\begin{equation}
    \mathbf{p}_{\text{est}, i}^{\text{aligned}} = s \mathbf{R} \mathbf{p}_{\text{est}, i} + \mathbf{t}
\end{equation}

\subsubsection{Absolute Pose Error (APE)}
As the primary metric for global accuracy, the Absolute Pose Error (APE) applied to the translational component was computed. For each pair of corresponding poses at time step $i$, the point-wise error is defined as the Euclidean distance:
\begin{equation}
    e_i = \left\Vert \mathbf{p}_{\text{est}, i}^{\text{aligned}} - \mathbf{p}_{\text{ref}, i} \right\Vert
\end{equation}
Across the set of $N$ samples comprising the temporal segment, the following statistical metrics were extracted:
\begin{itemize}
    \item Root Mean Square Error (RMSE):
    \begin{equation}
        \text{RMSE} = \sqrt{\frac{1}{N} \sum_{i=1}^{N} e_i^2}
    \end{equation}
    \item Mean Error ($\mu$) and Standard Deviation ($\sigma$):
    \begin{equation}
        \mu = \frac{1}{N} \sum_{i=1}^{N} e_i, \qquad \sigma = \sqrt{\frac{1}{N} \sum_{i=1}^{N} (e_i - \mu)^2}
    \end{equation}
    \item Median, minimum, and maximum error.
\end{itemize}

\subsection{Translation and orientation decomposition}
In parallel with the scalar APE, the following aspects were evaluated:
\begin{itemize}
    \item The individual Cartesian components ($x(t), y(t), z(t)$), to assess the presence of directional drift along specific axes (e.g., along the vertical $z$-axis).
    \item The spatial orientation expressed in Euler angles (roll, pitch, yaw), to confirm the stability of rotational tracking and verify the absence of temporal latency or phase shifts in the data streamed by the bridge.
\end{itemize}
\clearpage

\section{Experimental results and trajectory analysis}
\subsection{Validation on World Center Camera — Trajectory 1}
The initial validation session entailed the acquisition of a continuous indoor trajectory lasting approximately $13.5\text{ s}$, covering an estimated total path length of $5.304\text{ m}$ ($5.359\text{ m}$ recorded by the on-board reference system). Throughout execution, the bridge streamed the monochrome video feed at full resolution ($1016 \times 1016$ pixels at 30 Hz) along with native poses to the ROS 2 host workstation. Temporal alignment yielded 367 uniquely matched pose pairs with a maximum temporal discrepancy below $0.01\text{ s}$ ($\Delta t_{\max} \le 10\text{ ms}$), verifying the absence of systematic latency or timestamp drift within the middleware layer.

\begin{figure}[htbp]
    \centering
    \includegraphics[width=0.78\textwidth]{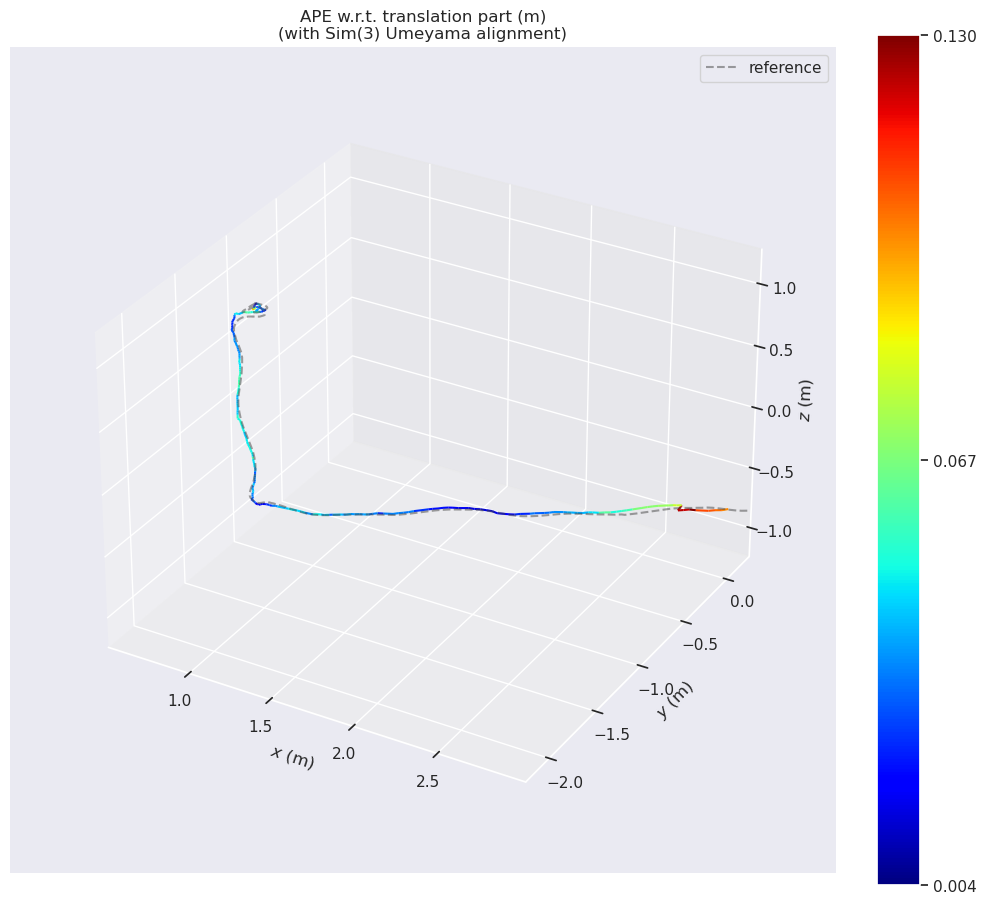}
    \caption{3D reconstruction of the estimated trajectory compared against the ground truth, colour-coded by translational APE following $\text{Sim}(3)$ Umeyama alignment.}
    \label{fig:Traiettoria_1_3D_map}
\end{figure}

\begin{table}[htbp]
\centering
\small
\renewcommand{\arraystretch}{1.2}
\begin{tabular}{@{}lcccccr@{}}
\hline
\textbf{Metric} & \textbf{RMSE [m]} & \textbf{Mean ($\mu$) [m]} & \textbf{Median [m]} & \textbf{Std ($\sigma$) [m]} & \textbf{Min [m]} & \textbf{Max [m]} \\ \hline
\textbf{APE (Translation)} & 0.0440 & 0.0389 & 0.0389 & 0.0206 & 0.0036 & 0.1297 \\ \hline
\end{tabular}
\caption{Statistical metrics for translational APE obtained via $\text{Sim}(3)$ Umeyama alignment for Trajectory 1 (World Center).}
\label{tab:ape_world_center_traj1}
\end{table}

\clearpage

\subsubsection{Analysis of the absolute pose error profile}

\begin{figure}[htbp]
    \centering
    \includegraphics[width=0.60\textwidth]{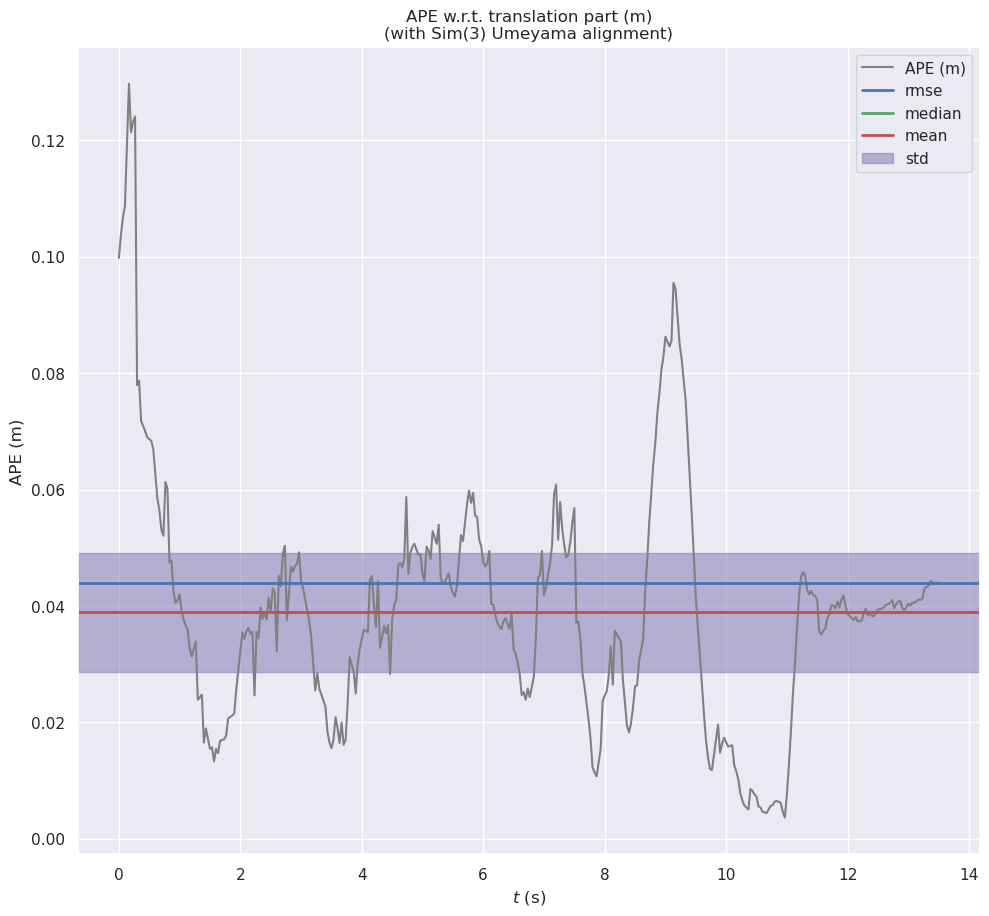}
    \caption{Temporal evolution of the translational Absolute Pose Error (APE) showing RMSE, mean, median, and standard deviation ($\pm\sigma$).}
    \label{fig:Traiettoria_1_ape}
\end{figure}

The root-mean-square error obtained on the translational component is $\text{RMSE} = 4.40\text{ cm}$, with a mean value of $\mu = 3.89\text{ cm}$ and low dispersion ($\sigma = 2.06\text{ cm}$). As highlighted by the temporal profile of the APE:
\begin{itemize}
    \item The error remains below $5\text{ cm}$ for almost the entire duration of the acquisition, reaching minimum values on the order of $3.6\text{ mm}$ during segments of uniform rectilinear motion.
    \item The maximum peak ($\approx 12.97\text{ cm}$) occurs during the initial transients of the monocular map initialisation (the first instants of local Bundle Adjustment convergence) and in correspondence with the abrupt change of direction at $t \approx 9\text{ s}$.
\end{itemize}
\clearpage
\subsubsection{Trajectory alignment and Cartesian axis decomposition ($X, Y, Z$)}

\begin{figure}[htbp]
    \centering
    \includegraphics[width=0.60\textwidth]{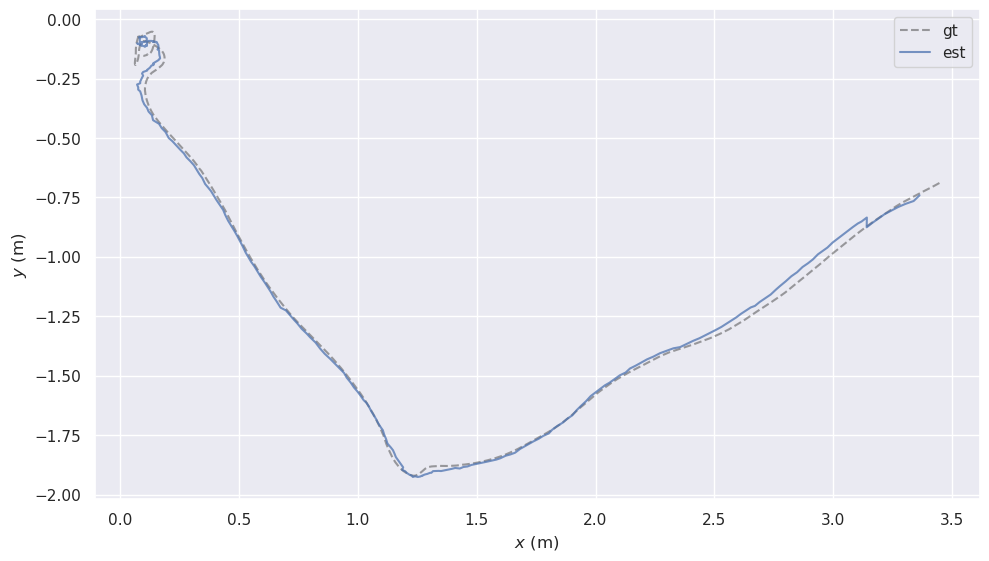}
    \caption{Two-dimensional projection on the $XY$ plane of the estimated trajectory compared against the reference trajectory.}
    \label{fig:Traiettoria_1_2D_map}
\end{figure}

\begin{figure}[htbp]
    \centering
    \includegraphics[width=0.60\textwidth]{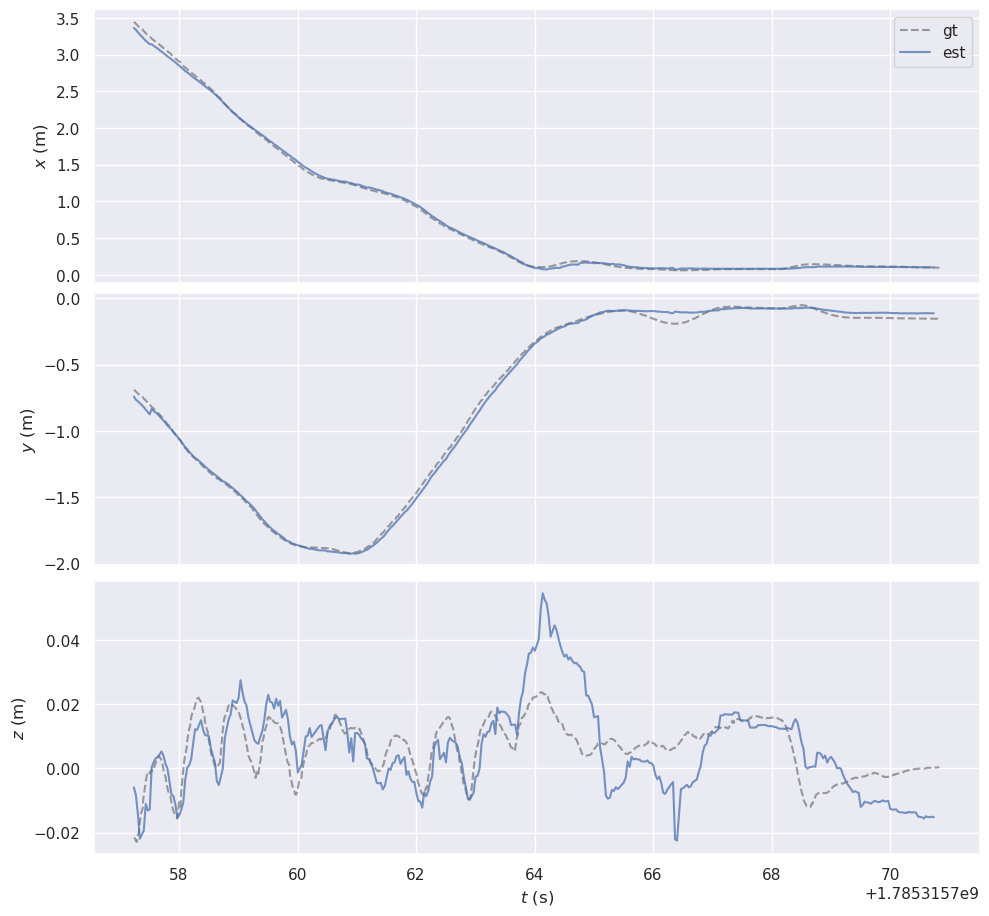}
    \caption{Temporal evolution of the individual Cartesian position components ($x(t)$, $y(t)$, $z(t)$): comparison between the estimate and the ground truth.}
    \label{fig:Traiettoria_1_traslation}
\end{figure}

The spatial evaluation across the $XY$ plane and within 3D space exhibits high geometric fidelity relative to the headset's ground truth. Time-domain Cartesian decomposition reveals that:
\begin{itemize}
    \item The horizontal axes $x(t)$ and $y(t)$ closely mirror the ground-truth kinematics, accurately preserving trajectory curvature changes and decelerations.
    \item The vertical axis $z(t)$ exhibits negligible fluctuations confined within a $\pm 2\text{ cm}$ envelope, characteristic of the operator's subtle walking oscillations.
\end{itemize}

\subsubsection{Orientation estimation (roll, pitch, yaw)}

\begin{figure}[htbp]
    \centering
    \includegraphics[width=0.75\textwidth]{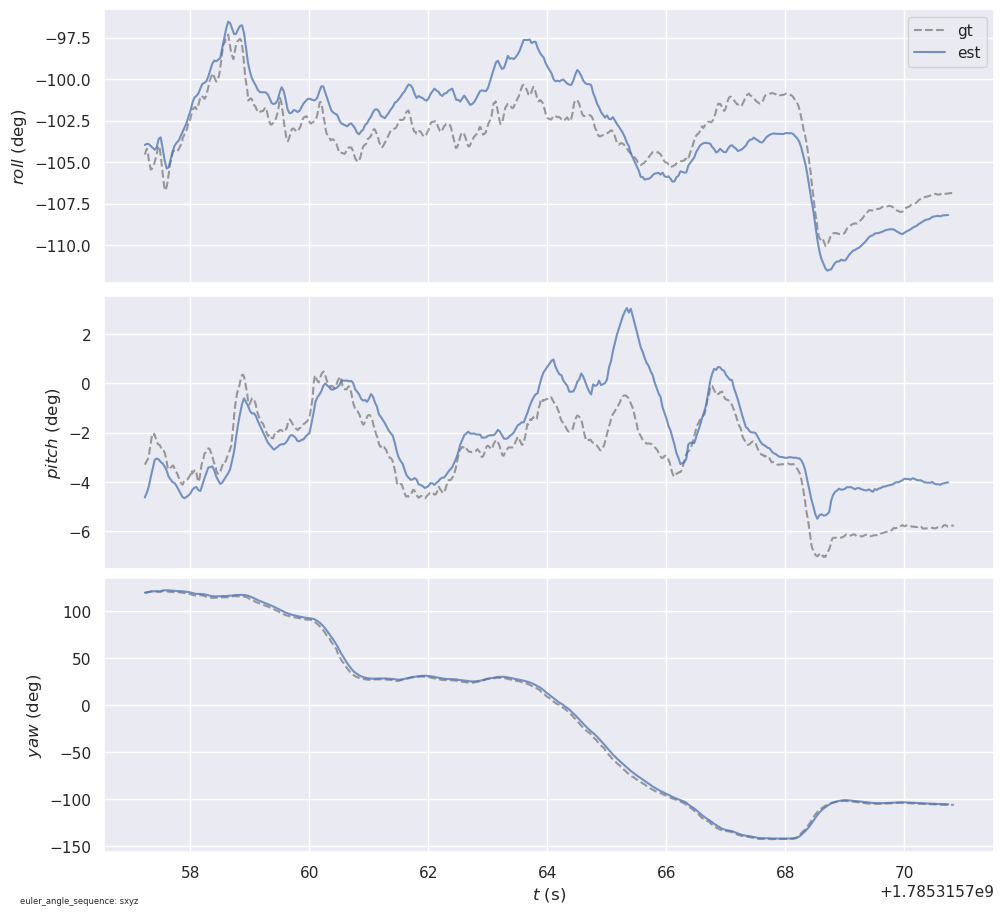}
    \caption{Temporal comparison of estimated Euler angles (Roll, Pitch, Yaw) against the headset's native on-board reference.}
    \label{fig:Traiettoria_1_rotation}
\end{figure}

The evaluation of the Euler angles substantiates the dynamic coherence of the frames streamed by the bridge:
\begin{itemize}
    \item The yaw angle, which undergoes an extensive rotation exceeding $200^\circ$ throughout exploration (ranging from $+120^\circ$ to $-105^\circ$), is tracked with nearly perfect overlap relative to the native reference trajectory.
    \item The roll and pitch angles exhibit mean angular discrepancies below $1.5^\circ$--$2^\circ$, confined strictly to the inherent differences between the purely vision-based estimation of ORB-SLAM3 and the on-board visual-inertial fusion executed by the Magic Leap 2.
\end{itemize}

\subsection{Validation on World Center Camera — Trajectory 2}
To corroborate the repeatability of the results, a second validation session was conducted along a different indoor path, characterised by a continuous trajectory lasting approximately $10.9\text{ s}$, covering an estimated total path length of $6.311\text{ m}$ ($5.960\text{ m}$ recorded by the on-board reference system). In this acquisition as well, the bridge streamed the monochrome video feed at $1016 \times 1016$ pixels at 30 Hz alongside the native headset poses. Temporal synchronisation yielded 305 uniquely matched pose pairs with a maximum temporal discrepancy below $0.01\text{ s}$ ($\Delta t_{\max} \le 10\text{ ms}$), reaffirming the robust stability of timestamp propagation.

\begin{figure}[htbp]
    \centering
    \includegraphics[width=0.82\textwidth]{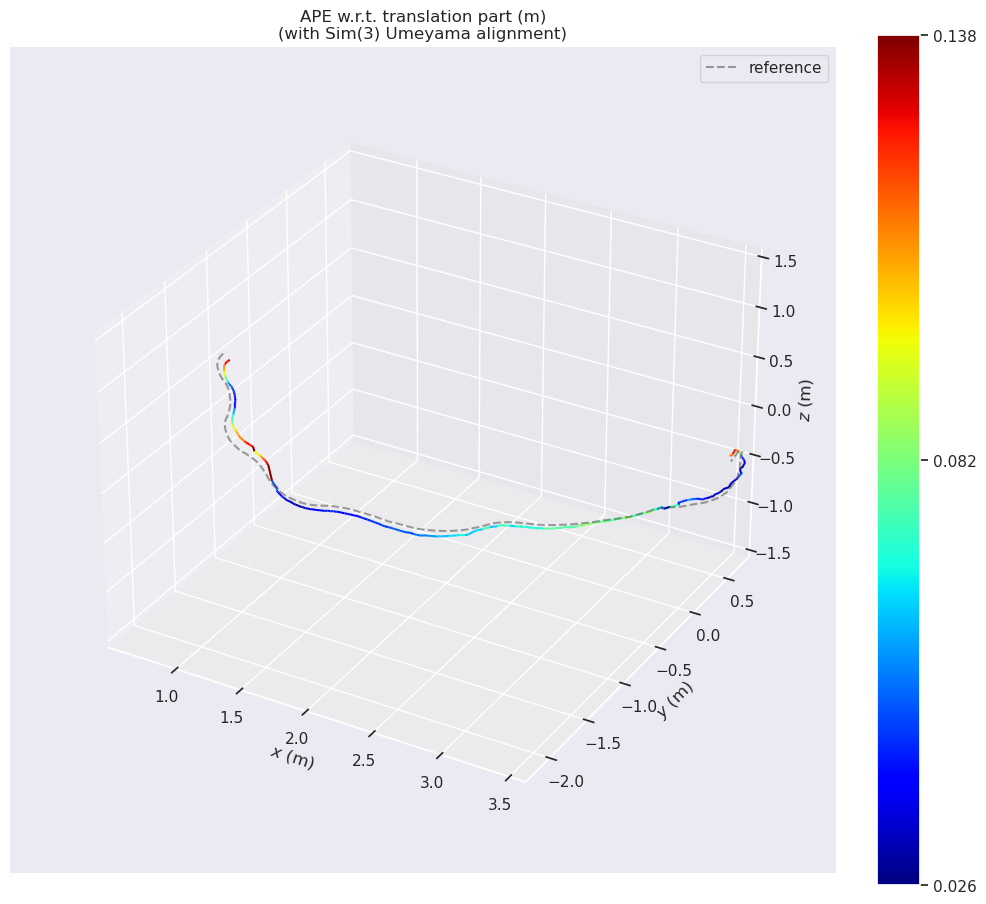}
    \caption{3D reconstruction of the estimated trajectory compared against the ground truth, colour-coded by translational APE following $\text{Sim}(3)$ Umeyama alignment.}
    \label{fig:Traiettoria_2_3D_map}
\end{figure}

\begin{table}[htbp]
\centering
\small
\renewcommand{\arraystretch}{1.2}
\begin{tabular}{@{}lcccccr@{}}
\hline
\textbf{Metric} & \textbf{RMSE [m]} & \textbf{Mean ($\mu$) [m]} & \textbf{Median [m]} & \textbf{Std ($\sigma$) [m]} & \textbf{Min [m]} & \textbf{Max [m]} \\ \hline
\textbf{APE (Translation)} & 0.0746 & 0.0681 & 0.0621 & 0.0305 & 0.0261 & 0.1378 \\ \hline
\end{tabular}
\caption{Statistical metrics for translational APE obtained via $\text{Sim}(3)$ Umeyama alignment for Trajectory 2 (World Center).}
\label{tab:ape_world_center_traj2}
\end{table}

\subsubsection{Analysis of the absolute pose error profile (APE)}

\begin{figure}[htbp]
    \centering
    \includegraphics[width=0.60\textwidth]{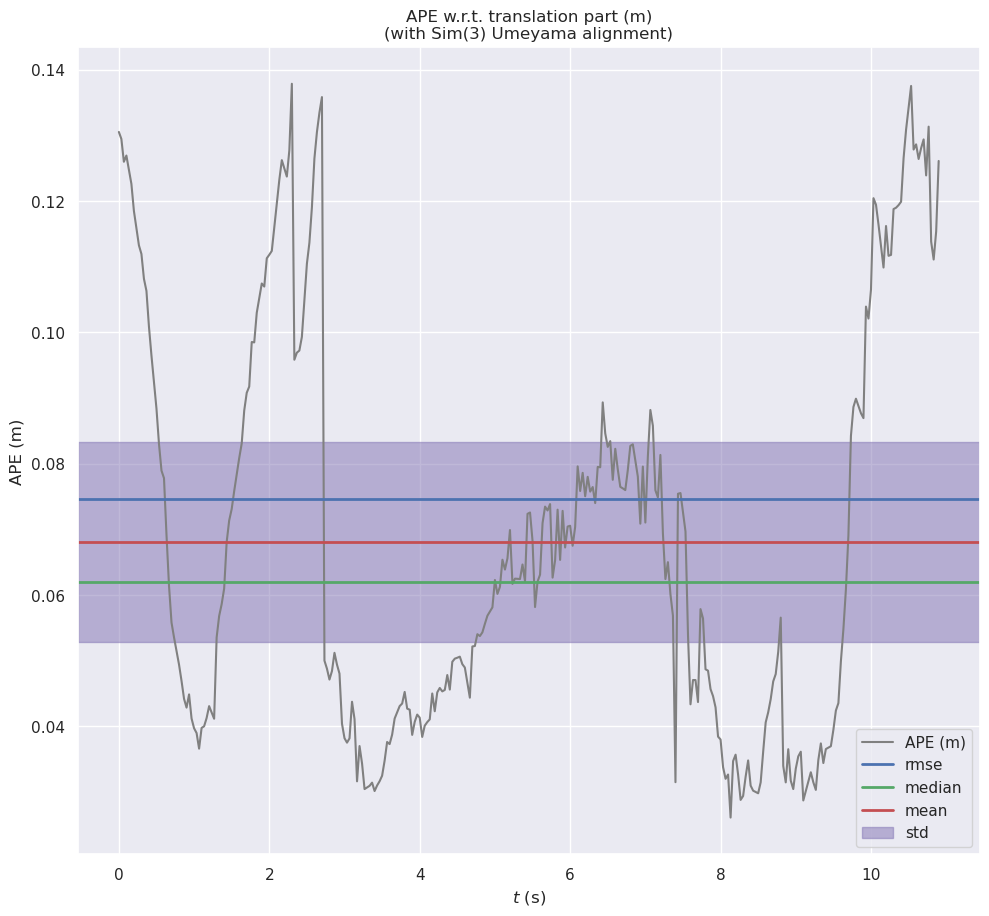}
    \caption{Temporal evolution of the translational Absolute Pose Error (APE) showing RMSE, mean, median, and standard deviation ($\pm\sigma$).}
    \label{fig:Traiettoria_2_ape}
\end{figure}

The root-mean-square error obtained on the translational component reaches $\text{RMSE} = 7.46\text{ cm}$, with a mean value of $\mu = 6.81\text{ cm}$, a median of $6.21\text{ cm}$, and a statistical dispersion of $\sigma = 3.05\text{ cm}$. As highlighted by the temporal evolution of the APE:
\begin{itemize}
    \item The error remains consistently below $8\text{ cm}$ throughout the central run, reaching minimum values of $2.61\text{ cm}$.
    \item The most pronounced error peaks ($\approx 13.78\text{ cm}$) are concentrated in the algorithm's initial convergence phase ($t \approx 0\text{--}0.5\text{ s}$) and in correspondence with the abrupt change of direction at $t \approx 9\text{--}10\text{ s}$, where rapid angular dynamics introduce a transient discrepancy into the monocular estimate.
\end{itemize}
\clearpage
\subsubsection{Trajectory alignment and Cartesian axis decomposition ($X, Y, Z$)}

\begin{figure}[htbp]
    \centering
    \includegraphics[width=0.60\textwidth]{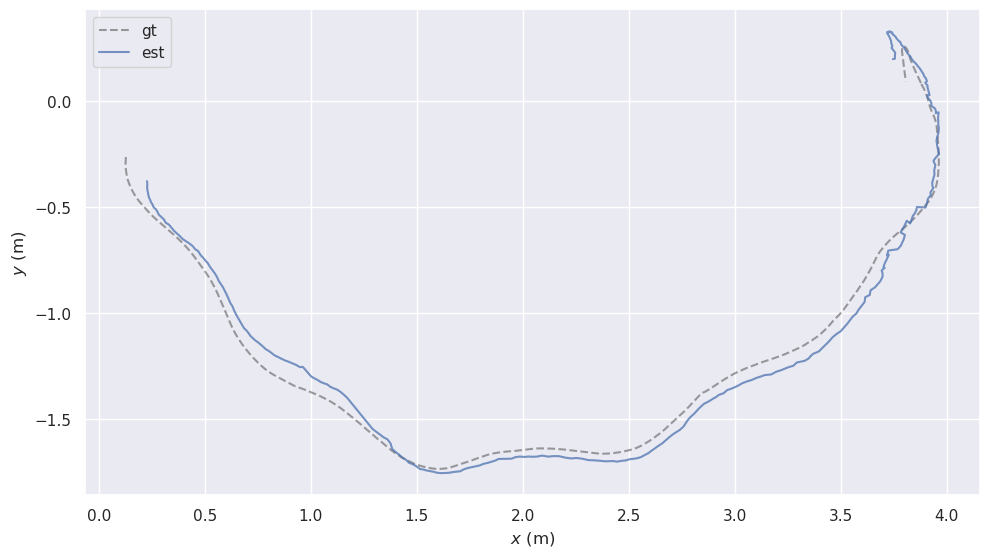}
    \caption{Two-dimensional projection on the $XY$ plane of the estimated trajectory compared against the reference trajectory.}
    \label{fig:Traiettoria_2_2D_map}
\end{figure}

\begin{figure}[htbp]
    \centering
    \includegraphics[width=0.60\textwidth]{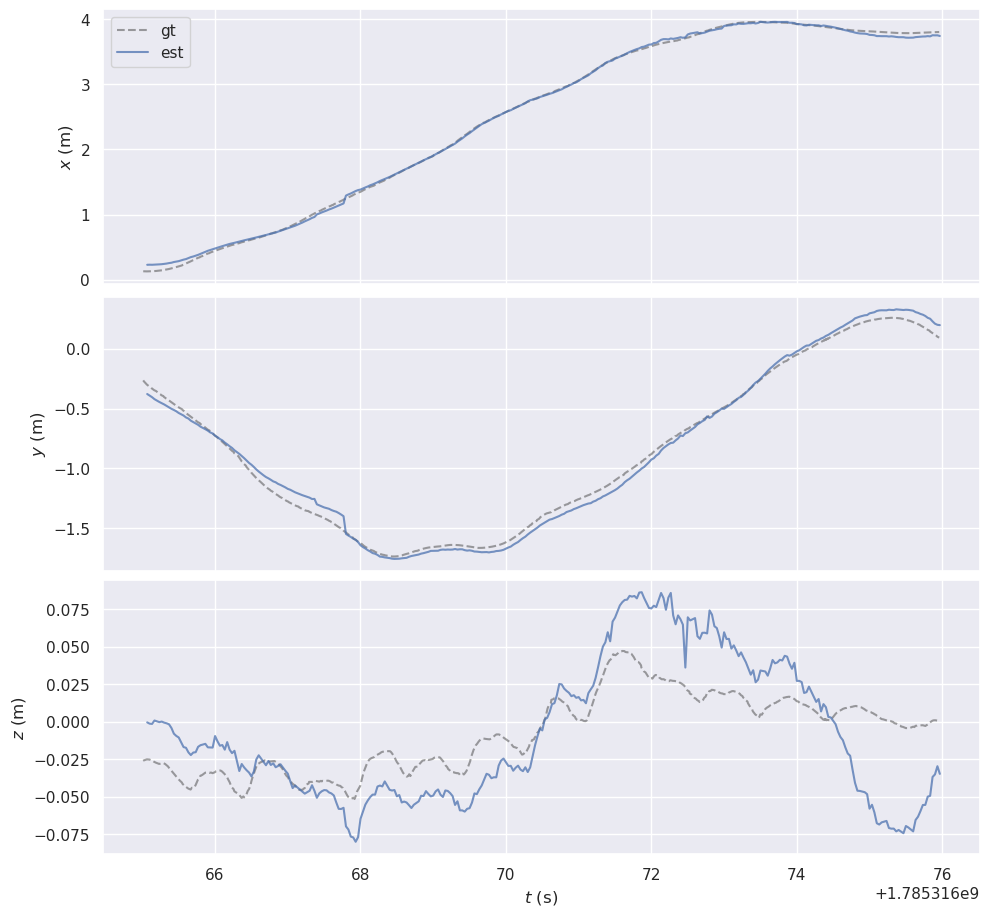}
    \caption{Temporal evolution of the individual Cartesian position components ($x(t)$, $y(t)$, $z(t)$): comparison between the estimate and the ground truth.}
    \label{fig:Traiettoria_2_traslation}
\end{figure}

The spatial evaluation across the $XY$ plane and within three-dimensional space exhibits remarkable geometric tracking along the entire curvilinear path:
\begin{itemize}
    \item The horizontal axes $x(t)$ and $y(t)$ accurately reproduce the temporal evolution of position, continuously matching both rectilinear segments and sections with varying curvature.
    \item The vertical axis $z(t)$ exhibits contained oscillations on the order of a few centimetres, demonstrating neither divergent vertical drift nor loss of metric consistency over time.
\end{itemize}

\subsubsection{Orientation estimation (roll, pitch, yaw)}

\begin{figure}[htbp]
    \centering
    \includegraphics[width=0.75\textwidth]{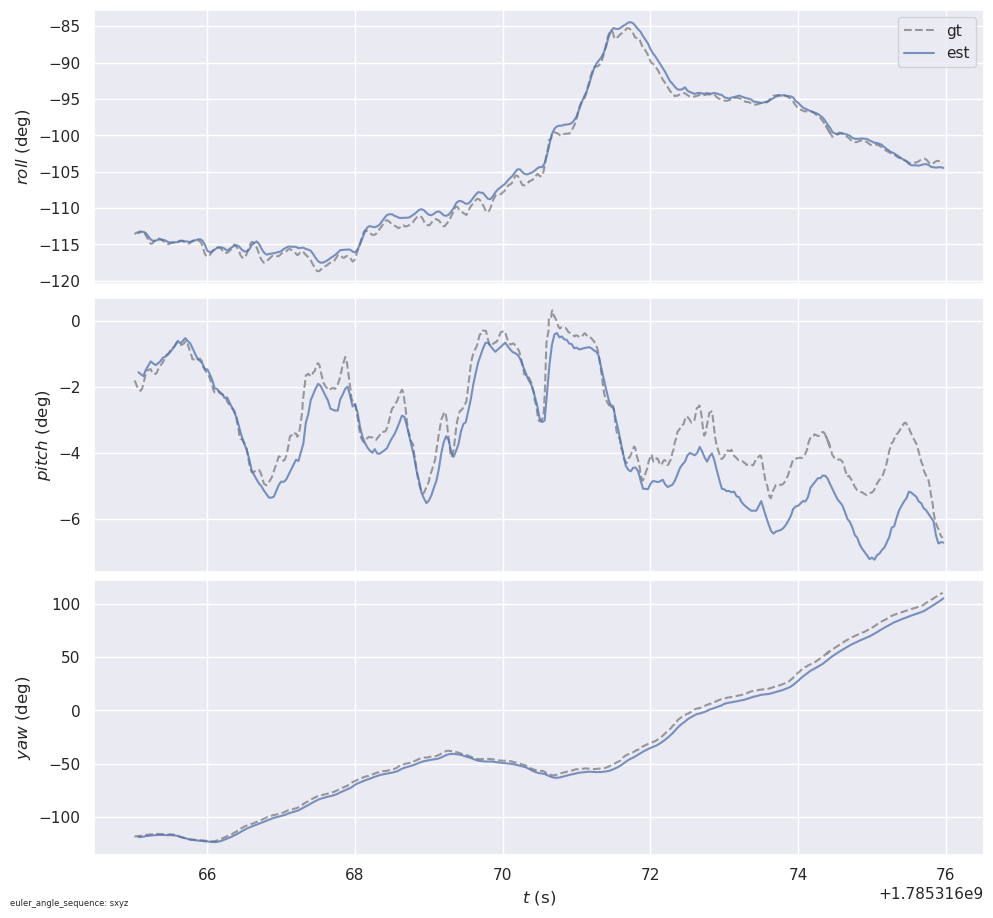}
    \caption{Temporal comparison of estimated Euler angles (Roll, Pitch, Yaw) against the headset's native on-board reference.}
    \label{fig:Traiettoria_2_rotation}
\end{figure}

The analysis of spatial orientation expressed in Euler angles confirms the absence of phase shifts or propagation delays between frames:
\begin{itemize}
    \item The yaw angle, which undergoes an extensive progressive rotation exceeding $220^\circ$ (from $-120^\circ$ up to over $+105^\circ$), is tracked with excellent accuracy and continuous overlap relative to the headset's ground truth.
    \item The roll and pitch angles faithfully follow the dynamic profile recorded by the Magic Leap 2, exhibiting minimal discrepancies attributable entirely to the differing characteristics of the two estimators (pure monocular optical estimation versus hardware visual-inertial fusion).
\end{itemize}

\subsection{Validation on World Left Camera — Trajectory 3}
The third experimental session was aimed at evaluating the sensory module of the lateral left camera (World Left) along an extended, closed-loop trajectory, in which the operator completed a full path returning to the vicinity of the starting position. The acquisition recorded a duration of approximately $22.77\text{ s}$, covering an estimated total path length of $11.961\text{ m}$ compared to the $12.277\text{ m}$ detected by the headset's on-board localisation system. Temporal synchronisation yielded 612 uniquely matched pose pairs with a maximum temporal discrepancy $\Delta t_{\max} \le 10\text{ ms}$, substantiating consistent frame ingestion even during prolonged scanning runs.

\begin{figure}[htbp]
    \centering
    \includegraphics[width=0.82\textwidth]{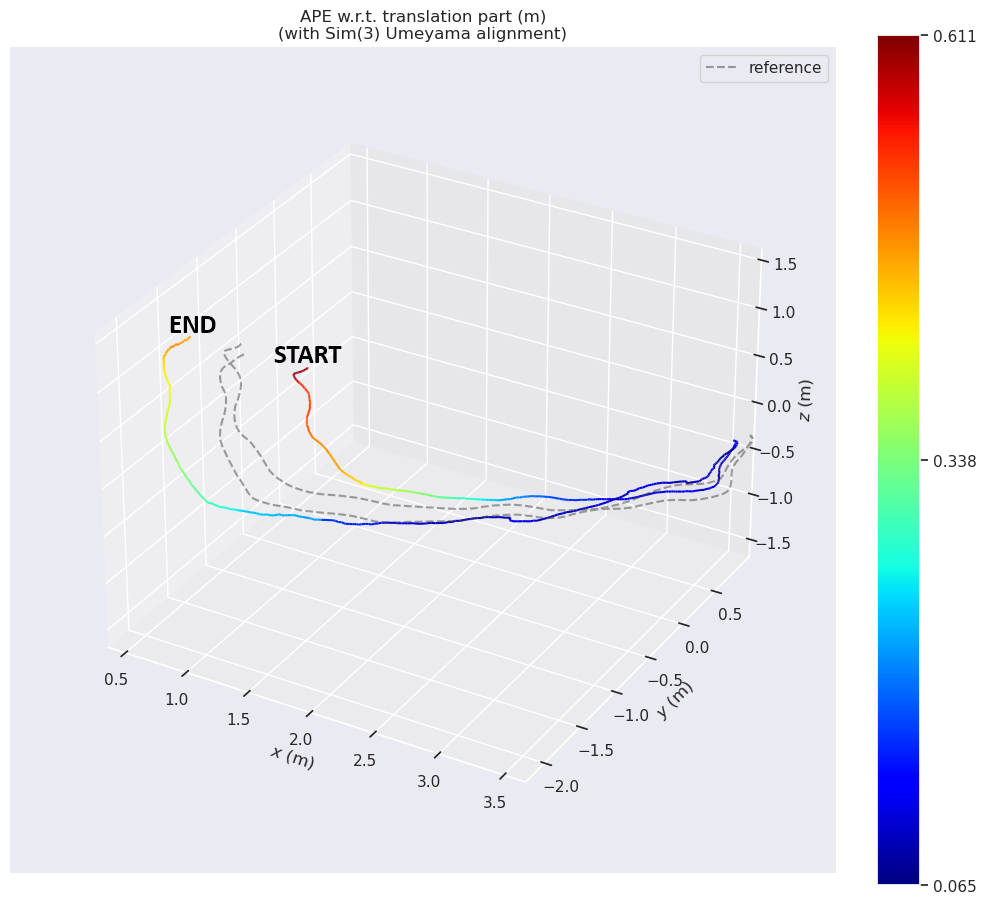}
    \caption{3D reconstruction of the estimated trajectory compared against the ground truth, colour-coded by translational APE following $\text{Sim}(3)$ Umeyama alignment.}
    \label{fig:Traiettoria_3_3D_map}
\end{figure}

\begin{table}[htbp]
\centering
\small
\renewcommand{\arraystretch}{1.2}
\begin{tabular}{@{}lcccccr@{}}
\hline
\textbf{Metric} & \textbf{RMSE [m]} & \textbf{Mean ($\mu$) [m]} & \textbf{Median [m]} & \textbf{Std ($\sigma$) [m]} & \textbf{Min [m]} & \textbf{Max [m]} \\ \hline
\textbf{APE (Translation)} & 0.3259 & 0.2765 & 0.2314 & 0.1724 & 0.0647 & 0.6109 \\ \hline
\end{tabular}
\caption{Statistical metrics for translational APE obtained via $\text{Sim}(3)$ Umeyama alignment for Trajectory 3 (World Left).}
\label{tab:ape_world_left_traj3}
\end{table}

\subsubsection{Analysis of the absolute pose error profile (APE)}

\begin{figure}[htbp]
    \centering
    \includegraphics[width=0.60\textwidth]{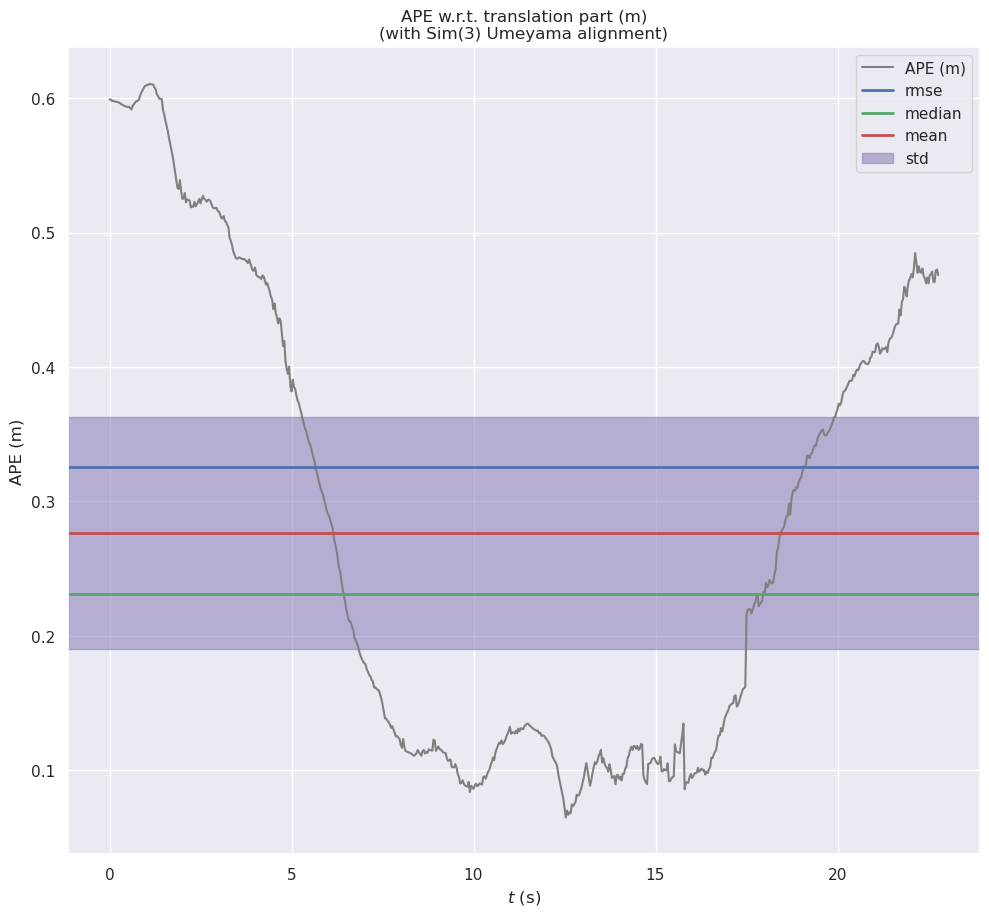}
    \caption{Temporal evolution of the translational Absolute Pose Error (APE) showing RMSE, mean, median, and standard deviation ($\pm\sigma$).}
    \label{fig:Traiettoria_3_ape}
\end{figure}

The root-mean-square error on the translational component reaches $\text{RMSE} = 32.59\text{ cm}$, with a mean value of $\mu = 27.65\text{ cm}$, a median of $23.14\text{ cm}$, and a statistical dispersion of $\sigma = 17.24\text{ cm}$. As this represents the longest and kinematically most challenging trajectory among those evaluated, the elevated absolute error relative to prior runs reflects the expected drift accumulation and subtle scale uncertainties characteristic of monocular SLAM in the absence of direct metric constraints. As highlighted by the temporal evolution of the APE:
\begin{itemize}
    \item The error decreases substantially throughout the central phase of the path ($t \approx 8\text{--}17\text{ s}$), remaining consistently below $15\text{ cm}$ and reaching a minimum of $6.47\text{ cm}$ across steady translation segments.
    \item The highest values ($\approx 61.09\text{ cm}$) manifest during the initial convergence phase and within the closing segment of the return path, where the combined effect of an extended odometric chain and perspective shifts accentuates the divergence between the pure vision-based estimate from ORB-SLAM3 and the headset's visual-inertial fusion used as pseudo-ground truth.
\end{itemize}
\clearpage
\subsubsection{Trajectory alignment and Cartesian axis decomposition ($X, Y, Z$)}

\begin{figure}[htbp]
    \centering
    \includegraphics[width=0.57\textwidth]{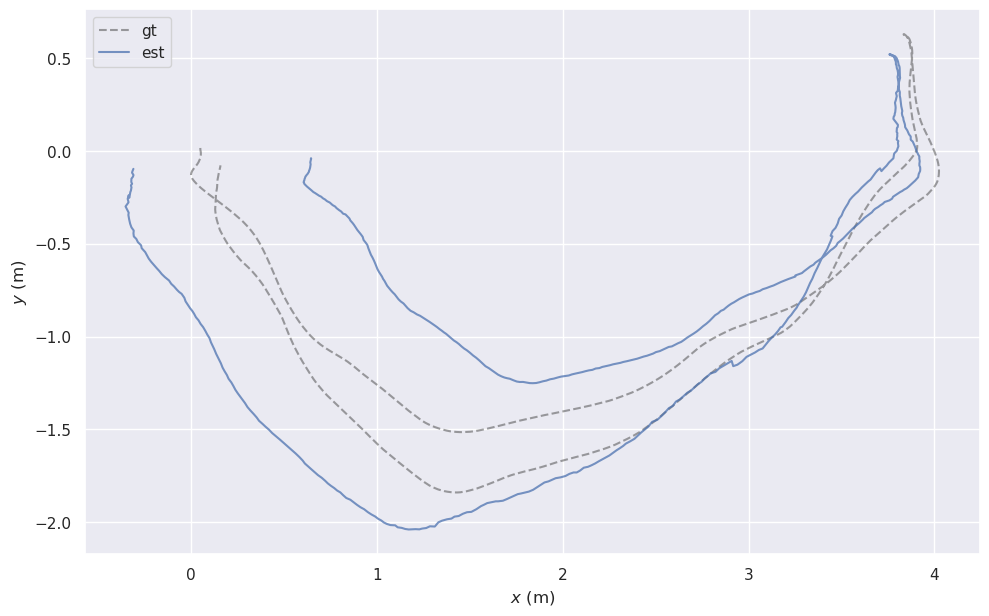}
    \caption{Two-dimensional projection on the $XY$ plane of the estimated trajectory compared against the reference trajectory.}
    \label{fig:Traiettoria_3_2D_map}
\end{figure}

\begin{figure}[htbp]
    \centering
    \includegraphics[width=0.57\textwidth]{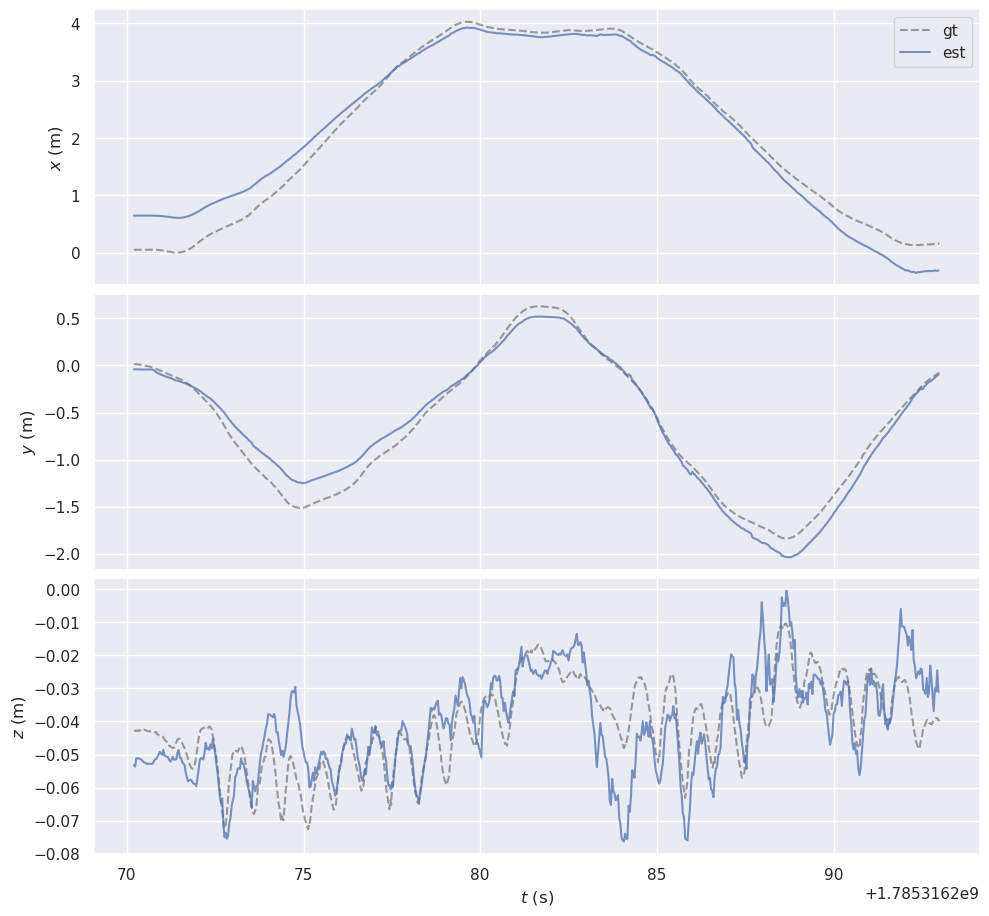}
    \caption{Temporal evolution of the individual Cartesian position components ($x(t)$, $y(t)$, $z(t)$): comparison between the estimate and the ground truth.}
    \label{fig:Traiettoria_3_traslation}
\end{figure}

The planar evaluation across the $XY$ plane and within three-dimensional space demonstrates robust topological consistency: the estimated trajectory closely traces the loop traversed by the operator. Decoupled analysis along the Cartesian axes indicates that:
\begin{itemize}
    \item The planar components $x(t)$ (exhibiting an excursion from $0\text{ m}$ to approximately $4.0\text{ m}$ and back) and $y(t)$ (exhibiting an excursion down to $-2.0\text{ m}$) reproduce the actual motion profile with high phase correlation, confirming the temporal integrity of the streams forwarded by the middleware.
    \item The vertical axis $z(t)$ remains bounded within a narrow envelope (between $-0.08\text{ m}$ and $0.00\text{ m}$), capturing the operator's gait without vertical divergence or structural map collapse.
\end{itemize}

\subsubsection{Orientation estimation (roll, pitch, yaw)}
\begin{figure}[htbp]
    \centering
    \includegraphics[width=0.75\textwidth]{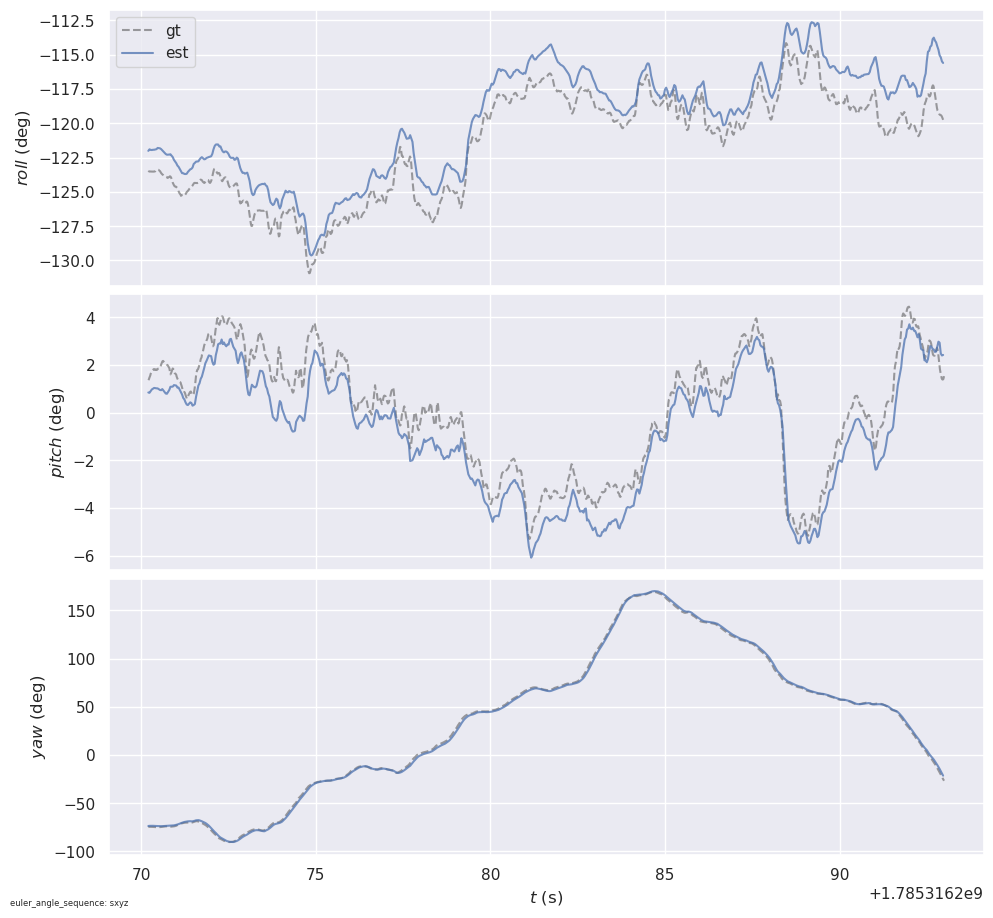}
    \caption{Temporal comparison of estimated Euler angles (Roll, Pitch, Yaw) against the headset's native on-board reference.}
    \label{fig:Traiettoria_3_rotation}
\end{figure}

The analysis of spatial orientation expressed in Euler angles confirms the high fidelity of the rotational estimate:
\begin{itemize}
    \item The yaw angle mirrors a demanding, wide rotation exceeding $250^\circ$ (spanning from $-80^\circ$ to over $+170^\circ$, subsequently settling back to $-20^\circ$) with excellent overlap, demonstrating precise synchronisation and zero frame loss during aggressive rotational manoeuvres.
    \item Roll and pitch accurately track high-frequency head dynamics, confirming streaming stability and fidelity even with lateral optics.
\end{itemize}

\subsection{Validation on World Right Camera — Trajectory 4}
To conclude the experimental validation across the optical tracking sensors, the fourth testing session evaluated the lateral right module (World Right). The acquisition encompassed a continuous indoor trajectory lasting approximately $15.97\text{ s}$, covering an estimated total path length of $9.206\text{ m}$ (against $7.966\text{ m}$ recorded by the native on-board tracking system). The bridge reliably streamed the monochrome video feed at full framerate alongside the headset poses. Temporal synchronisation yielded 438 uniquely matched pose pairs with a maximum temporal discrepancy $\Delta t_{\max} \le 10\text{ ms}$, reaffirming precise synchrony and the absence of cumulative latency introduced by the middleware layer.

\begin{figure}[htbp]
    \centering
    \includegraphics[width=0.82\textwidth]{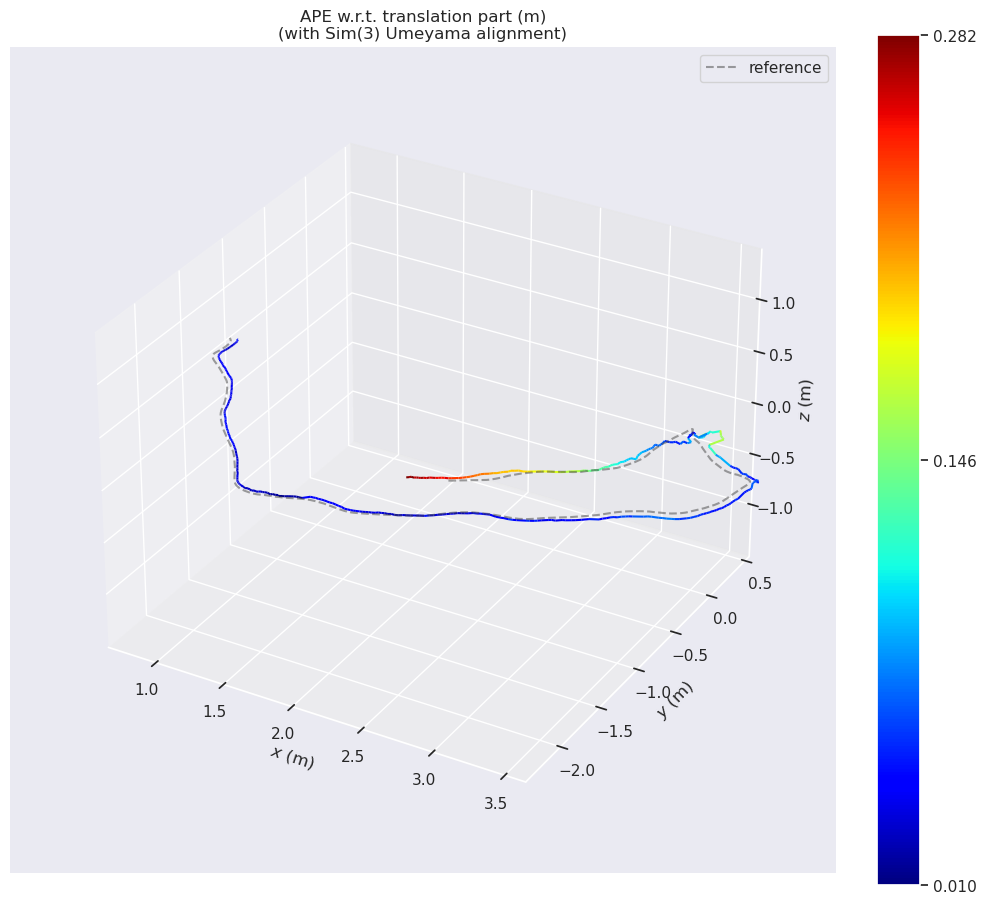}
    \caption{3D reconstruction of the estimated trajectory compared against the ground truth, colour-coded by translational APE following $\text{Sim}(3)$ Umeyama alignment.}
    \label{fig:Traiettoria_4_3D_map}
\end{figure}

\begin{table}[htbp]
\centering
\small
\renewcommand{\arraystretch}{1.2}
\begin{tabular}{@{}lcccccr@{}}
\hline
\textbf{Metric} & \textbf{RMSE [m]} & \textbf{Mean ($\mu$) [m]} & \textbf{Median [m]} & \textbf{Std ($\sigma$) [m]} & \textbf{Min [m]} & \textbf{Max [m]} \\ \hline
\textbf{APE (Translation)} & 0.0966 & 0.0746 & 0.0475 & 0.0614 & 0.0105 & 0.2819 \\ \hline
\end{tabular}
\caption{Statistical metrics for translational APE obtained via $\text{Sim}(3)$ Umeyama alignment for Trajectory 4 (World Right).}
\label{tab:ape_world_right_traj4}
\end{table}

\subsubsection{Analysis of the absolute pose error profile (APE)}

\begin{figure}[htbp]
    \centering
    \includegraphics[width=0.60\textwidth]{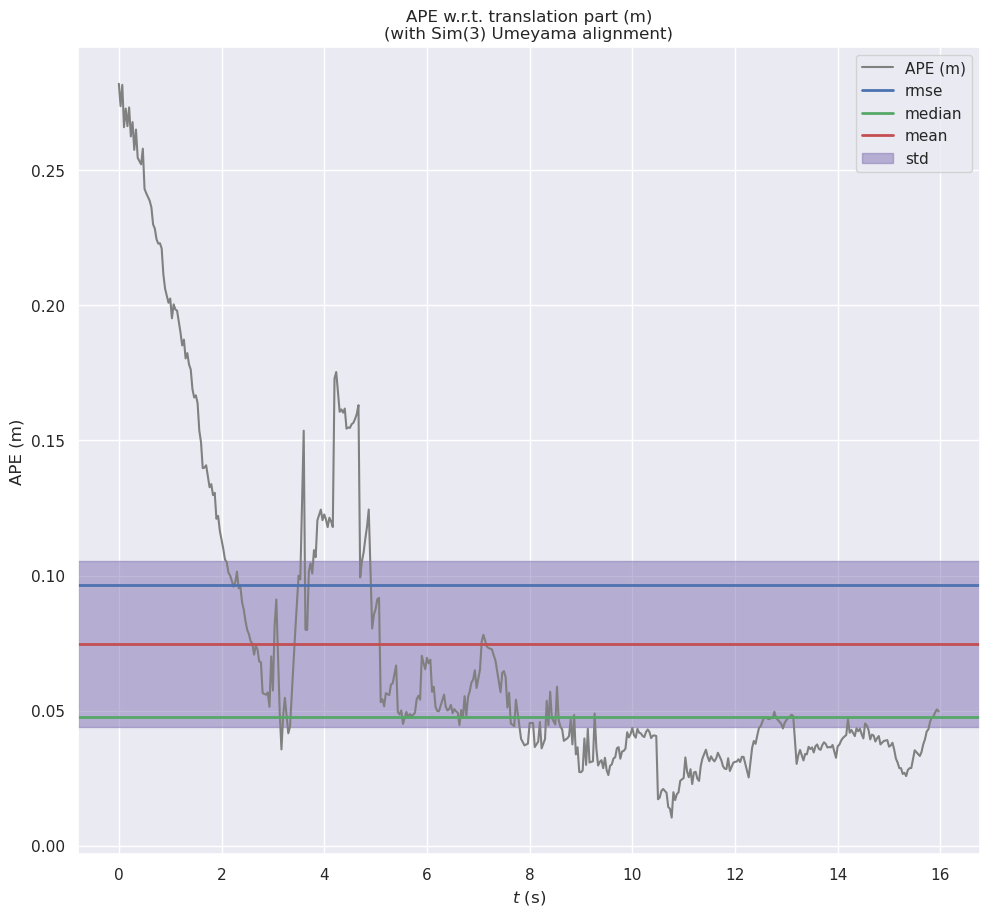}
    \caption{Temporal evolution of the translational Absolute Pose Error (APE) showing RMSE, mean, median, and standard deviation ($\pm\sigma$).}
    \label{fig:Traiettoria_4_ape}
\end{figure}

The root-mean-square error on the translational component reaches $\text{RMSE} = 9.66\text{ cm}$, with a mean value of $\mu = 7.46\text{ cm}$, a median of merely $4.75\text{ cm}$, and a low statistical dispersion ($\sigma = 6.14\text{ cm}$). The temporal profile of the APE highlights the excellent global tracking stability:
\begin{itemize}
    \item Following the initial transient phase, the error drops rapidly and stabilises for nearly the entire session ($t > 8\text{ s}$) below $5\text{ cm}$, reaching minimum values on the order of $1.05\text{ cm}$ during constant-velocity segments.
    \item The peak error ($\approx 28.19\text{ cm}$) is confined to the early initialisation instants ($t \approx 0\text{--}2\text{ s}$) and is subsequently mitigated by local bundle adjustment optimisation.
\end{itemize}

\clearpage

\subsubsection{Trajectory alignment and Cartesian axis decomposition ($X, Y, Z$)}

\begin{figure}[htbp]
    \centering
    \includegraphics[width=0.60\textwidth]{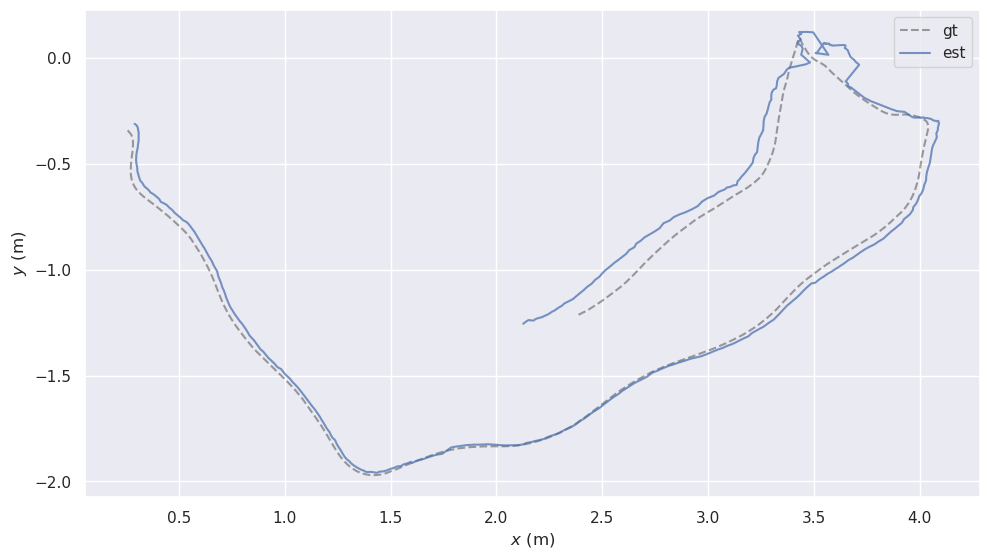}
    \caption{Two-dimensional projection on the $XY$ plane of the estimated trajectory compared against the reference trajectory.}
    \label{fig:Traiettoria_4_2D_map}
\end{figure}

\begin{figure}[htbp]
    \centering
    \includegraphics[width=0.60\textwidth]{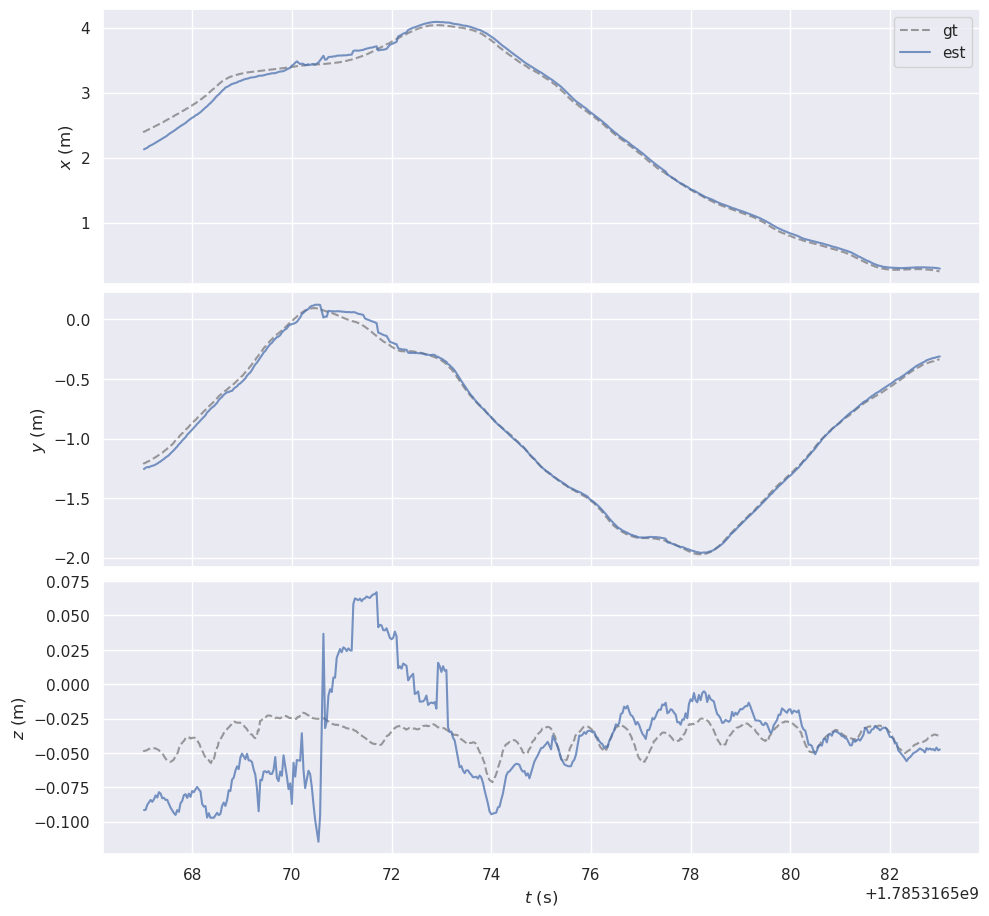}
    \caption{Temporal evolution of the individual Cartesian position components ($x(t)$, $y(t)$, $z(t)$): comparison between the estimate and the ground truth.}
    \label{fig:Traiettoria_4_traslation}
\end{figure}

The planar projection across the $XY$ plane and within three-dimensional space demonstrates remarkable geometric agreement with the ground-truth baseline. Time-domain Cartesian decomposition indicates that:
\begin{itemize}
    \item The horizontal axes $x(t)$ (spanning from $0.5\text{ m}$ to approximately $4.0\text{ m}$) and $y(t)$ (from $0.1\text{ m}$ to approximately $-2.0\text{ m}$) faithfully track the actual kinematics, capturing directional changes and wide-radius curves with excellent phase correlation.
    \item The vertical axis $z(t)$, while exhibiting a minor local fluctuation in the intermediate phase ($t \approx 7.0\text{--}7.4\text{ s}$) due to abrupt local scale adjustments, promptly recovers steady-state tracking; deviations remain bounded within a few centimetres without exhibiting divergent vertical drift.
\end{itemize}

\subsubsection{Orientation estimation (roll, pitch, yaw)}

\begin{figure}[htbp]
    \centering
    \includegraphics[width=0.75\textwidth]{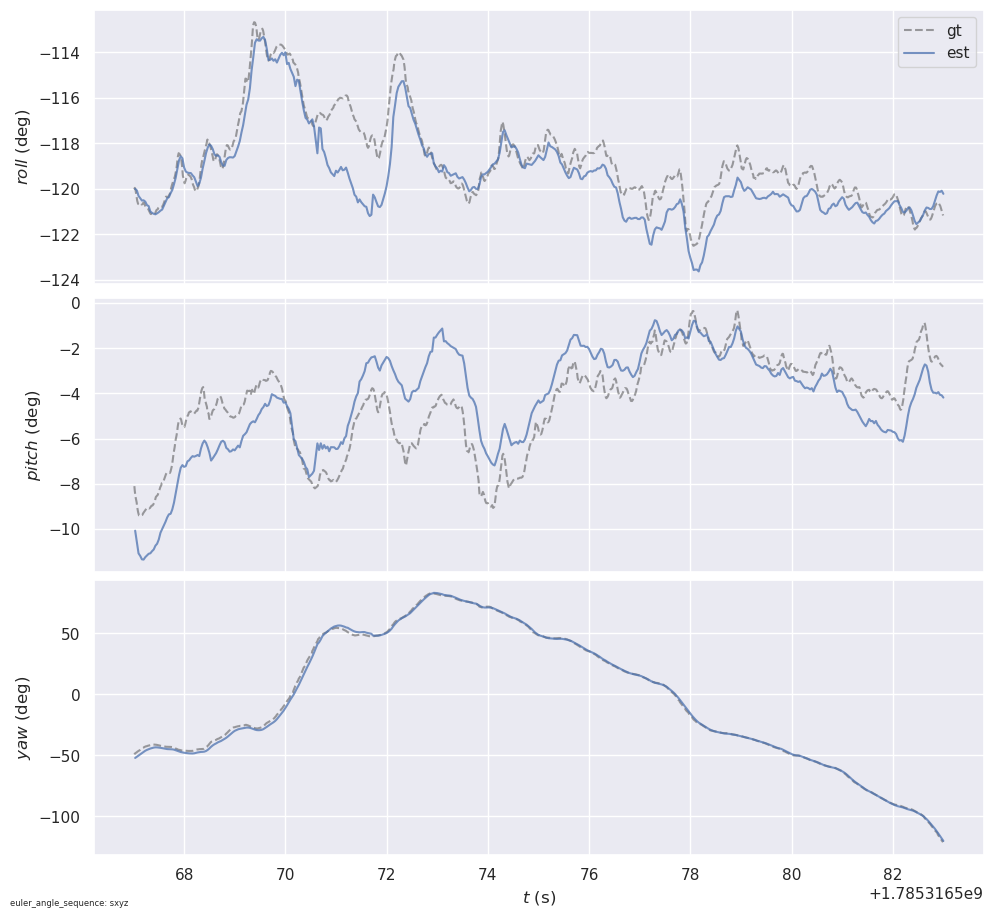}
    \caption{Temporal comparison of estimated Euler angles (Roll, Pitch, Yaw) against the headset's native on-board reference.}
    \label{fig:Traiettoria_4_rotation}
\end{figure}

The evaluation of spatial orientation expressed in Euler angles confirms the high kinematic accuracy of the transmitted frames:
\begin{itemize}
    \item The yaw angle tracks the continuous $\approx 200^\circ$ rotation with exceptional overlap (ranging from $-50^\circ$ to over $+80^\circ$ before descending to $-120^\circ$), confirming the absence of latency or timestamp discontinuities in the video stream.
    \item The roll and pitch angles accurately reproduce the dynamic oscillations of the operator's head, maintaining mean angular discrepancies confined to $1.5^\circ\text{--}2.5^\circ$, attributable solely to the methodological divergence between the vision-only estimator and the headset's proprietary inertial fusion pipeline.
\end{itemize}

\subsection{Final considerations on the experimental results}
The experimental phase enabled the practical efficacy of the developed architecture to be evaluated under real-world motion scenarios. The benchmark tests conducted with the ORB-SLAM3 visual localisation algorithm rigorously assessed all ambient tracking cameras on the headset, substantiating the bridge's capability to stream continuous, coherent, and uninterrupted data towards ROS 2.
Across all experimental trials, the framework ensured precise temporal synchronisation alongside the absence of perceptible latency or packet drops. The transmitted visual streams enabled the traversed paths to be reconstructed with remarkable fidelity relative to the headset's native tracking system:
\begin{itemize}
    \item Medium-range trajectories: Utilizing both the central and the lateral right optical modules, the estimated trajectory closely mirrored the actual physical displacements, maintaining mean discrepancies confined to a few centimetres while accurately tracing directional changes and curves.
    \item Head orientation dynamics: Rotational movements and operator gaze inclinations were reproduced with high accuracy throughout the entire duration of the tests, verifying the absence of temporal skew or phase lag during frame ingestion.
    \item Extended trajectories and loop closure: In the closed-loop acquisition using the left camera, the system preserved the global topology of the path; the progressive error growth observed towards the end of the run reflects the inherent challenge of estimating absolute scale using a solitary monocular camera over extended distances, yet without triggering tracking failure or filter divergences.
\end{itemize}
The experimental findings thus corroborate that the engineered infrastructure is robust, responsive, and primed for deployment in advanced applications, establishing a dependable foundation for spatial co-presence and real-time human-robot interaction.
    \cleardoublepage

    \chapter{Conclusions and Future Work}

\section{Conclusions}
Within modern industrial paradigms, Human-Robot Collaboration (HRC) and Industry 4.0 mandate continuous, intuitive information sharing across the shared workspace. Augmented Reality represents the ideal medium to establish this link; however, the absence of direct, native communication between commercial AR headsets and ROS 2-based robotic systems constitutes a substantial practical bottleneck. The primary objective of this thesis was specifically to overcome this constraint by developing an adaptable and efficient software bridge capable of seamlessly interfacing the Magic Leap 2 headset with the ROS 2 ecosystem. Leveraging the Unity development environment alongside the \texttt{ROS-TCP-Connector} package, an on-board application was engineered to acquire and stream data from all primary integrated sensors. The resulting infrastructure enables the streaming of the three ambient tracking cameras, the high-resolution front RGB camera (utilisable both for conventional video streaming and for mixed reality compositing with superimposed holograms), the Time-of-Flight depth sensor configurable for short-range manual interaction or long-range environmental mapping, the four dedicated eye tracking cameras for operator gaze monitoring, the Inertial Measurement Units (IMUs), and the ambient light sensor. A cornerstone of the project is the dynamic management of these streams: via a straightforward configuration file (\texttt{config.yaml}) handled by a dedicated ROS 2 node, the user can selectively determine at launch which sensors to activate, their target resolutions, and their respective framerates. This design prevents local network bandwidth saturation (particularly over Wi-Fi links), precludes device thermal throttling, and permits rapid system reconfiguration without requiring source code modifications or recompilation.
System robustness was experimentally validated by feeding real-time data from the World Cameras streamed through the bridge directly into the monocular ORB-SLAM3 pipeline. Benchmarking against the headset's internal tracking confirmed the high quality of the transmitted sensory data, as the estimated trajectories closely mirrored the physical displacements, exhibiting mean translational errors on the order of a few centimetres (between 4 and 9 cm across medium-range paths). This deviation is predominantly attributable to the intrinsic scale drift of monocular SLAM; conversely, the headset fuses multiple cameras, IMUs, and an active depth sensor, yielding substantially superior tracking accuracy. Furthermore, head motion and rotational dynamics (Roll, Pitch, Yaw) were reproduced with high fidelity without perceptible latency, and even along extended closed-loop trajectories (exceeding 12 metres traversed in approximately 23 seconds), the video stream remained unbroken and free from pipeline stalls. In conclusion, the developed framework constitutes a robust, responsive, and modular infrastructure. The open-source release of the complete project on GitHub~\cite{MagicLeap2_ROS2_Repo} makes this tool readily available to the scientific community and, in particular, to the IAS-Lab at the University of Padua, establishing a dependable, off-the-shelf foundation for engineering advanced applications wherein the headset ceases to be a mere display peripheral and operates as a fully fledged perceptual suite supporting human-robot interaction.

\section{Future work}
The modular software architecture and experimental findings established in this thesis provide a versatile foundation for extensions and real-world deployments within robotics and Human-Robot Collaboration (HRC). Future research trajectories encompass several fronts, with three being particularly prominent.
The first consists of bidirectional interaction and robotic intent visualisation. Integrating a downlink communication channel from ROS 2 back into the Unity graphical environment enables the augmented reality optical see-through display to project context-aware holograms directly into the operator's field of view. Specifically, the system will be able to render real-time computational decisions, planned manipulator trajectories, and safety bounding volumes in Euclidean space. This communicative transparency substantially mitigates user cognitive load and averts hazardous situations during close-proximity collaborative tasks. Furthermore, the same visual infrastructure can be deployed for procedural guidance in industrial assembly lines, projecting spatially registered step-by-step instructions directly onto physical workpieces during assisted assembly sequences.
The second development front concerns intuitive gesture-based control and multimodal interfaces. Exploiting the visual and depth streams provided by the bridge, the framework can be extended to integrate advanced hand tracking pipelines and natural gesture recognition algorithms. Operators will thus be enabled to dispatch direct commands to the manipulator, select targets in 3D space, or teleoperate collaborative robots rapidly and intuitively, superseding conventional industrial teach pendants. This paradigm aligns with cutting-edge hybrid interfaces for HRC, wherein augmented reality and gestural feedback collaborate to ensure dynamic control and active obstacle avoidance~\cite{Yan2024Complementary}.
The final front of development involves egocentric perception for wearable robotic systems and exoskeletons. In practice, the AR headset can be deployed not merely as a user interface peripheral, but as an auxiliary head-mounted egocentric vision module to assist mobile robotic systems and lower-limb exoskeletons~\cite{Mihailovic2025Egocentric}. Fusing head-mounted visual perception with ground-level exoskeleton sensor arrays facilitates dense 3D surface reconstruction, enabling early detection of distant obstacles or uneven terrain profiles, and empowering environment-adaptive gait planning strategies augmented by holographic visual feedback.
    \cleardoublepage

    \chapter*{Acknowledgements}
First and foremost, I would like to express my sincere gratitude to my supervisor, Prof.~Stefano Ghidoni, for involving me in this stimulating thesis project and for his continuous availability throughout its development. My profound gratitude extends equally to my co-supervisor, Dr~Matteo Terreran, for mentoring me step by step with remarkable dedication, for his invaluable technical guidance, and for his promptness in addressing every uncertainty, expertly navigating me towards the completion of this work.
My heartfelt thanks also go to my fellow student Federico Compagno, with whom I shared the trials, challenges, and milestones throughout the development of this project.\\ \\
The author used Gemini to assist in translating and polishing the language from the original Italian thesis. All content, technical claims, and final edits were reviewed and verified by the author.
    \cleardoublepage

    \printbibliography[heading=bibintoc]
\end{document}